%% file: main_neurips.tex
\pdfoutput=1
\documentclass{article}

\usepackage{styles/debstyle}   %

\usepackage[main, final]{styles/neurips_2026}

\usepackage[utf8]{inputenc}
\usepackage[T1]{fontenc}
\usepackage{microtype}
\usepackage{hyperref}
\usepackage{url}
\usepackage{booktabs}
\usepackage{amsfonts}
\usepackage{nicefrac}
\usepackage{xcolor}

\definecolor{darkblue}{rgb}{0, 0, 0.5}
\hypersetup{
  colorlinks=true,
  citecolor=darkblue,
  linkcolor=darkblue,
  urlcolor=darkblue,
  pdfborder={0 0 0},   %
  pdftitle={Don't Pay Attention, PLANT It: Pretraining Attention via Learning-to-Rank},
  pdfauthor={Debjyoti Saha Roy, Byron C. Wallace, Javed A. Aslam}
}

\title{Don't Pay Attention, PLANT It: Pretraining \\
Attention via Learning-to-Rank}

\author{%
Debjyoti Saha Roy, Byron C. Wallace \& Javed A. Aslam \\
Khoury College of Computer Sciences,
Northeastern University \\
\texttt{\{saharoy.d,b.wallace,j.aslam\}@northeastern.edu}
}

\begin{document}

\maketitle

\input{sections/content.tex}

\begin{ack}
We thank the anonymous reviewers for their constructive feedback, which helped us strengthen this paper.

\textbf{Funding.} This work received no third-party funding.
\textbf{Competing interests.} Javed A. Aslam is a co-founder of CodaMetrix, a startup in AI for administrative healthcare. CodaMetrix did not fund or take part in this work. The other authors declare no competing interests.
\end{ack}

\section*{Author Contributions}
Debjyoti Saha Roy conceived \plant, developed the method, wrote the code, designed and ran all experiments and analyses, and wrote the paper.
Byron C. Wallace reviewed the manuscript and provided suggestions that improved it.
Javed A. Aslam, as Debjyoti Saha Roy's PhD advisor, supervised the doctoral research of which this work is part.

\bibliographystyle{plainnat}   %
\bibliography{bib/references}

\appendix
\input{sections/appendix.tex}

\clearpage
\input{sections/neurips_checklist.tex}

\end{document}

%% file: sections/content.tex
\begin{abstract}

State-of-the-art Extreme Multi-Label Text Classification models rely on multi-label attention to focus on key tokens in input text, but learning good attention weights is challenging.
We introduce \plant---\textbf{P}retrained and \textbf{L}everaged \textbf{A}tte\textbf{NT}ion---a plug-and-play strategy for initializing the label-attention module that operates on top of an LLM's output representations.
\plant works by \emph{planting} label-specific attention using a Learning-to-Rank model trained on mutual information gain, before end-to-end fine-tuning.
This architecture-agnostic approach integrates with LLM backbones (e.g., \mistral, \llama, \deepseek, and \phiThree).
\plant outperforms SOTA methods across tasks like ICD coding, legal topic classification, and content recommendation, and improves fine-tuned LLMs in controlled comparisons outside the medical domain.
Gains are largest on rare labels, although a training-free LLM extraction baseline remains competitive on rarely seen labels. Ablation studies confirm that attention initialization is a key driver of these gains.
We make our code \href{https://github.com/deb-official/plant}{available}.
\end{abstract}

\section{Introduction}
Extreme Multi-Label Text Classification (XMTC) entails assigning the most relevant subset of labels to a given instance from a (very) large label set.
This setting emerges naturally in domains featuring vast, structured taxonomies such as e-commerce, legal categorization, and healthcare.
In such settings, manual labeling is both costly and error-prone.
For example, in clinical settings (see examples in Table~\ref{tab:icdehr}, App.~\ref{sec:appendix_implementation}), ICD coding---the task of assigning standardized codes for diagnoses and procedures based on clinical notes~\citep{moons2020comparison, who2025icd}---may be viewed as an instance of XMTC.

Building XMTC models is challenging due to the high-dimensional label space and heavily skewed label distributions \cite{bhatia2016extreme}.
For example, ICD-10-CM and ICD-10-PCS together define about 154,000 codes \citep{cms2026icd10}.
Many are rare: In the MIMIC-III dataset \cite{johnson2016mimic} approximately $5411$ out of $8929$ codes appear $<$$10$ times.
The task is further exacerbated by the often lengthy narratives in clinical texts.
For example, in the MIMIC-III dataset, discharge summaries frequently contain detailed clinical histories comprising a median of 1,375 words, and often exceeding 1500 tokens \citep{johnson2016mimic, johnson2023mimic, mullenbach2021clip, nguyen2023mimic}.
However, only a small fraction of these tokens are informative for assigning relevant ICD codes.

LLMs can be used zero-shot for XMTC tasks, but this poses challenges.
For instance, prompts for such tasks tend to include long and flat label lists, resulting in \textit{attention dilution}: The fixed attention budget is spread thin across thousands of tokens, weakening focus on rare tail labels~\citep{peysakhovich2023attention, vandemoortele2025haystack}.
This limitation is similarly evident in long-context retrieval tasks ~\citep{kamradt2023needle, hsieh2024ruler, liu2024lost}, where LLMs struggle to locate relevant items.
Task-specific fine-tuning may address such issues by embedding  knowledge of the labels directly into model parameters during training, obviating the need for attention over long label lists in the prompt and thus mitigating attention dilution~\citep{yang2023surpassing, boukhers2024large, zhang2025general, barreiros2025explainable}.

In current approaches to XMTC, \emph{attention mechanisms} \cite{bahdanau2014neural} help address the challenges of high-dimensional, skewed label spaces.
Existing XMTC models
~\citep{
lu2023towards, li2023towards, nguyen2023mimic, yang2023multi, chen2023rare, zhang2024novel, luo2024corelation} almost always  include a multi-label attention layer that allocates per-label attention weights to the input tokens ~\citep{wang2023multi,xiong2023xrr, yuan2024research,liu2025all}.
Intuitively, this is akin to a dedicated ``spotlight'' for each label: in high-dimensional spaces, it avoids the inefficiency of a single global focus by creating tailored text representations that highlight most relevant tokens per label.
In skewed distributions, this prevents subtle tail-label cues from being overwhelmed by head labels, improving rare-class prediction.

Regardless of the specific encoder architecture, removing this attention layer significantly harms performance.
A recent study by~\citet{xiong2023xrr} highlights the importance of label-specific attention for \emph{product-to-tag matching}
by showing that removing this component leads to a sharp drop in \patone (-15.69 points).
Elsewhere, results on \emph{scientific paper classification} show that stacking attention layers further boosts performance: Micro-F1 improves by a few points,
showing that deeper attention enhances the model’s capacity to represent label-specific features~\cite{liu2025all}.

\usetikzlibrary{arrows.meta,positioning,fit,calc,decorations.pathreplacing,matrix,backgrounds}
\begin{figure*}[t]
\centering
\begin{adjustbox}{max width=\textwidth}
\begin{tikzpicture}[
  font=\footnotesize,
  >=Latex,
  token/.style={draw, thick, rounded corners, minimum width=9mm, minimum height=6mm, inner sep=2pt},
  prob/.style={fill=purple!10, draw, thick, rounded corners, inner sep=3pt},
  ndcg/.style={fill=Teal!25, draw, thick, rounded corners, inner sep=3pt},
  outerbox/.style={draw=BurntOrange, very thick, rounded corners, inner sep=12pt},
  lossbox/.style={draw, rounded corners, very thick, minimum height=12mm, minimum width=50mm, inner sep=3pt},
  summationbox/.style={draw=black, very thick, rounded corners, fill=BurntOrange!50, minimum height=12mm, minimum width=12mm, align=center, text=black},
  labelbox/.style={font=\large},
  blocksmall/.style={draw, rounded corners, align=center, minimum height=28mm, minimum width=28mm, inner sep=3pt, font=\normalsize},
  thinblocksmall/.style={draw, rounded corners, align=center, minimum height=6mm, inner sep=1pt, minimum width=16mm, font=\scriptsize},
  axisarrow/.style={-Latex, line width=0.6pt}
]

\matrix[matrix of nodes, nodes=token, column sep=3mm] (rank) {
  \node (r1) {$t_{(1)}$}; &
  \node (r2) {$t_{(2)}$}; &
  \node (r3) {$t_{(3)}$}; &
  \node (r4) {$t_{(4)}$}; &
  \node (r5) {$t_{(5)}$}; \\
};

\node[draw, rounded corners, ultra thick, red!70, fit=(r2)] (Jbox) {};
\node[draw, rounded corners, ultra thick, blue!70, fit=(r4)] (Hbox) {};
\node[below=1mm of Jbox] {\scriptsize $\boldsymbol{j}$};
\node[below=1mm of Hbox] {\scriptsize $\boldsymbol{h}$};

\coordinate (midJH) at ($(Jbox.north)!0.5!(Hbox.north)$);
\draw[<->, bend left=25, very thick, Teal] (Jbox.north) to (Hbox.north);

\node[above=12mm of midJH, labelbox] (swaplabel) {$\mathsf{incorrect\;rank\;penalty}$};
\node[ndcg, below=1mm of swaplabel] (ndcgbox) {$|\Delta \mathsf{nDCG}@k|_{jh}$};

\node[prob, below=8mm of rank] (probbox)
  {$P(j \succ h)=\big(1+e^{-\sigma(s_{lj}-s_{lh})}\big)^{-1}$};
\draw[->, very thick, purple!80] (probbox.north) -- (Jbox.south);
\draw[->, very thick, purple!80] (probbox.north) -- (Hbox.south);

\node[below=6mm of probbox, labelbox] (title)
  {$\mathsf{Predicted\;ranking\;on}\;\mathcal{T}_l\;\mathsf{(by}\;s_{lj}\mathsf{)}$};

\node[outerbox, fit=(rank) (Jbox) (Hbox) (swaplabel) (probbox) (title)] (outer) {};
\node[below=6mm of outer.south, text=BurntOrange, labelbox] (miglabel)
  {$\mathsf{MIG\;relevance}\;r_{lj}\;\mathsf{selects\;top\!-\!k\;tokens\;per\;label}\;(\mathcal{T}_l)$};

\coordinate (outerE) at (outer.east);
\node[lossbox, right=16mm of outerE, anchor=west] (loss) {};

\coordinate (Lsw) at (loss.south west);
\coordinate (Lnw) at (loss.north west);
\coordinate (Lse) at (loss.south east);
\coordinate (Lne) at (loss.north east);
\def\lossSplit{0.52}
\coordinate (LmidN) at ($(Lnw)!\lossSplit!(Lne)$);
\coordinate (LmidS) at ($(Lsw)!\lossSplit!(Lse)$);

\path[fill=Teal!25]   (Lsw) -- (Lnw) -- (LmidN) -- (LmidS) -- cycle;
\path[fill=purple!15] (Lse) -- (Lne) -- (LmidN) -- (LmidS) -- cycle;
\draw[rounded corners, very thick] (Lsw) rectangle (Lne);

\node[align=center] at (loss.center)
  {$-\;|\Delta \mathsf{nDCG}@k|_{jh}\;\;\log_2 P(j\succ h)$};

\node[summationbox] (sum) at ($(loss.west)+(-8mm,0)$) {$\displaystyle\sum$};
\node[below=1mm of sum.south, text=BurntOrange, align=center, font=\scriptsize] (sumsub)
  {\parbox{12mm}{\centering
    $\mathsf{token\;pairs}$\\$\mathsf{from}\;\mathcal{T}_l$
  }};

\coordinate (sumloss-mid) at ($(sum.south west)!0.5!(loss.south east)$);

\path let
  \p1 = (sum.south west),
  \p2 = (loss.south east)
in
  node[below=10mm of sumloss-mid, align=center, labelbox] (losslabel)
    {$\underbrace{\hspace{\dimexpr\x2-\x1\relax}}_{\mathsf{\plant\;objective}}$};

\node[blocksmall] (mha) at ($(outer.west)+(-45mm,0)$)
  {\parbox{28mm}{\centering
    $\mathsf{MultiHead\!-\!Attn.}$\\$W_{\mathsf{attn}}$
  }};

\foreach \y/\name in {6/Q,0/K,-6/V} {
  \coordinate (in\name) at ($(mha.west)+(0,\y mm)$);
}

\coordinate (outS) at (mha.east);

\node[thinblocksmall, left=8mm of inQ] (Q) {$\mathbf{Q}=\mathbf{E}$\\ $\mathbb{R}^{|\mathcal L|\times h}$};
\node[thinblocksmall, left=8mm of inK] (K) {$\mathbf{K}=\mathbf{H}_i$\\ $\mathbb{R}^{n\times h}$};
\node[thinblocksmall, left=8mm of inV] (V) {$\mathbf{V}=\mathbf{H}_i$\\ $\mathbb{R}^{n\times h}$};

\node[thinblocksmall, right=8mm of outS] (S) {$\mathbf{S}$\\ $\mathbb{R}^{|\mathcal L|\times n}$};

\draw[axisarrow] (Q.east) -- (inQ);
\draw[axisarrow] (K.east) -- (inK);
\draw[axisarrow] (V.east) -- (inV);
\draw[axisarrow] (outS) -- (S.west);

\node[draw=gray!70, fill=gray!10, rounded corners, align=center, minimum height=20mm, minimum width=14mm, font=\normalsize, left=6mm of $(K.west)!0.5!(V.west)$] (llm)
  {$\mathsf{LLM}$\\$\mathcal{M}_{\mathsf{adapt}}$};
\draw[axisarrow] (llm.east) -- (K.west);
\draw[axisarrow] (llm.east) -- (V.west);
\node[draw=gray!70, dashed, rounded corners, inner sep=2.5mm, fit=(Q) (K) (V) (mha) (S),
  label={[font=\large, text=gray!40!black]above:$\mathsf{external\;attention\;on\;LLM\;outputs}$}] (extbox) {};

\draw[axisarrow] (S.east) -- (outer.west);

\node[below=1mm of S]
  {\parbox{13mm}{\centering\scriptsize
    $\mathsf{token\;attn.}$\\$\mathsf{per\;label}$
  }};

\end{tikzpicture}
\end{adjustbox}
\caption{
\textbf{\plant} Attention.
On the left, the $\mathsf{MultiHead\!-\!Attention}$ module \citep{vaswani2017attention}, parameterized by $\bs{W}_{\mathsf{attn}}$, takes as input
queries $\mathbf{Q}=\mathbf{E}$ (label embeddings), keys $\mathbf{K}=\mathbf{H}_i$,
and values $\mathbf{V}=\mathbf{H}_i$, and produces $\mathbf{S}\in\mathbb{R}^{|\mathcal L|\times n}$, the token-level attention for each label.
This module (dashed box) is \emph{external} to the LLM and distinct from its internal self-attention: its keys and values are the final hidden states $\mathbf{H}_i$ of the adapted backbone $\mathcal{M}_{\mathsf{adapt}}$.
The \textcolor{BurntOrange}{\textbf{orange box}} highlights the top-$k$ tokens per label, $\mathcal{T}_l$,
selected by \textcolor{BurntOrange}{Mutual Information Gain $r_{lj}$}.
Two tokens $j$ (\textcolor{red!70}{\textbf{red}}) and $h$ (\textcolor{blue!70}{\textbf{blue}}) in this set
are compared, with $j$ more relevant than $h$.
The module is trained to rank $j$ above $h$ (\textcolor{purple!80!black}{$P(j \succ h)$}), with each pair weighted by the change in $\mathsf{nDCG@k}$ if they were swapped (\textcolor{Teal!70!black}{$|\Delta \mathsf{nDCG}@k|_{jh}$});
the \textcolor{BurntOrange}{\textbf{summation box}} over token pairs in $\mathcal{T}_l$ gives the \plant objective used to initialize $\bs{W}_{\mathsf{attn}}$.
}
\label{fig:PLANT_arch}
\end{figure*}

{\bf The premise of this work is that we can be smarter about how we initialize attention module weights}.
Throughout, this attention module is an \emph{external} label--text cross-attention layer applied to the LLM's output hidden states (Figure~\ref{fig:PLANT_arch}); it is distinct from the self-attention layers inside the LLM backbone, which \plant does not modify.
SOTA XMTC models begin with random label attention weights, so each label must learn to rank all tokens from scratch. This is data-intensive in high-dimensional label spaces, and skew makes it worse: rare labels have the least data, and longer training tends to overfit them.
\citet{edin2023automated} show that SOTA models struggle to predict rare ICD codes: for PLM-ICD on MIMIC-IV, per-code \macrof\ rises with code frequency, and rare codes are predicted least accurately (Figure~\ref{fig:rare_codes_zoom}, left, digitized from their Figure~5).

Since removing label-specific attention sharply hurts XMTC models, we hypothesize that how this attention is initialized also matters: random initialization of the label-attention layer may keep rare labels from discovering shared attention structure.
We test this hypothesis in the experiments: removing \plant's attention pretraining costs more than any other ablation (Table~\ref{tab:plant_2stage_training_ablation_mimiciv}), and when training sees only common labels, randomly initialized attention reaches 0.5--1.1\% F1 on the held-out rare labels, versus 6.9--8.1\% with \plant (RQ7; Figure~\ref{fig:cosine_hisograms_and_freq_bin_transfer_bars}, right).

\begingroup
\begin{figure}[t]
    \centering
    \resizebox{\textwidth}{!}{
    \begin{subfigure}{0.5\linewidth}
        \centering
        \includegraphics[height=4.1cm]{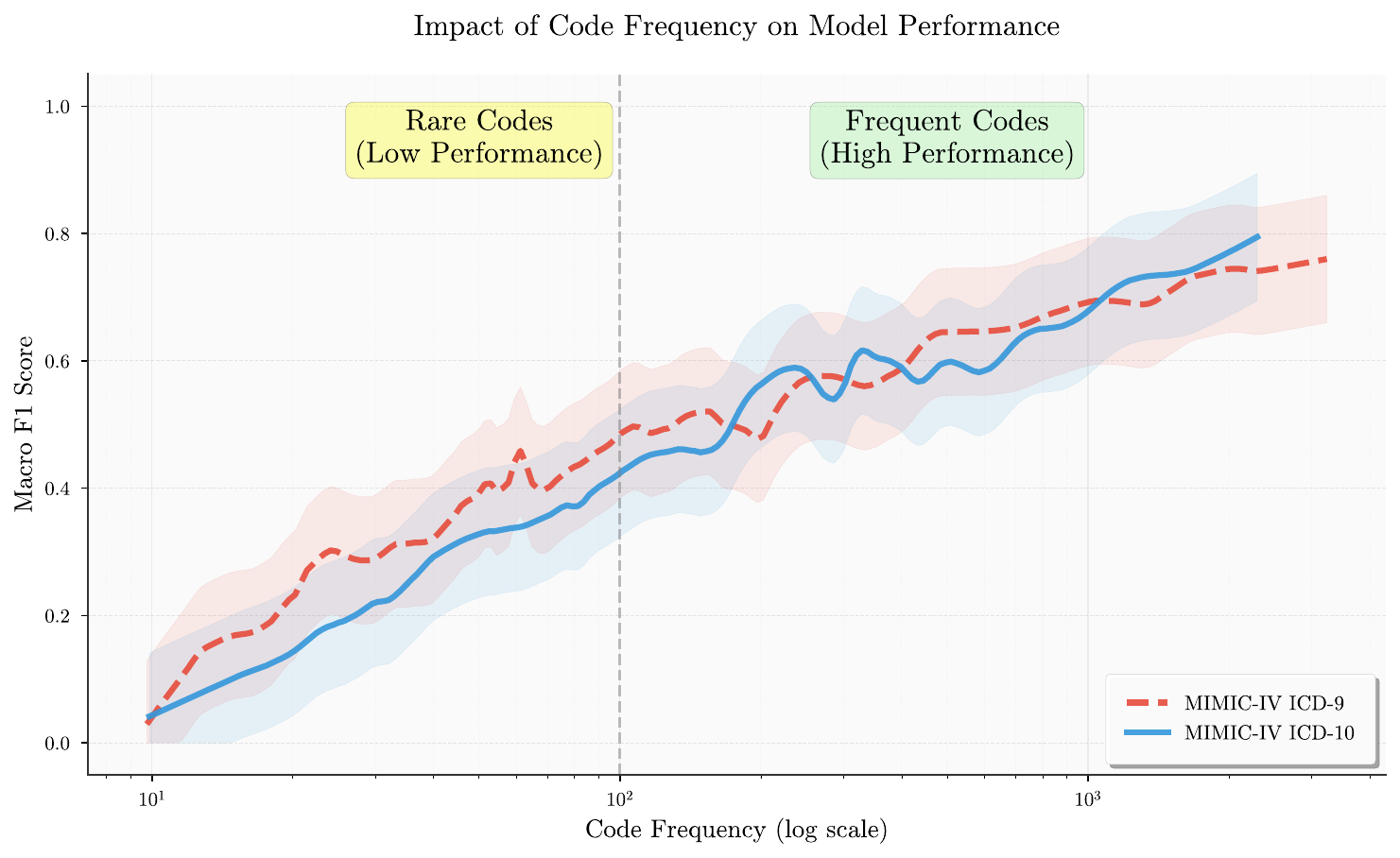}
    \end{subfigure}
    \begin{subfigure}{0.5\linewidth}
        \centering
        \includegraphics[height=4.1cm]{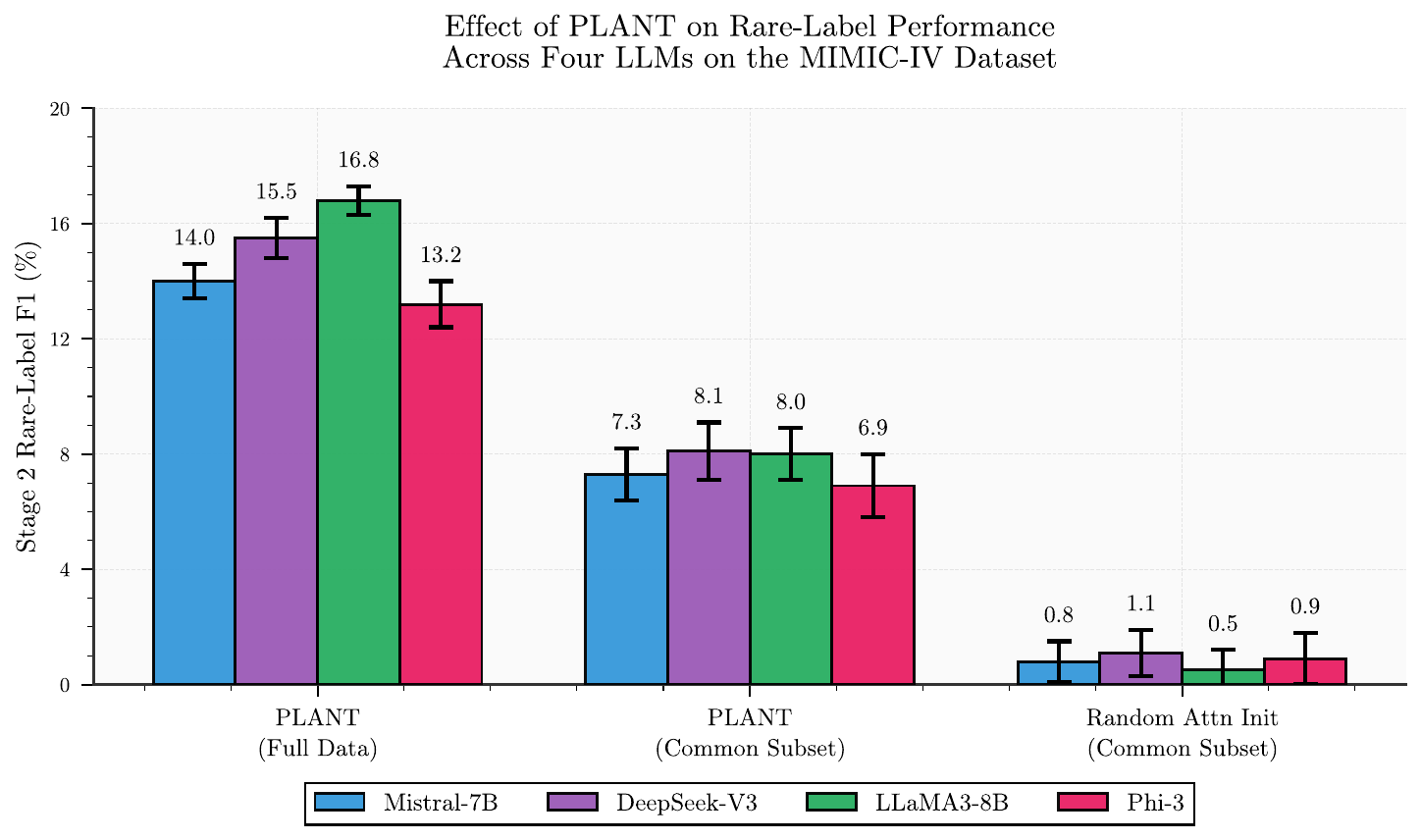}
    \end{subfigure}
    }
    \caption{\emph{(Left)} Per-code \macrof\ of PLM-ICD on MIMIC-IV against code frequency, digitized from Figure~5 of \citet{edin2023automated}: rare codes have near-zero \macrof\ (RQ3).
    \emph{(Right)} Rare-F1 when training only on common labels ($>1\%$ frequency). \plant retains zero-shot performance (6.9--8.1\%); random attention initialization collapses (0.5--1.1\%). See Section~\ref{subsection:freq_binned_transfer} (RQ7).}
    \label{fig:rare_codes_zoom}\label{fig:cosine_hisograms_and_freq_bin_transfer_bars}
\end{figure}
\endgroup

Our {\bf main contributions} are as follows:
(1) We introduce \plant(\textbf{P}retrained and \textbf{L}everaged \textbf{A}tte\textbf{NT}ion), a plug-and-play strategy for initializing the external label-attention module. \plant replaces random initialization with relevance-guided attention weights via a two-stage framework: Stage~1 pretrains the attention layer as a \emph{Learning-to-Rank} (L2R) module using \emph{mutual information}; Stage~2 leverages these weights to train the full model end-to-end, improving rare-label performance. \plant is architecture-agnostic and can be integrated with LLM backbones --- such as \mistral, \llama, \deepseek, or \phiThree --- without modifying them;
(2) In experiments across ICD coding, legal topic classification, and content recommendation, we report gains using \plant across backbones and datasets, compare against the same LLMs fine-tuned without \plant and against training-free zero-/few-shot LLM baselines, report where \plant does not help, and analyze through ablations which aspects of \plant are responsible for the gains.

\section{\plant}
\label{sec:approach}
In Extreme Multilabel Classification (XMTC) tasks, the goal is to assign to an input text multiple relevant labels from a very large label set.
Formally, denote the dataset by \(\mathcal{D} = \left\{ (\bs{x}_i, \bs{y}_i) \mid \bs{y}_i \in \{0,1\}^{\modu{\mathcal{L}}}, i = 1, \ldots, N \right\}\), where:
\(\bs{x}_i\) is an input instance (e.g., a text document), and
\(\bs{y}_i\) is a binary vector indicating the presence (\(y_{il} = 1\)) or absence (\(y_{il} = 0\)) of each label \(l \in \mathcal{L}\), where \(\mathcal{L}\) denotes the label set (which may contain tens of thousands of unique labels).
The objective is to learn a prediction function \(f_\theta : \bs{x}_i \mapsto \mathbb{R}^{\modu{\mathcal{L}}}\) parameterized by \(\theta\) that outputs labels for each input \(\bs{x}_i\).
For each label \(l \in \mathcal{L}\), the output \(f_\theta(\bs{x}_i)_l \in \mathbb{R}\) is the score assigned for the \(l\)-th label.

\textbf{Model Architecture.} \label{subsection:architecture}
We start with a pretrained transformer-based LLM \(\mathcal{M}_{\mathsf{base}}\), selected from widely used models such as \mistral, \llama, \deepseek, and \phiThree, known for strong general~\citep{team2023gemini, grattafiori2024llama, jiang2024mixtral, abdin2024phi} and domain-specific performance in ICD coding~\citep{yang2022large, yang2023transformehr, falis2024can, madan2024transformer, nerella2024transformers, asensio2025admission, he2025survey, liu2025application, yuan2025transformers}.
Due to computational constraints, we use both Low-Rank Adaptation (LoRA) and 4-bit quantization in all experiments~\citep{frantar2022gptq, hu2022lora, dettmers2023qlora, liu2024dora, aidouni2024masteringQLoRA}.
We adapt \(\mathcal{M}_{\mathsf{base}}\) into \(\mathcal{M}_{\mathsf{adapt}}\); see Appendix~\ref{app:training_details} for details.

Given an input
\(\bs{x}_i\), we tokenize it into \(\bs{t}_i\) and pass it through the adapted model \(\mathcal{M}_{\mathsf{adapt}}\) to obtain hidden states:
$\bs{H}_i = \mathcal{M}_{\mathsf{adapt}}(\bs{t}_i) \in \mathbb{R}^{n \times h}$,
where \(n\) is the sequence length and \(h\) the hidden size.

Specifically, we define trainable label embeddings \(\bs{E} \in \mathbb{R}^{\modu{\mathcal{L}} \times h}\), one per label.
In our main experiments, we initialize each label embedding from its textual description when available, encoded with a Sentence-Transformers model~\citep{reimers2019sentence}; \(\bs{E}\) is then refined in both training stages.
This semantic initialization benefits rare labels, whose representations need not be learned from only a few positive instances.
For datasets that provide only opaque label identifiers, \(\bs{E}\) is learned without this prior; \plant does not require label descriptions.
These embeddings serve as query vectors in a multi-head attention module. The module, denoted as \(\mathsf{MultiHead}\), defines learnable parameters \(\bs{W}_{\mathsf{attn}}\). It is applied on top of the backbone: its keys and values are the LLM's output hidden states \(\bs{H}_i\), so it is separate from the backbone's internal self-attention:
\begin{align}
\bs{Q}=\bs{E},\; \bs{K} \,\&\, \bs{V}=\bs{H}_i,\quad
\bs{A},\bs{S}=\mathsf{MultiHead}\footnotemark(\bs{Q},\bs{K},\bs{V};\bs{W}_{\mathsf{attn}}),\quad
\bs{A}\!\in\!\mathbb{R}^{\modu{\mathcal{L}}\times h},\; \bs{S}\!\in\!\mathbb{R}^{\modu{\mathcal{L}}\times n},
\label{eq:multiheadattention}
\end{align}
where \(\bs{Q}\), \(\bs{K}\), and \(\bs{V}\) represent the query, key, and value inputs, respectively. During training, the parameters \(\bs{W}_{\mathsf{attn}}\) are optimized to learn label-specific attention weights \(\bs{S}\), which determine how each label query attends to tokens in the sequence, and the output \(\bs{A}\), which represents the attended representations for each label query.

\footnotetext{See Appendix~\ref{app:hypers} for the number of attention heads in Equation~\ref{eq:multiheadattention} and output size \(p\) used in pooling.}

The resulting attention output is boosted by learnable matrices \(\mathbf{B}_a, \mathbf{B}_m \in \mathbb{R}^{|\mathcal{L}| \times h}\) as:
$
\mathbf{A}_{\text{boost}} = \mathbf{B}_a + \mathbf{A} \cdot \mathbf{B}_m$,
where \(\cdot\) denotes element-wise multiplication. This ``boosts'' label-specific signals in a learned, differentiable manner. This is motivated by the need to enhance task-specific signals in recent Mixture-of-Experts (MoE) frameworks~\cite{cai2024survey, yu2024boosting, chen2023mod}. An adaptive average pooling layer reduces the dimensionality of \(\bs{A}_{\text{boost}}\) to:
$\bs{P}_i = \mathsf{Pool}(\bs{A}_{\text{boost}}) \in \mathbb{R}^{\modu{\mathcal{L}} \times p}$.\footnotemark[\thefootnote]
\phantomsection\label{eq:avg_pooling}
A shared linear projection \(\bs{W}_c \in \mathbb{R}^{p \times 1}\) then computes the final logits as
$\hat{\bs{y}}_i = \bs{P}_i \bs{W}_c \in \mathbb{R}^{\modu{\mathcal{L}}}$,
each entry in $y_i$ is the predicted (raw) relevance score for the corresponding label.

\subsection*{\textbf{Two-Stage Training}}
\label{subsec:training}
\plant entails a two-stage optimization strategy.
The first focuses on pretraining \(\mathsf{MultiHead}\) (Equation~\ref{eq:multiheadattention}) label-wise attention weights; the second entails fine-tuning the model end-to-end.
We use \emph{attention pretraining} to mean only Stage~1: an L2R initialization of the external label-attention module before the task loss is used. It is not language-model pretraining; the backbone's own pretraining is left unchanged.

\textbf{\textsc{Stage 1: Pretraining Attention as L2R (Figure~\ref{fig:PLANT_arch})}}
In Stage 1 we train \(\mathcal{M}_{\text{adapt}}\) (through its LoRA adapters), \(\mathbf{E}\), and \(\mathsf{MultiHead}\) with a ranking loss.
The \(\mathsf{MultiHead}\) module outputs label-specific attention scores \(\bs{S} = [s_{lj}] \in \mathbb{R}^{|\mathcal{L}| \times n}\), where \(s_{lj}\) is the score for the token at position \(j\) of the input for label \(l\).

We train this following a learning-to-ranking objective focused on the top-\(k\) tokens per label selected by \hlorange{\textbf{Mutual Information Gain} (MIG)} computed from the training set.\footnotemark

\usetikzlibrary{tikzmark,calc,arrows,shapes,positioning}
\begingroup
\begin{figure*}[htp]
\centering
\begin{equation}
    \mathcal{L}_{\textsf{rank}}^{(i)} (\bs{S}) =
    - \sum_{l=1}^{|\mathcal{L}|} 
    \sum_{\tikzmarknode{mig}{\highlight{BurntOrange}{$\substack{j,h \in \mathcal{T}_l \\ r_{lj} > r_{lh}}$}}}
    \tikzmarknode{ndcg}{\highlight{Teal}{$|\Delta \mathsf{nDCG@k}|_{jh}$}}
    \times \log_2
    \tikzmarknode{prob}{\highlight{purple}{$\left(1 + e^{-\sigma (s_{lj} - s_{lh} )}\right)^{-1}$}}
    \label{eq:ranking_loss}
\end{equation}
\begin{tikzpicture}[overlay,remember picture,>=stealth,nodes={align=left,inner ysep=1pt},<-]
\path (mig.west) ++ (-2em, 0) node[anchor=east, color=BurntOrange!85, align=center] (migtext) {\textsf{\footnotesize MIG relevance}};
\draw [color=BurntOrange](mig.west) -- (migtext.east);
\path (ndcg.north) ++ (-0.5em,1.5em) node[anchor=east,color=Teal!85, align=center] (ndcgtext){\textsf{\footnotesize wt. for pairwise swap impact}};
\draw [color=Teal](ndcg.north) |- ([xshift=-0.0ex,color=Teal]ndcgtext.east);
\path (prob.south) ++ (-0.5ex,-1.5em) node[anchor=east,color=purple!85, text width=3cm,align=center] (probtext){\textsf{\footnotesize (differentiable) approx ranking prob.}};
\draw [color=purple](prob.south) |- (probtext.east);
\end{tikzpicture}
\end{figure*}
\endgroup
\footnotetext{See Appendix~\ref{app:mig} for details on pre-computing MIG for a corpus.}

where
\(\mathbf{r}_l = [r_{lj}] \in \mathbb{R}^n\) represents the ground-truth relevance of tokens for label \(l\): \(r_{lj}\) is the MIG, computed per vocabulary token, of the token at position \(j\);
\hlorange{$\mathcal{T}_l$ is the set of positions whose tokens are among the top-$k$ tokens for label $l$ by MIG}, so \(j\) and \(h\) index positions, not vocabulary entries;
\hlteal{$|\Delta \mathsf{nDCG@k}|_{jh}$ is the change in $\mathsf{nDCG@k}$ after swapping $j$ and $h$ in the predicted ranking}, where the predicted ranking is determined by sorting tokens in $\mathcal{T}_l$ by their predicted scores $s_{lj}$ in descending order; and
\(\sigma = 0.5\) is the sigmoid scaling factor. We test \(k = \{500, 1000, 2000\}\) to evaluate sensitivity.

\phead{\plant{ed} Attention via MIG Ranking}
\hlorange{MIG scores, denoted by $ r_{lj} $, quantifies how informative token $ t_j $ is for predicting label $ l $}. Higher \( r_{lj} \) indicates stronger relevance of the token to the label.
We compute MIG between token occurrences and label assignments over the training corpus (Appendix~\ref{app:mig}); the ranking loss (Equation~\ref{eq:ranking_loss}) pushes \(\mathsf{MultiHead}\) to give higher scores \(s_{lj}\) to more relevant tokens.
\hlorange{It considers token pairs $(j, h)$ where token $j$ is more relevant than token $h$ for a given label $l$, i.e., $r_{lj} > r_{lh}$}.
\hlpurple{The term $\left(1 + e^{-\sigma (s_{lj} - s_{lh} )}\right)^{-1}$ approximates the probability that token $j$ is ranked above token $h$.}
\hlteal{Each pair is weighted by $|\Delta \mathsf{nDCG@k}|_{jh}$, which penalizes incorrect rankings in proportion to their impact on the $\mathsf{nDCG@k}$ metric.}
This loss formulation encourages attention scores to align with the MIG.

\phead{Negation and context}
Because MIG is computed per token--label pair, generic negation cues (e.g., \emph{no}) usually receive low label-specific relevance.
\plant, however, attends over the LLM's \emph{contextualized} hidden states, so the representation of a condition token (e.g., \emph{pneumonia}) can encode a surrounding negation even when the cue itself has low MIG; Stage~2 then trains the full model end-to-end and can correct an imperfect Stage~1 prior.
We have not validated this with a controlled negation analysis (Section~\ref{sec:limitations}).

\textbf{\textsc{Stage 2: Leveraging Attention \--- Full Training}}
\label{sec:stage2_training}
In Stage 2 we train the entire model (Section~\ref{subsection:architecture}) end-to-end.
We start with the finetuned \(\mathcal{M}_{\text{adapt}}\) and the initialized weights \(\mathbf{W}_{\text{attn}}\) and \(\mathbf{E}\) from Stage 1.
We optimize the model under focal loss with label smoothing and hard negative mining to address label imbalance ~\citep{ben2020asymmetric, xiong2023xrr}.
The detailed formulation of the focal loss is provided in Appendix~\ref{app:focal_loss} (Eq.~\ref{eq:focal_loss}).

To address the challenge of imbalanced labels, where negative labels often dominate, hard negative mining is applied to focus the loss on the most informative examples. This selects all positive labels (\(y_{il} = 1\)) and the top-\(m\) negative labels (\(y_{il} = 0\)) with the highest predicted probabilities \(\sigma(\hat{y}_{il})\), where \(m = 1000\).
The focal loss is then computed over this selected subset \(\mathcal{S}_i \subseteq \mathcal{L}\) by restricting the summation in Eq.~\ref{eq:focal_loss} to \(\mathcal{S}_i\). We refer to this as the \emph{HNM-augmented focal loss}.
This approach ensures the model prioritizes learning from difficult negative examples, improving performance on challenging cases (Section~\ref{sec:ablation_study}).

\section{Experiments}
\label{sec:exp}
\textbf{Datasets, Baselines \& Implementation:}
We evaluated \plant against SOTA models on the \mimicboth datasets, which comprise discharge summaries annotated with ICD-10 and ICD-9 codes, respectively. For few-shot learning, we used \mimicrare and \mimicfew subsets to focus on rare codes.
To assess generalizability, we also evaluated \plant on publicly available
legal topic classification (\eurlex, over long legal documents) and
content recommendation (\wikiten, tag prediction for Wikipedia-style texts).
We further compare \plant with the same LLMs fine-tuned without \plant on sample-scale \eurlex and \wikiten subsets (RQ5), and with zero-/few-shot LLM baselines on a low-resource MIMIC-IV subset (RQ6).
\emph{Training protocol.} Both stages use early stopping with patience 2: Stage~1 runs for at most 10 epochs (validation $\mathsf{nDCG@k}$) and Stage~2 for at most 20 epochs (validation \macrof). In practice Stage~1 stopped at epochs 7--8 and Stage~2 at 12--15. Single-stage baselines use the same Stage~2 schedule and stopping rule, so their convergence criteria match. Reported training times are observed wall-clock times, not time caps. On \mimicfour, Stage~1 adds 11.8--27.6\% to total training time across the four backbones (e.g., 18.7\,h to 22.9\,h for \mistral; +20.8\% on average) and about 10\% to per-document inference time (App.~\ref{app:efficiency_benchmarks}, Table~\ref{tab:efficiency-comparison-mimiciv}).
Efficiency, datasets, implementation, baselines, and metrics are in App.~\ref{app:efficiency_benchmarks}, \ref{sec:appendix_implementation}, \ref{sec:appendix_baselines}, and \ref{sec:appendix_metrics}.

\textit{Notation.} \textsuperscript{\blueup}/\textsuperscript{\reddown} mark significant gains/drops ($\alpha{=}0.05$, Wilcoxon~\citealt{demvsar2006statistical}; see App.~\ref{sec:appendix_stats_significance}), shown if the test passes and 95\% CI excludes 0.
\textcolor{blue}{Gains}/\textcolor{red}{drops} followed by CI in \textcolor{CIPlum}{plum}.
\textbf{Bold} = best per metric.
\begingroup
\begin{table*}[t]
\caption{\label{tab:sota_llm_vs_plant_mimiciv}
        \textbf{\plant consistently boosts all LLM backbones on \mimicfour.}
        The \mimicfull{} results are in Table~\ref{tab:sota_llm_vs_plant_full} (App.~\ref{sec:appendix_results_detailed}); propensity scores are reported in Table~\ref{tab:sota_llm_vs_plant_mimiciv_psp}.
    }
    \centering
    \small
    \resizebox{\columnwidth}{!}{
    \begin{tabular}{l cc cc c}
        \toprule
        \multirow{2}{*}{Model}
        & \multicolumn{2}{c}{AUC}
        & \multicolumn{2}{c}{F1}
        & \multirow{2}{*}{P@15} \\
        \cmidrule(lr){2-3} \cmidrule(lr){4-5}
        & Macro & Micro & Macro & Micro & \\
        \midrule
        \mistral                 & 90.2 & 98.7 & 20.0 & 57.0 & 53.8 \\
        \rowcolor{green!20} \mistral + \plant       &
        \gain{97.4}{7.2} & \gain{99.5}{0.8} &
        \gain{23.0}{3.0} & \gain{59.2}{2.2} &
        \gain{56.9}{3.1} \\

        \llama                  & 90.5 & 98.8 & 20.5 & 57.5 & 54.0 \\
        \rowcolor{green!20} \llama + \plant         &
        \gain{\textbf{97.6}}{7.1} & \gain{\textbf{99.6}}{0.8} &
        \gain{\textbf{23.5}}{3.0} & \gain{\textbf{59.5}}{2.0} &
        \gain{\textbf{57.0}}{3.0} \\

        \deepseek               & 90.0 & 98.6 & 19.8 & 56.8 & 53.5 \\
        \rowcolor{green!20} \deepseek + \plant     &
        \gain{97.2}{7.2} & \gain{99.4}{0.8} &
        \gain{22.8}{3.0} & \gain{59.0}{2.2} &
        \gain{56.5}{3.0} \\

        \phiThree               & 89.8 & 98.5 & 19.5 & 56.5 & 53.2 \\
        \rowcolor{green!20} \phiThree + \plant      &
        \gain{97.0}{7.2} & \gain{99.3}{0.8} &
        \gain{22.5}{3.0} & \gain{58.8}{2.3} &
        \gain{56.3}{3.1} \\

        \midrule
        \rowcolor{green!20}
        \textbf{Avg.\ gain with \plant} &
        \avgain{7.2} & \avgain{0.8} &
        \avgain{3.0} & \avgain{2.2} & \avgain{3.1} \\
        \bottomrule
    \end{tabular}
    }
\end{table*}
\endgroup

\textbf{RQ1: How effective is \plant's two-stage across LLM backbones?}
Table~\ref{tab:sota_llm_vs_plant_mimiciv} compares the performance of four LLM backbones (\mistral, \llama, \deepseek, \phiThree)\footnotemark trained end-to-end with \emph{HNM-augmented focal loss}
versus \plant's two-stage training (Section~\ref{subsec:training}) which, in stage 2, adopts the same \emph{HNM-augmented focal loss}, on \mimicfull and \mimicfour.
We report the \emph{average absolute gains} obtained by computing the mean difference between each LLM and its \plant-enhanced counterpart for a given metric.

Notably, much smaller models integrated with \plant outperform significantly larger LLMs used alone. For instance, on \mimicfull, \llama+\plant (8B) outperforms \deepseek (671B total, 37B activated parameters) by \textbf{+1.8} in \textsf{F1} (Macro) and \textbf{+3.0} in \textsf{P@15}. Similarly, \phiThree+\plant (3.8B) surpasses \deepseek by \textbf{+2.0} in \textsf{F1} (Micro) and a substantial \textbf{+7.1} in \textsf{AUC} (Macro). This trend persists across \mimicfour as well: \llama+\plant achieves gains of \textbf{+3.7} in \textsf{F1} (Macro) and \textbf{+3.5} in \textsf{P@15} over \deepseek.
These results highlight the efficiency of \plant, which permits smaller models to surpass much larger LLMs across key metrics.
\plant's gains do not shrink systematically with backbone size: on \mimicfour it improves \deepseek by $+7.2$ \textsf{AUC} (Macro) and $+3.0$ \textsf{F1} (Macro), comparable to the 7--8B backbones. Where gains are small (e.g., \textsf{AUC} (Micro), already near 99\%), performance saturation is a plausible but untested explanation.

\footnotetext{See Appendix~\ref{app:training_details} for LLM details.}

\begin{table*}[t]
\caption{\label{tab:results_mimicfour}
    \textbf{\plant sets a new SOTA on \mimicfour.} CoRelation's \patfifteen is from our own evaluation.
}
\centering
\small
\resizebox{\textwidth}{!}{
\begin{tabular}{l ll ll lll}
\toprule
\multirow{2}{*}{Model} & \multicolumn{2}{c}{$\mathsf{AUC}$} & \multicolumn{2}{c}{$\mathsf{F1}$} & \multicolumn{2}{c}{$\mathsf{Precision}$} \\
\cmidrule(lr){2-3} \cmidrule(lr){4-5} \cmidrule(lr){6-7}
 & Macro & Micro & Macro & Micro & \pateight & \patfifteen \\
\midrule
CoRelation~\citep{luo2024corelation} & $97.2$ & $\mathbf{99.6}$ & $6.3$ & $57.8$ & $70.0$ & $55.3$ \\
PLM-CA~\citep{edin2024unsupervised} & $91.8$ & $99.1$ & $22.3$ & $58.9$ & $70.5$ & $55.8$ \\
GKI-ICD~\citep{zhang2025general} & $97.1$ & $99.3$ & $20.6$ & $58.5$ & $70.7$ & $55.8$ \\
GPT-4 Zero-Shot~\citep{yuan2025transformers} & $90.5$ & $98.8$ & $5.0$ & $56.0$ & $68.0$ & $53.5$ \\

\midrule

\rowcolor{green!20}
\plant (Ours) & \textbf{97.4} & 99.5 & \textbf{23.0} & \textbf{59.2} & \textbf{72.1} & \textbf{56.9} \\

\rowcolor{green!10}
&
\gainonly{+0.2}{0.08}{0.31} & \diponly{-0.1}{-0.27}{0.06} & \gainonly{+0.7}{0.48}{0.95} &
\gainonly{+0.3}{0.12}{0.44} & \gainonly{+1.4}{1.01}{1.88} & \gainonly{+1.1}{0.72}{1.55} \\

\rowcolor{green!10}
&
\cionly{0.08}{0.31} & \cionly{-0.27}{0.06} & \cionly{0.48}{0.95} &
\cionly{0.12}{0.44} & \cionly{1.01}{1.88} & \cionly{0.72}{1.55} \\
\bottomrule
\end{tabular}
}
\end{table*}
\textbf{RQ2: Does \plant \footnotemark\ outperform SOTA models on ICD-10 code classification?}
\label{subsection:main_results}
Table~\ref{tab:results_mimicfour} compares \plant with SOTA models on \mimicfour.
Performance comparison across \mimicfull and \mimicfifty is provided in Table~\ref{tab:results_full_and_top50} in Appendix~\ref{sec:appendix_results_detailed}.
On \mimicfour, which exhibits a more skewed label distribution (see Table~\ref{tab:data-stat}), \plant demonstrates gains of up to \textbf{+1.4}, including a \textbf{+0.7} (95\% CI: \textbf{0.5--1.0}) gain in \textsf{F1} (Macro) and a \textbf{+1.4} (95\% CI: \textbf{1.0--1.9}) improvement in \textsf{Precision@8} over SOTA baselines. \plant's larger gains on macro- than micro-averaged \textsf{AUC} (both datasets) and \textsf{F1} (\mimicfour) highlight its efficacy in addressing label imbalance.

\footnotetext{In this setting the base LLM \(\mathcal{M}_{\mathsf{base}}\) for \plant was \mistral.}

\begingroup
\begin{table*}[t]
\caption{\label{tab:results_xmtc}
    \textbf{\plant performs strongly across domains}---legal topic classification \& tag prediction on the \emph{full} \eurlex and \wikiten datasets (all labels), with a DistilBERT backbone for \plant (App.~\ref{app:training_details}).
    Shaded rows give \plant's gain over the strongest baseline per metric, with 95\% CIs.
    \plant is best on five of six metrics; on \wikiten \textsf{P@5}, XRR is 0.01 higher.
}
\centering
\resizebox{0.9\textwidth}{!}{
\begin{tabular}{l lll lll}
\toprule
\multirow{2}{*}{Model}
& \multicolumn{3}{c}{\makecell[c]{Legal Topic\\Classification (\textbf{\eurlex})}}
& \multicolumn{3}{c}{\makecell[c]{Content\\Recommendation (\textbf{\wikiten})}} \\
\cmidrule(lr){2-4} \cmidrule(lr){5-7}
 & $\mathsf{P@1}$ & $\mathsf{P@3}$ & $\mathsf{P@5}$ & $\mathsf{P@1}$ & $\mathsf{P@3}$ & $\mathsf{P@5}$ \\
\midrule
PECOS~\citep{yu2022pecos} & $88.41$ & $75.97$ & $63.18$ & $88.69$ & $80.17$ & $70.91$ \\
ICXML~\citep{zhu2023icxml} & $87.10$ & $74.32$ & $62.70$ & $87.61$ & $79.11$ & $69.78$ \\
XRR~\citep{xiong2023xrr} & $87.96$ & $78.88$ & $68.52$ & $89.54$ & $85.38$ & $\mathbf{81.34}$ \\
X-Transformer + RDE~\citep{shi2024residual} & $84.60$ & $72.61$ & $61.35$ & $86.15$ & $76.99$ & $68.75$ \\
MatchXML~\citep{ye2024matchxml} & $88.12$ & $75.00$ & $62.22$ & $89.30$ & $80.45$ & $70.89$ \\
DE~\citep{guptadual} & $87.60$ & $74.39$ & $67.80$ & $88.21$ & $80.29$ & $69.91$ \\
\makecell[l]{InceptionXML~\citep{kharbanda2023inceptionxml} + \\ GANDALF~\citep{kharbanda2024labelcor}} & $86.98$ & $75.89$ & $68.78$ & $88.76$ & $80.32$ & $69.89$ \\
CG~\citep{chai2024compositional} & $87.82$ & $76.71$ & $68.42$ & $87.29$ & $79.81$ & $68.45$ \\
ELMO~\citep{zhang2025elmo} & $87.95$ & $76.48$ & $65.91$ & $88.76$ & $83.94$ & $77.45$ \\
ViXML~\citep{ortego2026large} & $88.74$ & $77.12$ & $66.85$ & $89.28$ & $84.67$ & $78.92$ \\
\midrule

\rowcolor{green!20}
\plant (Ours)
& \textbf{90.61} & \textbf{81.35} & \textbf{70.24}
& \textbf{90.91} & \textbf{87.61} & 81.33 \\

\rowcolor{green!10}
&
\gainonly{+1.87}{1.14}{2.92} & \gainonly{+2.47}{1.73}{3.49} & \gainonly{+1.46}{0.94}{2.11}
& \gainonly{+1.37}{0.76}{1.93} & \gainonly{+2.23}{1.60}{2.84} & \diponly{-0.01}{-0.27}{0.19} \\

\rowcolor{green!10}
&
\cionly{1.14}{2.92} & \cionly{1.73}{3.49} & \cionly{0.94}{2.11}
& \cionly{0.76}{1.93} & \cionly{1.60}{2.84} & \cionly{-0.27}{0.19} \\
\bottomrule
\end{tabular}
}
\end{table*}
\endgroup

\textbf{RQ3: Why \plant\ Is Superior to Few-Shot SOTA Models?}
\label{subsection:fig2vsfig3}
In the per-code results of \citet{edin2023automated} (Figure~\ref{fig:rare_codes_zoom}, left), codes seen about 10 times in training have near-zero \macrof.
We therefore zoom in on \mimicfew, whose 685 codes each appear 1--5 times in training. There, per-code \macrof\ averages \textbf{0.663} for \mistral+\plant versus 0.054 for \textsc{Co-Relation} \citep{luo2024corelation}, and \textbf{54.8\%} of these codes reach \macrof{} $>0.7$ with \plant, versus 0.6\% for \textsc{Co-Relation}. Integrating \plant with a base LLM not only surpasses specialized few-shot methods (Table~\ref{tab:results_mimic3_few_and_rare}), but also substantially improves rare-label modeling (case studies in App.~\ref{sec:appendix_attention_analysis}). For example, on \mimicfour the tokens that \plant's attention ranks highest for \texttt{C83.18} (mantle cell lymphoma, multiple lymph-node regions) include \emph{mant} (mantle), \emph{bend} and \emph{rit} (plausibly bendamustine and rituximab, standard treatments), and \emph{ki} (plausibly Ki-67).

\textbf{RQ4: How generalizable is \plant \footnotemark\ to other  imbalanced classification tasks?}
Table~\ref{tab:results_xmtc} evaluates \plant on two diverse tasks: legal topic classification (\eurlex) and content recommendation (\wikiten), both characterized by extreme label spaces and imbalanced distributions. Against the strongest baseline for each metric, \plant gains \textbf{+1.5--2.5} on \eurlex across \textsf{P@1}, \textsf{P@3}, and \textsf{P@5},
including
a \textbf{+2.5} (95\% CI: \textbf{1.7--3.5}) gain in \textsf{P@3}.
On \wikiten, \plant gains \textbf{+1.4} in \textsf{P@1} and \textbf{+2.2} (95\% CI: \textbf{1.6--2.8}) in \textsf{P@3}, and is tied with XRR on \textsf{P@5} ($-0.01$; CI includes 0).

\footnotetext{In this setting the backbone \(\mathcal{M}_{\mathsf{base}}\) for \plant was DistilBERT (66M parameters), matching the scale of the XMTC baselines.}

\begin{table*}[t]
\caption{\label{tab:w1_w2}
\textbf{Left (RQ5):} same LLM fine-tuned without (FT) and with \plant on \eurlex/\wikiten subsets (600 documents, 60 for testing; 150 most frequent labels).
\textbf{Right (RQ6):} MIMIC-IV-few precision by training-frequency bucket: zero-shot (0 occurrences), few-shot (1--7), head ($\geq$8), all labels. TF-0/TF-2: training-free extract-then-link with zero/two in-context examples.
All values $\times100$; single runs; \textbf{bold}: best per backbone and column.
}
\centering
\footnotesize
\setlength{\tabcolsep}{2.6pt}
\begin{minipage}[t]{0.535\textwidth}
\vspace{0pt}
\centering
\resizebox{\linewidth}{!}{%
\begin{tabular}{ll ccc ccc}
\toprule
 & & \multicolumn{3}{c}{EUR-Lex subset} & \multicolumn{3}{c}{Wiki10 subset} \\
\cmidrule(lr){3-5} \cmidrule(lr){6-8}
Backbone & & P@25 & R@25 & PSP@25 & P@25 & R@25 & PSP@25 \\
\midrule
\multirow{2}{*}{Mistral} & FT & 3.2 & 31.6 & 26.2 & 14.6 & 57.5 & 52.3 \\
 & \plant & \textbf{4.7} & \textbf{48.5} & \textbf{42.5} & \textbf{19.1} & \textbf{77.2} & \textbf{67.5} \\
\midrule
\multirow{2}{*}{Llama-3.1} & FT & 3.9 & 38.1 & 31.9 & 15.0 & 62.5 & 46.5 \\
 & \plant & \textbf{5.4} & \textbf{54.0} & \textbf{49.5} & \textbf{19.4} & \textbf{78.6} & \textbf{68.6} \\
\midrule
\multirow{2}{*}{Phi-3-small} & FT & 3.1 & 29.1 & 24.6 & 3.9 & 12.7 & 16.7 \\
 & \plant & \textbf{4.3} & \textbf{39.0} & \textbf{39.7} & \textbf{15.5} & \textbf{64.7} & \textbf{49.5} \\
\bottomrule
\end{tabular}}
\end{minipage}\hfill
\begin{minipage}[t]{0.44\textwidth}
\vspace{0pt}
\centering
\resizebox{\linewidth}{!}{%
\begin{tabular}{ll cccc}
\toprule
 & & Zero & Few & Head & All \\
Backbone & Method & (P@1) & (P@5) & (P@8) & (P@15) \\
\midrule
\multirow{4}{*}{Mistral} & FT & 0.6 & 5.5 & 27.7 & 22.2 \\
 & TF-0 & 27.5 & 14.9 & 29.6 & 22.4 \\
 & TF-2 & 25.8 & \textbf{15.5} & 29.3 & 22.7 \\
 & \plant & \textbf{30.9} & 11.6 & \textbf{30.8} & \textbf{24.6} \\
\midrule
\multirow{4}{*}{Llama-3.1} & FT & 0.8 & 0.9 & 17.4 & 9.9 \\
 & TF-0 & 22.8 & \textbf{13.9} & 27.6 & 21.0 \\
 & TF-2 & 23.4 & \textbf{13.9} & 28.4 & 22.3 \\
 & \plant & \textbf{28.4} & 13.4 & \textbf{34.2} & \textbf{27.0} \\
\midrule
\multirow{2}{*}{Phi-3-small} & FT & 1.1 & \textbf{6.7} & 17.2 & 10.8 \\
 & \plant & \textbf{11.2} & 5.8 & \textbf{29.0} & \textbf{21.4} \\
\bottomrule
\end{tabular}}
\end{minipage}
\end{table*}

\textbf{RQ5: Does \plant improve fine-tuned LLMs outside the medical domain?}
\label{subsection:w1_controlled}
To isolate \plant's contribution to a fine-tuned LLM, we fine-tuned \mistral, \llama, and \phiThreeSmall with and without \plant, using the same data split, backbone, and training configuration, on deterministic subsets of \eurlex and \wikiten (Table~\ref{tab:w1_w2}, left).
For compute reasons, each subset has 600 documents (540 train / 60 test) and is restricted to the 150 most frequent labels; label frequencies remain skewed (94/150 \eurlex and 42/150 \wikiten labels have fewer than 10 training examples).
\plant improves \textsf{P@25}, \textsf{R@25}, and \textsf{PSP@25} for all three backbones on both subsets.
These are single runs on small test sets: directional evidence that complements, but does not replace, a full-label-space comparison, which we did not complete.
The comparison is also sensitive to the training budget: \wikiten needed 3 epochs per stage (\eurlex used 1), and with 1 epoch per stage \plant trailed the \mistral baseline on ranking metrics (App.~\ref{app:w1_details}).
\phiThreeSmall's large relative gain on \wikiten reflects an unusually weak baseline rather than an unusually strong \plant result.

\textbf{RQ6: How does \plant compare with zero- and few-shot LLM baselines?}
\label{subsection:w2_fewshot}
We keep two notions of zero/few-shot separate: a label's \emph{frequency bucket} (how often it occurs in training) and a training-free LLM's \emph{prompting regime} (how many in-context examples it receives).
We built MIMIC-IV-few, an 800-note subset of \mimicfour (516/117/167 train/val/test notes) in which each note keeps its complete ICD-10 annotation (3,343 labels; median 17 per note); notes were selected through codes with 3--7 training occurrences (App.~\ref{app:w2_details}).
We compare \plant with the same backbone fine-tuned without \plant, and with a training-free extract-then-link pipeline: an off-the-shelf instruction-tuned LLM lists diagnoses and procedures without seeing the label set, given zero or two fixed in-context examples, and SapBERT~\citep{liu2021self} links the extracted mentions to the 3,343 code descriptions.
\plant improves over the fine-tuned baseline on zero-shot, head, and overall labels for all three backbones, and on few-shot labels for \mistral and \llama but not for \phiThreeSmall.
The training-free pipeline (run for \mistral and \llama only) is strong on rarely seen labels and exceeds \plant's few-shot-label \textsf{P@5} for both, whereas \plant is best on unseen labels, head labels, and overall ranking, which extraction alone does not match. \deepseek was not evaluated in either comparison because of its cost.

\textbf{RQ7: How effective is \plant on zero-shot transfer to unseen rare labels?}
\label{subsection:freq_binned_transfer}
Across 4 LLMs,
\plant trained on the \emph{full} dataset achieves the strongest Rare-F1 (13.2--16.8\%).
Figure~\ref{fig:cosine_hisograms_and_freq_bin_transfer_bars} (right) reports Rare-F1 when the model is trained exclusively on documents containing only common labels and evaluated on held-out rare labels.  The \textbf{second and third batches of bars} in the figure correspond to models trained only on the common-label subset: here, \plant still retains substantial performance (6.9--8.1\%), whereas the same models with \textbf{random atten-init} collapse to 0.5--1.1\% Rare-F1.
This gap (up to $+7.5$\,pp, for \llama) shows that \plant's initialization enables zero-shot transfer to rare labels that receive no supervision during training.
This is transfer to unseen labels \emph{within} a dataset; we do not transfer planted attention across datasets with different label spaces, since MIG and the attention initialization are computed on each target dataset's training split (Section~\ref{sec:limitations}).

\section{Ablation Analysis}
\label{sec:ana}
\textbf{\textsc{Ablating components in \plant (Table~\ref{tab:plant_2stage_training_ablation_mimiciv})}}
\label{sec:ablation_study}
To assess the contribution of individual components in \plant's two-stage training pipeline (Section~\ref{subsec:training}), we run ablations (1)--(7) and analysis (8) with \mistral on \mimicfull and \mimicfour.
\textbf{Each ablation configuration is compared against the full \plant setup} (bottom row of Table~\ref{tab:plant_2stage_training_ablation_mimiciv}), intentionally isolating the effect of a component.
We evaluate using (1)~\macroAUC for per-label classification, (2)~\macrof for rare-label accuracy, and (3)~\patfifteen for top-$k$ prediction quality.
For each ablation, we report the average decrease relative to the full \plant setup (\avdip{x}).

    \textbf{(1)+(2) \textbf{Benefit of Stage~1 attention initialization.}}\label{sec:stage1_benefit}
    End-to-end training from scratch without Stage~1 degrades performance: using \textsf{BCE} loss yields avg dips (\macroAUC/\macrof/\patfifteen) of \avdip{6.2}/\avdip{4.8}/\avdip{5.3}, while \textsf{Focal{+}HNM} (\(\gamma=2,\epsilon=0.1,m=1000\)) reduces the dips to \avdip{4.8}/\avdip{3.4}/\avdip{3.8}.
    Focal+HNM helps, but pretrained attention helps much more.

    \textbf{(3) \textbf{Removing focal loss and HNM in Stage~2.}}
   Training with vanilla BCE after Stage~1 (\(\gamma = 0\), no label smoothing, no HNM), yields
    changes of  (\macroAUC/\macrof/\patfifteen): \avdip{2.5}/\avdip{2.0}/\avdip{2.5};
    this still beats both single-stage variants, so \emph{Stage~1 alone provides meaningful gains}.

    \textbf{(4)+(5) Effects of label smoothing and hard negative mining in Stage~2.}
    Removing label smoothing (\(\epsilon{=}0\)) causes mild but consistent drops:
    \avdip{0.3}/\avdip{1.1}/\avdip{1.2} in \macroAUC/\macrof/\patfifteen.
    Removing hard negative mining (computing loss over all labels instead of top-$m$ negatives) leads to larger degradations:
    \avdip{0.6}/\avdip{1.9}/\avdip{2.8}.
    HNM thus matters more than label smoothing, by focusing learning on informative negatives.

    \textbf{(6) \textbf{Importance of MIG vs.\ naive frequency in Stage~1.}}
    Replacing MIG relevance scores with normalized token frequency per label yields (\macroAUC/\macrof/\patfifteen): \avdip{1.0}/\avdip{2.4}/\avdip{3.3}: the coarser signal harms rare-label and top-$k$ accuracy.

    \textbf{(7) \textbf{Importance of ranking objective in Stage~1.\;}}
    Replacing Stage~1 pairwise ranking loss (Eq.~\ref{eq:ranking_loss}) with MSE, yields
    \emph{avg dips} (\macroAUC/\macrof/\patfifteen): \avdip{1.5}/\avdip{2.9}/\avdip{3.5}: the ranking signal is central to aligning attention with MIG relevance.

    \textbf{(8) \textbf{Effect of attention initialization quality of Stage~1.}}
    We vary the number of Stage~1 epochs (1–10) before switching to Stage~2, and measure attention initialization quality at the end of Stage~1 via $\mathsf{nDCG}@k$ (computed against MIG relevance as ground truth), alongside final \macrof\ and \patfifteen\ after Stage~2 (Figure~\ref{fig:effect_plant_initialization_plus_plant_vary_trn_size}, \textit{left}).
    Compared to 10 Stage~1 epochs, training for only 1 epoch yields
    \emph{avg dips} ($\mathsf{nDCG}@k$/\macrof/\patfifteen): \avdip{0.24}/\avdip{2.4}/\avdip{2.6}.
    As Stage~1 training length increases, $\mathsf{nDCG}@k$ steadily improves (e.g., 0.68$\to$0.94 on \mimicfour), and final \macrof\ and \patfifteen\ rise with it (Figure~\ref{fig:effect_plant_initialization_plus_plant_vary_trn_size}, left; App.~\ref{sec:appendix_results_detailed}, Fig.~\ref{fig:effect_plant_initialization}).

\begingroup
\begin{figure*}[t]
    \centering
    \begin{subfigure}[t]{0.49\textwidth}
        \centering
        \raisebox{0mm}{\includegraphics[height=3.9cm, keepaspectratio]{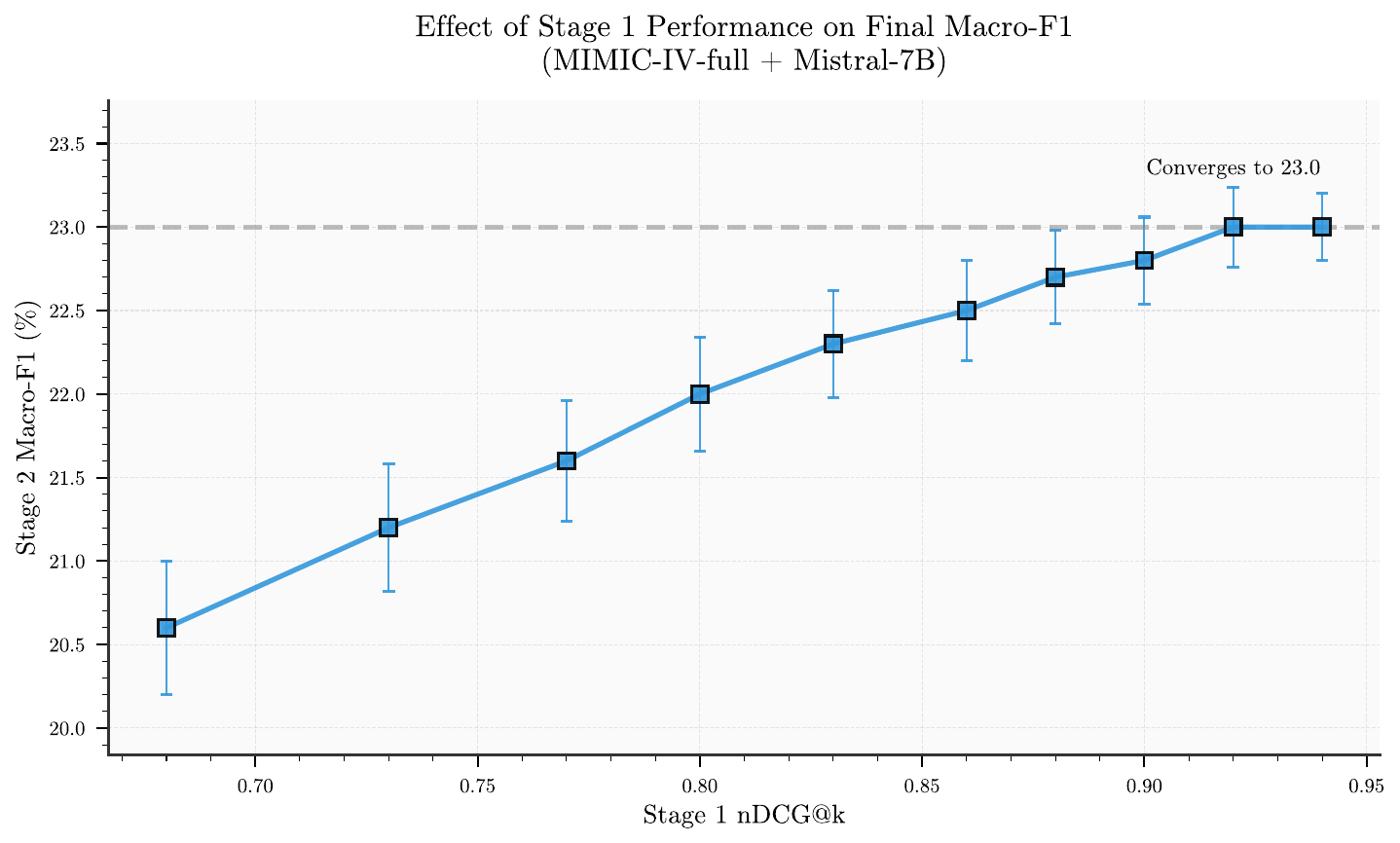}}
    \end{subfigure}
    \hfill
    \begin{subfigure}[t]{0.49\textwidth}
        \centering
        \raisebox{-3mm}{\includegraphics[height=3.9cm, keepaspectratio]{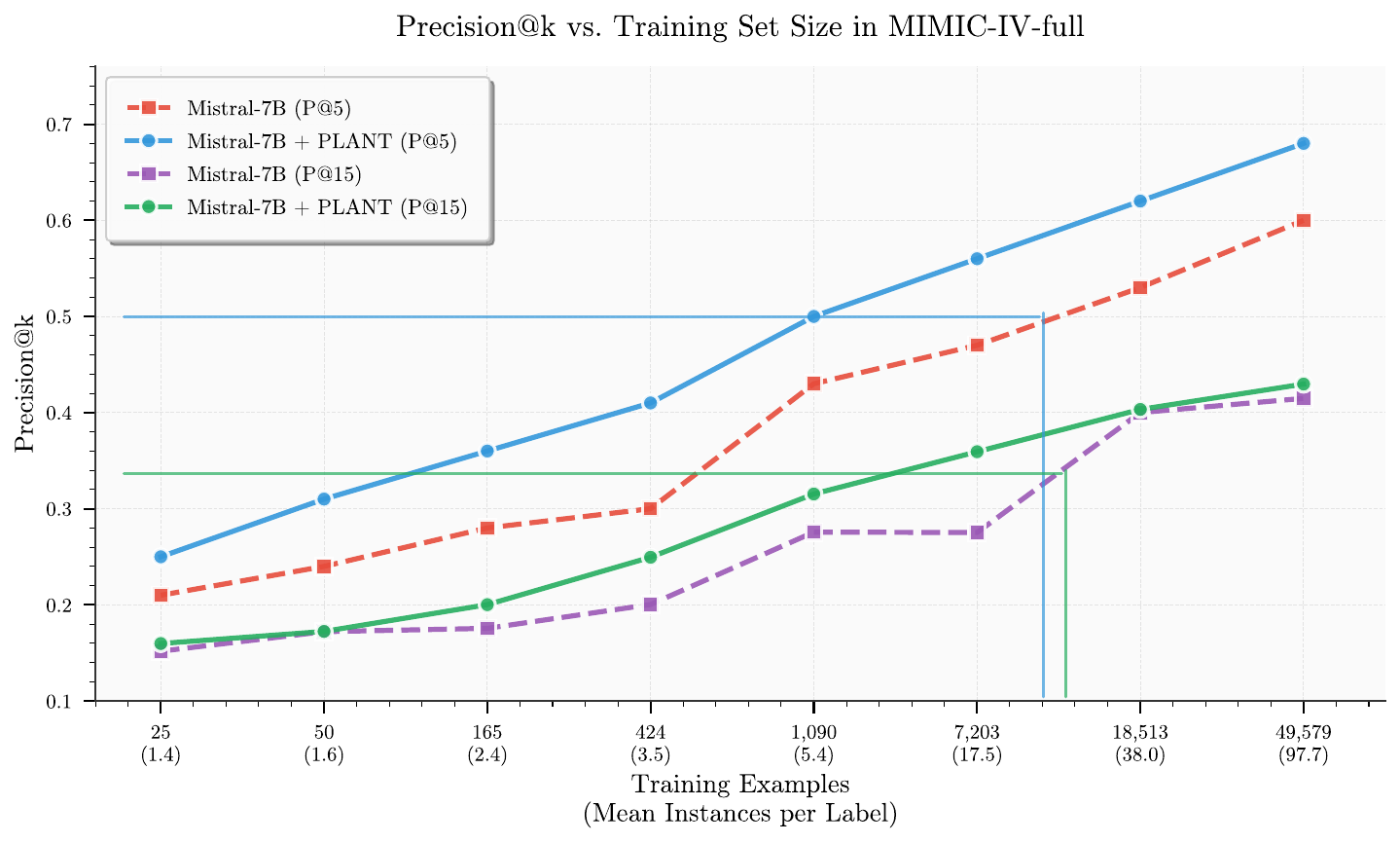}}
    \end{subfigure}
    \caption{\textit{(Left)} On \mimicfour{} w/ \mistral, better Stage~1 $\mathsf{nDCG}@k$ (attn. init. quality) leads to higher Stage~2 (downstream) \macrof.
Extended results (\mimicfull, \mimicfour; \macrof, \patfifteen) appear in
App.~\ref{sec:appendix_results_detailed}, Fig.~\ref{fig:effect_plant_initialization}. \textit{(Right)} \plant boosts \mistral on \mimicfour across training sizes:
solid lines (\mistral+\plant) beat dashed baselines on \patfive/\patfifteen, with largest gains in low-data regimes.
Paired \mimicfull+\mimicfour results are in App.~\ref{sec:appendix_results_detailed}, Fig.~\ref{fig:plant_vary_trn_size}.
}
    \label{fig:effect_plant_initialization_plus_plant_vary_trn_size}
\end{figure*}
\endgroup

\textbf{Key Takeaways (1--8):} \ul{Stage~1 attention pretraining is the single most impactful component of \plant}: removing it costs \avdiprange{4.8}{6.2} in \macroAUC, \avdiprange{3.4}{4.8} in \macrof, and \avdiprange{3.8}{5.3} in \patfifteen, and within Stage~1 both MIG relevance and the ranking objective matter.

\begingroup
\begin{table*}[t]
\caption{
Ablation on \mimicfour with base \mistral.
The single-stage rows use a different configuration from the Table~\ref{tab:sota_llm_vs_plant_mimiciv} baselines, so their scores differ.
Full results (\mimicfull and \mimicfour) appear in
Table~\ref{tab:plant_2stage_training_ablation_full}, App.~\ref{sec:appendix_results_detailed}.
}
\label{tab:plant_2stage_training_ablation_mimiciv}
\centering
\small
\resizebox{\columnwidth}{!}{
\begin{tabular}{lccc}
\toprule
\textbf{Ablation Config} & \macroAUC & \macrof & \patfifteen \\
\midrule
\stageOne Single-Stage BCE
    & 91.0\dipci{-6.4}{-7.2}{-5.4}
    & 18.0\dipci{-5.0}{-5.9}{-3.8}
    & 52.0\dipci{-4.9}{-5.7}{-3.9} \\
\stageOne Single-Stage Focal Loss
    & 92.5\dipci{-4.9}{-5.8}{-4.0}
    & 19.5\dipci{-3.5}{-4.2}{-2.7}
    & 53.5\dipci{-3.4}{-4.0}{-2.7} \\
\stageOne \plant w/ Vanilla BCE
    & 95.0\dipci{-2.4}{-3.0}{-1.7}
    & 21.0\dipci{-2.0}{-2.6}{-1.4}
    & 54.8\dipci{-2.1}{-2.7}{-1.5} \\
\stageTwo \plant w/o Label Smoothing
    & 97.0\dipci{-0.4}{-0.7}{-0.2}
    & 21.8\dipci{-1.2}{-1.8}{-0.7}
    & 55.8\dipci{-1.1}{-1.7}{-0.6} \\
\stageTwo \plant w/o Hard Neg Mining
    & 96.8\dipci{-0.6}{-1.0}{-0.3}
    & 21.0\dipci{-2.0}{-2.7}{-1.3}
    & 54.5\dipci{-2.4}{-3.1}{-1.7} \\
\stageOne \plant w/ Term Frequency
    & 96.5\dipci{-0.9}{-1.4}{-0.5}
    & 20.5\dipci{-2.5}{-3.2}{-1.8}
    & 54.0\dipci{-2.9}{-3.6}{-2.1} \\
\stageOne \plant w/ MSE
    & 96.0\dipci{-1.4}{-2.1}{-0.8}
    & 20.0\dipci{-3.0}{-3.9}{-2.1}
    & 53.8\dipci{-3.1}{-3.9}{-2.3} \\
\rowcolor{green!20}
\plant (full setup)
    & \textbf{97.4} & \textbf{23.0} & \textbf{56.9} \\
\bottomrule
\end{tabular}
}
\end{table*}
\endgroup

\textbf{\textsc{\plant under Varying Training Size}}
\label{subsec:ablation_few_shot}
Can \plant's pretrained attention reduce the labeled data needed for competitive performance?
We compare \plant's two-stage training (Section~\ref{subsec:training}) with single-stage end-to-end training using the same architecture (Section~\ref{subsection:architecture}) and base LLM \mistral,
on \mimicfull and \mimicfour under varying training sizes
(Figure~\ref{fig:effect_plant_initialization_plus_plant_vary_trn_size}, right).
Both methods are trained on different fractions of balanced training splits, with fixed test sets, up to 5 epochs, and evaluated on \patfive and \patfifteen.
The sole difference is that \plant starts from Stage~1 pretrained attention, while the baseline learns attention from scratch.
\phead{Takeaway:} As shown in Figure~\ref{fig:effect_plant_initialization_plus_plant_vary_trn_size} (right), \plant consistently matches or exceeds end-to-end \mistral across all training sizes, often with several times fewer labels.
On \mimicfour, \plant reaches \patfive{=}0.50 with \textbf{1,090} training examples, a level the baseline reaches only between 7,203 and 18,513 (Figure~\ref{fig:effect_plant_initialization_plus_plant_vary_trn_size}, right).
On \mimicfull, it reaches \patfive{=}0.47 with \textbf{136}, which the baseline reaches only between 750 and 1,756 (Fig.~\ref{fig:plant_vary_trn_size}).

\textbf{\textsc{Random Attention Initialization Causes Rare-Label Failures}}
In a controlled analysis on \mimicfour with \llama (App.~\ref{app:causal_ablation}, Table~\ref{tab:init_causal}), random attention initialization trained on common labels only yields the highest correlation between per-label F1 and log label frequency (0.78, vs.\ 0.45 for \plant with the same training) and the lowest alignment between learned attention $\mathbf{S}_l$ and MIG relevance for rare labels (cosine 0.08, vs.\ 0.52): \plant's initialization mitigates both frequency bias and attention misalignment.

\section{Related Work}
\noindent

Attention has long been used to capture label–text interactions.
Label-wise attention has been central to XMTC and ICD coding since AttentionXML and CAML \citep{you2019attentionxml, mullenbach-etal-2018-explainable}.
Transformer-based models introduced multi-resolution self-attention for large label spaces \citep{zhang2021fast, kharbanda2022cascadexml}, while multi-head attention across text granularities improved weak supervision \citep{kargupta2023megclass}.
Other work leverages contrastive or knowledge-enhanced attention \citep{lu2023towards, li2023towards}.
Dynamic pipelines filtered candidate labels using structured signals like diagnoses, procedures, and medications, relying on attention to prioritize relevant labels \citep{wang2024multi, wang2024auxiliary}.
Other studies show that label-guided, dictionary, or bi-attention mechanisms improve alignment between labels and text \citep{wang2023benchmark, wu2024dila, wang2024label}.
Meta-learning and label tree structures further advance attention-driven few-shot generalization \citep{teng2024few, wang2024icdxml}.
Recent work includes attention-based co-ranking \citep{yan2025labelcorank},
contrastive dual-attention for rare labels \citep{huang2025contrastive},
and knowledge-integrated attention for medical coding \citep{zhang2025general}.
Prior work has also initialized label embeddings for label-wise attention, e.g., from code descriptions \citep{dong2021explainable}.
To our knowledge, \plant is the first to initialize the label-attention module itself with a learning-to-rank objective over MIG relevance.

LLMs are increasingly used for XMTC~\citep{ma2025large, tabatabaei2025can, asensio2025admission, yuan2025transformers, nerella2024transformers}. Yet large LLMs in zero-shot mode can underperform smaller fine-tuned models~\citep{boyle2023automated, zhang2025general}. Heavy finetuning, in turn, raises concerns about compute cost and overfitting~\citep{huang2022plm, michalopoulos2022icdbigbird, ng2023modelling, kang2023automatic}. \citet{sakai2025large} further show that such finetuning often fails to improve rare-label performance in high-dimensional, skewed label spaces.

\section{Limitations}
\label{sec:limitations}
(1)~\emph{Supervision.} Stage~1 estimates MIG from token--label co-occurrences in labeled training data, so \plant cannot pretrain attention in a fully unsupervised setting, and very sparse or noisy annotations can make MIG estimates unreliable.
(2)~\emph{Scale and scope.} We did not evaluate larger XMTC benchmarks such as Wiki-500K or Amazon-3M. Outside the medical domain, the full-dataset results (Table~\ref{tab:results_xmtc}) use a DistilBERT backbone, and the controlled comparisons with fine-tuned LLMs (RQ5) are single runs on 600-document, 150-label subsets; matched full-label-space comparisons remain future work.
(3)~\emph{Rare labels.} A training-free extract-then-link LLM pipeline exceeds \plant on few-shot-label \textsf{P@5} in MIMIC-IV-few, \plant does not improve \phiThreeSmall's few-shot bucket, and these zero-/few-shot comparisons omit \deepseek and are single runs (RQ6).
(4)~\emph{Mechanism.} Our account of negation (Section~\ref{sec:approach}) relies on contextualized hidden states and has not been validated by a controlled negation analysis. Planted attention is dataset-specific; transfer across datasets with different label spaces is untested.
(5)~\emph{Cost.} Stage~1 adds 12--28\% training and ${\sim}$10\% inference time (App.~\ref{app:efficiency_benchmarks}). See App.~\ref{app:limitations} for details.

\section{Conclusion}
This work proposed \plant, a plug-and-play strategy that pretrains the label-attention module as a \ltr model with mutual information and then leverages it in end-to-end training, turning attention into a pretrainable component.
\plant is architecture-agnostic and boosts performance across tasks and backbones: smaller LLMs with \plant outperform much larger models used alone, it matches baselines trained on several times more data, and it substantially improves rare-label prediction over fine-tuned baselines, although a training-free extraction baseline remains competitive on rarely seen labels.

%% file: sections/appendix.tex
\section{Training Details}
\label{app:training_details}
\phead{LLM Backbone}  
We use the following pretrained instruction-tuned LLMs as base models \(\mathcal{M}_{\mathsf{base}}\) in our experiments, all publicly available on the Hugging Face Model Hub and compatible with the Transformers library:
(1) \textbf{Mistral-7B-Instruct-v0.3} (7B, Mistral AI): \url{https://huggingface.co/mistralai/Mistral-7B-Instruct-v0.3}, (2) \textbf{Llama-3.1-8B-Instruct} (8B, Meta AI): \url{https://huggingface.co/meta-llama/Llama-3.1-8B-Instruct}, (3) \textbf{DeepSeek-V3} (671B total, 37B activated per token; mixture-of-experts, DeepSeek-AI): \url{https://huggingface.co/deepseek-ai/DeepSeek-V3}, and (4) \textbf{Phi-3-mini-4k-instruct} (3.8B, Microsoft): \url{https://huggingface.co/microsoft/Phi-3-mini-4k-instruct}.
These serve as the LLM backbones for fine-tuning; the main text refers to them as \mistral, \llama, \deepseek, and \phiThree. The legend of Figure~\ref{fig:cosine_hisograms_and_freq_bin_transfer_bars} (right) uses the shorter labels LLaMA3-8B and Phi-3 for the same \llama and \phiThree checkpoints. %

The controlled comparisons in RQ5 and RQ6 (App.~\ref{app:w1_w2_details}) run on a single NVIDIA H100 80GB GPU and use the same \mistral and \llama checkpoints; in place of \phiThree they use \textbf{Phi-3-small-8k-instruct} (7B, Microsoft; \phiThreeSmall in the text): \url{https://huggingface.co/microsoft/Phi-3-small-8k-instruct}.

For the extreme multi-label text classification (XMTC) results reported in Table~\ref{tab:results_xmtc}, \plant was additionally adapted to a compact \textbf{DistilBERT} encoder backbone (66M parameters) to ensure a fair comparison with the listed baselines (e.g., XRR~\citep{xiong2023xrr}, MatchXML~\citep{ye2024matchxml}, and InceptionXML~\citep{kharbanda2023inceptionxml}), which similarly employ encoder models of comparable scale. The DistilBERT model is initialized from \texttt{distilbert-base-uncased} and is publicly accessible at \url{https://huggingface.co/distilbert/distilbert-base-uncased}.

\phead{Quantization \& LoRA Adaptation}
Starting with a pretrained model \(\mathcal{M}_{\mathsf{base}}\), such as \mistral, \llama, or \phiThree, we apply Parameter-Efficient Fine-Tuning (PEFT) using Low-Rank Adaptation (LoRA)~\cite{hu2022lora, dettmers2023qlora, liu2024dora}. 

To enable memory-efficient fine-tuning on resource-constrained hardware, we first quantize \(\mathcal{M}_{\mathsf{base}}\) to 4-bit precision using the \texttt{NormalFloat4} (NF4) format with double quantization \citep{dettmers2023qlora}, yielding \(\mathcal{M}_{\mathsf{quant}}\).
NF4 is a non-uniform, quantile-based format: each block of weights is scaled by its absolute maximum and mapped to the nearest of 16 levels placed at quantiles of a standard normal distribution; double quantization additionally quantizes the per-block scaling constants.
The quantized layers compute in \texttt{bfloat16}; training otherwise uses FP16 mixed precision (below).

We then apply LoRA to a subset of the attention projection layers (query, key, value, and output), introducing trainable low-rank matrices:
\[
\Delta \bs{W} = \bs{A} \bs{B}, \quad \text{with } \bs{A} \in \mathbb{R}^{d \times r}, \bs{B} \in \mathbb{R}^{r \times d},
\]
using rank \(r = 16\), scaling factor \(\alpha = 32\), and dropout \(p = 0.05\). The adapted model becomes:
\[
\mathcal{M}_{\mathsf{adapt}} = \mathcal{M}_{\mathsf{quant}} + \alpha \cdot \Delta \bs{W}.
\]

\phead{Optimization and Training Regimen}
To address potential overwriting of Stage~1 attention signals during Stage~2 fine-tuning, we employ a gradual unfreezing strategy combined with discriminative learning rates, ensuring stable transfer of the MIG-seeded priors while allowing task-specific refinement. 
All experiments are conducted on 8$\times$A100--80GB GPUs using DeepSpeed ZeRO-3 offloading for memory efficiency, with a global batch size of 256 (gradient accumulation steps{=}4) and mixed-precision (FP16) training via Hugging Face Accelerate.

Stage~1 pretraining optimizes \(\mathcal{M}_{\text{adapt}}\), the multi-head attention module ($\mathsf{MultiHead}$) and label embeddings $\mathbf{E}$ via the ranking loss (Eq.~\ref{eq:ranking_loss}) for at most 10 epochs, using AdamW with a cosine learning-rate schedule (peak $\eta = 5\times10^{-4}$, 10\% warmup) and weight decay $\lambda = 0.01$. %

In Stage~2, we initialize from Stage~1 checkpoints and apply discriminative fine-tuning to preserve attention integrity: the attention module ($\mathsf{MultiHead}$, $\mathbf{E}$) starts frozen for the first 5 epochs (allowing downstream layers to adapt), followed by gradual unfreezing of the full model in three phases---attention last (epochs~6--10, $\eta = 1\times10^{-5}$), intermediate layers (epochs~11--15, $\eta = 5\times10^{-6}$), and all parameters (epochs~16--20, $\eta = 2\times10^{-6}$)---each with cosine decay and 5\% warmup. This layered schedule, inspired by progressive distillation in large-scale vision--language models~\citep{hou2018lifelong}, is paired with the AdamW optimizer~\cite{loshchilov2017decoupled} (weight decay 0.01) and gradient clipping (max-norm 1.0) for stability. We use a per-device batch size of~8 with 4-step gradient accumulation (effective per-device batch 32; global batch 256 across 8 GPUs), PyTorch’s \texttt{autocast} for FP16, and gradient checkpointing to manage memory. Stage~1 runs for at most 10 epochs and Stage~2 for at most 20 epochs, each with early stopping (patience{=}2) on validation $\mathsf{nDCG@k}$ (Stage~1) or macro-F1 (Stage~2); in practice Stage~1 stops at epochs 7--8 and Stage~2 at epochs 12--15, so the last unfreezing phase is usually not reached. Single-stage baselines use the same Stage~2 schedule and stopping rule. To ensure reproducibility, we fix random seeds across \texttt{random}, \texttt{numpy}, \texttt{torch}, and \texttt{torch.cuda}; experiments use distributed data-parallelism (DDP) where applicable, with metrics logged via Weights \& Biases.

\phead{Token Selection Sensitivity}
To test sensitivity to token selection in the ranking loss (Equation~\ref{eq:ranking_loss}), we vary the top-\(k\) token threshold with \(k \in \{500, 1000, 2000\}\).

\subsubsection*{Hyperparameters in Architecture (Section~\ref{subsection:architecture})}
\label{app:hypers}
The multi-head attention module \(\mathsf{MultiHead}\) (Equation~\ref{eq:multiheadattention}) uses \(k = 8\) attention heads. The adaptive average pooling layer (Equation~\ref{eq:avg_pooling}) produces an output size of \(p = 128\).

\subsubsection*{Focal Loss with Label Smoothing}
\label{app:focal_loss}
For completeness, we present the explicit formulation of the focal loss used in Stage~2 training (Section~\ref{sec:stage2_training}). 
The focal loss for an input \(\mathbf{x}_i\) is defined as:
\begin{equation}
\mathcal{L}_{\text{focal}}^{(i)}(\tilde{y}, \hat{y}, \theta) 
= -\frac{1}{|\mathcal{L}|} \sum_{l=1}^{|\mathcal{L}|} \biggl[ 
\tilde{y}_{il} \left(1 - \sigma(\hat{y}_{il})\right)^{\gamma} %
\log\left(\sigma(\hat{y}_{il})\right)
+ 
\left(1 - \tilde{y}_{il}\right) 
\left(\sigma(\hat{y}_{il})\right)^{\gamma} 
\log\left(1 - \sigma(\hat{y}_{il})\right) 
\biggr],
\label{eq:focal_loss}
\end{equation}
where \(\theta\) denotes all trainable model parameters, \(\sigma(\cdot)\) is the sigmoid function, \(\hat{y}_{il} \in \mathbb{R}\) is the predicted logit for label \(l\), \(\gamma = 2\) is the focusing parameter that emphasizes harder examples, and \(\tilde{y}_{il}\) is the smoothed label:
\(\tilde{y}_{il} = (1-\epsilon)^{y_{il}} \, (\epsilon)^{1-y_{il}}\)
\label{eq:label_smoothing}
with \(\epsilon = 0.1\) to prevent overconfidence in predictions.

\subsubsection*{Causal ablation details}
\label{app:causal_ablation}
To establish that random initialization of the label-specific attention module is the primary cause of poor rare-label performance, we run controlled ablations on \textsc{MIMIC-IV} ICD-10 using \llama. Labels are stratified by training-set frequency: \textbf{rare} ($<0.1\%$) and \textbf{common} ($>1\%$). We evaluate two matched setups:  
(1) \textbf{Random Attn Init (common only)}—Stage~1 skipped; attention weights $\mathbf{W}_{\mathsf{attn}}$ and label embeddings $\mathbf{E}$ remain Xavier-initialized;  
(2) \textbf{\plant (common only)}—full Stage~1 applied only to the common-label subset,
updating $  \bs{W}_{\mathsf{attn}}  $, $  \bs{E}  $, $  \mathbf{B}_a  $, and $  \mathbf{B}_m  $ exclusively for common labels. Rare-label parameters receive no gradient updates and keep their initial values. %
Both models are trained solely on common labels and evaluated on held-out rare labels.

All experiments use 5-fold cross-validation on MIMIC-IV-full (80/10/10 split). Backbone: \llama with QLoRA (rank 16, $\alpha=32$, 4-bit). Stage~2: 20 epochs, AdamW (lr $=1\text{e}^{-5}$), focal loss ($\gamma=2$), label smoothing ($\epsilon=0.1$), and hard-negative mining ($m=1000$). MIG top-$k=1000$ is computed on the corresponding training subset. Common-only training removes all rare-label instances. Metrics include micro-F1 per frequency bin and Pearson correlation between per-label F1 and log-frequency. Cosine similarity is averaged over 50 randomly sampled rare labels using full-corpus MIG as reference. All results use seed 42; paired t-tests show $p<0.001$ for Rare-F1 differences, with Cohen’s $d>1.4$.

Table~\ref{tab:init_causal} reports the resulting frequency bias and attention alignment (summarized in Section~\ref{sec:ana}).

\begingroup
\setlength{\textfloatsep}{0.5em}
\begin{table*}[ht]
\caption{Impact of \plant on rare-label performance in \mimicfour using \llama.}
\vspace{-5pt}
\label{tab:init_causal}
\centering
\small
\resizebox{\textwidth}{!}{
\begin{tabular}{lcccc}
\toprule
Variant (training data) & Macro-F1 (\%) & Rare-F1 (\%) & Corr(F1, log-freq) & Sim($\mathbf{S}_l$, MIG)-Rare \\
\midrule
1 Epoch Stage~1 \plant (full data)  & 20.5 $\pm$ 0.6 & 10.3 $\pm$ 0.9 & 0.55 $\pm$ 0.05 & 0.35 $\pm$ 0.07 \\
\plant (full data)       & 23.0 $\pm$ 0.3 & 16.8 $\pm$ 0.5 & \cellcolor{green!30}0.30 $\pm$ 0.03 & \cellcolor{green!30}0.78 $\pm$ 0.04 \\
\midrule
Random Attn Init (common only)    & 12.5 $\pm$ 1.0 &  0.5 $\pm$ 0.4 & \cellcolor{red!30}0.78 $\pm$ 0.03 & \cellcolor{red!30}0.08 $\pm$ 0.05 \\
\plant (common only)    & 18.0 $\pm$ 0.7 &  8.0 $\pm$ 0.6 & \cellcolor{green!30}0.45 $\pm$ 0.04 & \cellcolor{green!30}0.52 $\pm$ 0.07 \\
\bottomrule
\end{tabular}
}
\vspace{-10pt}

\end{table*}
\endgroup

\section{Precomputations}
\subsection{Mutual Information Gain in XMTC}
\label{app:mig}
In extreme multilabel classification (e.g., ICD coding on MIMIC-IV), MIG quantifies the informativeness of a token $t_j$ for predicting label $l$ presence, grounded in information theory as the KL divergence between the joint and product-of-marginals distributions. Formally:
\[
r_{l,j} = \sum_{(x, y) \in \{0,1\}^2} P(x, y) \log \left( \frac{P(x, y)}{P(x)\, P(y)} \right),
\]
where $x = \mathbb{I} [l \text{ present}]$ (marginal $P(x)$ = label frequency), $y = \mathbb{I}[t_j \text{ present}]$ (marginal $P(y)$ = token frequency), and $P(x,y)$ is the empirical joint from corpus co-occurrences. This measures bits of mutual information: how much $y$ reduces entropy in $x$, penalizing spurious correlations (e.g., high $P(y)$ but low $P(x|y) > P(x)$).

Probabilities are estimated via maximum-likelihood on the full training corpus (no subsampling), with Laplace smoothing ($\alpha=1$) for zero-count cells to avoid undefined logs. Scores $r_{l,j}$ are max-normalized per label to $[0,1]$ (dividing by $\max_j r_{l,j}$) for stability, then thresholded at top-$k$ (tested $k\in\{500,1000,2000\}$) to select tokens for Stage~1 ranking. Unlike raw co-occurrence (e.g., $P(y|x)$), MIG corrects for frequency bias: high-frequency tokens inflate joints but are downweighted if independent of $l$.

\textbf{Example.} Consider a toy corpus ($N=100$ docs): rare label A ($P(x)=0.05$, 5 docs), common B ($P(x)=0.50$, 50 docs); token ``fever'' ($P(y)=0.40$, co-occurs with B in 25 docs); ``rare\_disease'' ($P(y)=0.06$, co-occurs with A in 5 docs). Raw co-occurrence ranks ``fever'' higher for B (support=25 vs. 5), but MIG elevates ``rare\_disease'' for A due to stronger conditional dependence.

For ``fever'' w.r.t. B, the contingency table yields joints: $P(x=1,y=1)=0.25$, $P(x=1,y=0)=0.25$, $P(x=0,y=1)=0.15$, $P(x=0,y=0)=0.35$. MI computation (log base 2):
\begin{align*}
&\text{(1,1): } 0.25 \cdot \log_2(0.25 / (0.50 \cdot 0.40)) \approx 0.25 \cdot 0.322 = 0.0805, \\
&\text{(1,0): } 0.25 \cdot \log_2(0.25 / (0.50 \cdot 0.60)) \approx 0.25 \cdot (-0.263) = -0.066, \\
&\text{(0,1): } 0.15 \cdot \log_2(0.15 / (0.50 \cdot 0.40)) \approx 0.15 \cdot (-0.415) = -0.062, \\
&\text{(0,0): } 0.35 \cdot \log_2(0.35 / (0.50 \cdot 0.60)) \approx 0.35 \cdot 0.223 = 0.078, \\
&\text{Total MI} \approx 0.030 \text{ bits (weak dependence)}.
\end{align*}

For ``rare\_disease'' w.r.t. A: $P(x=1,y=1)=0.05$, $P(x=1,y=0)=0.00$, $P(x=0,y=1)=0.01$, $P(x=0,y=0)=0.94$. MI:
\begin{align*}
&\text{(1,1): } 0.05 \cdot \log_2(0.05 / (0.05 \cdot 0.06)) \approx 0.05 \cdot 4.06 = 0.203, \\
&\text{(0,1): } 0.01 \cdot \log_2(0.01 / (0.95 \cdot 0.06)) \approx 0.01 \cdot (-2.51) = -0.025, \\
&\text{(0,0): } 0.94 \cdot \log_2(0.94 / (0.95 \cdot 0.94)) \approx 0.94 \cdot 0.074 = 0.0696, \\
&\text{(1,0): } 0 \cdot \cdots = 0, \\
&\text{Total MI} \approx 0.248 \text{ bits (strong dependence)}.
\end{align*}
MIG ranks ``rare\_disease'' higher for A ($0.248 > 0.030$), capturing precision (\( P(A \mid \text{rare\_disease}) = 83\% \) vs. marginal 5\%) without volume bias. 

\section{Implementation Details}
\label{sec:appendix_implementation}

\subsection*{Datasets} 
\label{subsec:appendix_datasets}
We compare \plant to SOTA ICD coding models using the MIMIC-III \citep{johnson2016mimic} and MIMIC-IV \citep{johnson2023mimic} datasets, which include rich textual and structured records from ICU settings, primarily discharge summaries annotated with ICD-9 (MIMIC-III) and ICD-10 (MIMIC-IV) codes. MIMIC-III contains 52,723 discharge summaries with 8,929 unique ICD-9 codes, and MIMIC-IV includes 122,279 summaries with 7,942 ICD-10 codes. We follow established methodologies for patient ID-based splits and frequent code subsets. For few-shot learning, we evaluate \plant on the MIMIC-III-rare50 dataset \citep{yang2022knowledge}, which features 50 rare ICD codes, and the MIMIC-III-few dataset \citep{yang2023multi}, a subset with 685 unique ICD-9 codes occurring between 1 and 5 times in the training set. We denote these datasets as \mimicfull, \mimicfifty, \mimicrare, \mimicfew, and \mimicfour (refer to Table~\ref{tab:data-stat} for statistics). Following prior research \citep{mullenbach-etal-2018-explainable, xie-knowgraph19, li2020icd}, we tokenize and lowercase all text while eliminating non-alphabetic tokens containing numbers or punctuation.

To assess generalizability beyond the clinical domain,  we also experiment with two large-scale extreme multilabel datasets. The \eurlex dataset, comprising 15,449 training and 3,865 test European Union legal documents annotated with 3,956 EUROVOC labels, supports automated legal topic classification, compliance analysis, and cross-lingual information retrieval (\url{http://manikvarma.org/downloads/XC/XMLRepository.html}). The \wikiten dataset, with 14,146 training and 6,616 test Wikipedia articles associated with 30,938 categories, facilitates automatic tagging, web-scale document organization, and content recommendation (\url{http://manikvarma.org/downloads/XC/XMLRepository.html}). Both datasets are used to study large-scale label spaces and imbalanced label distributions(refer to Table~\ref{tab:data-stat-xmtc} for statistics).

\begin{table*}[ht]
\caption{\label{tab:icdehr}
Examples of clinical text with ICD codes \citep{wang2024icdxml,zhang2025general}.
\textcolor{blue}{Blue}: code/label;
\textcolor{red}{red bold}: disease mentions;
\textcolor{teal}{teal}: other relevant clinical findings.
}
\centering
\small
\renewcommand{\arraystretch}{0.95}
\begin{tabularx}{\columnwidth}{X @{\hspace{4mm}} X @{\hspace{4mm}} X}
\toprule
\textcolor{blue}{$428.0$:} \emph{Congestive heart failure} &
\textcolor{blue}{$202.8$:} \emph{Other malignant lymphomas} &
\textcolor{blue}{$770.6$:} \emph{Transitory tachypnea of newborn} \\
$\cdots$ DIAGNOSES: \textcolor{red}{\textbf{Acute congestive heart failure}},
\textcolor{teal}{Diabetes mellitus}, \textcolor{teal}{Pulmonary edema} $\cdots$ &
$\cdots$ 55 year-old female with \textcolor{red}{\textbf{non Hodgkin's lymphoma}}
and \textcolor{teal}{C1 esterase inhibitor deficiency} $\cdots$ &
$\cdots$ Chest x-ray: \textcolor{red}{\textbf{transient tachypnea of the newborn}} with
\textcolor{teal}{respiratory distress} $\cdots$ \\
\bottomrule
\end{tabularx}
\end{table*}

\begin{table*}[ht]
\caption{\label{tab:data-stat} 
    Descriptive statistics for \mimicfull and \mimicfour discharge summary training sets.
}
\centering
\small
\resizebox{\columnwidth}{!}{%
\begin{tabular}{l ll}
\toprule
 & \mimicfull & \mimicfour \\
\midrule
Number of documents & 52,723 & 122,279 \\
Number of patients & 41,126 & 65,659 \\
Number of unique codes & 8,929 & 7,942 \\
Codes per instance: Median (IQR) & $14(10\text{--}20)$ & $14(9\text{--}20)$ \\
Words per document: Median (IQR) & $1,375(965\text{--}1,900)$ & $1,492(1,147\text{--}1,931)$ \\
Documents: Train/val/test [\%] & $90.5 / 3.1 / 6.4$ & $72.9 / 10.9 / 16.2$ \\
\bottomrule
\end{tabular}%
}

\end{table*}

\begin{table*}[ht]
\caption{\label{tab:data-stat-xmtc} 
    Descriptive statistics for publicly available XMTC datasets \eurlex and \wikiten.
}
\centering
\small
\resizebox{\columnwidth}{!}{%
\begin{tabular}{l ll}
\toprule
 & \eurlex & \wikiten \\
\midrule
Number of train documents & $15,449$ & $14,146$ \\
Number of test documents & $3,865$ & $6,616$ \\
Number of unique labels & $3,956$ & $30,938$ \\
Average number of labels per instance & $5.30$ & $18.64$ \\
Average number of instances per label & $20.79$ & $8.52$ \\
\bottomrule
\end{tabular}%
}

\end{table*}

\section{Baselines for comparisons}
\label{sec:appendix_baselines}
\phead{ICD Baselines:}
We compare \plant against a diverse set of ICD coding baselines spanning classical, recent, and few-shot paradigms.

\noindent\emph{Early deep learning models}:
CAML~\citep{mullenbach-etal-2018-explainable}, MSATT-KG~\citep{xie-knowgraph19}, MUltiResCNN~\citep{li2020icd}, and HyperCore~\citep{cao-etal-2020-hypercore}.

\noindent\emph{Attention- and hierarchy-based models:}
LAAT and JointLAAT~\citep{vu-label-attn2021}, ISD~\citep{zhou-etal-2021-automatic}, Effective-CAN~\citep{liu-etal-2021-effective}, Hierarchical~\citep{dai2022revisiting}, and MSMN~\citep{yuan-etal-2022-code}.

\noindent\emph{Recent pretraining and architecture innovations:}
DiscNet~\citep{zhang-etal-2022-automatic}, KEPTLongformer~\citep{yang2022knowledge}, PLM-ICD~\citep{huang2022plm}, AHDD~\citep{zhang2024novel}, CoRelation~\citep{luo2024corelation}, Contrastive~\citep{lu2023towards}, MIMIC-IV-Benchmark~\citep{nguyen2023mimic}, and PLM-CA~\citep{edin2024unsupervised}.

\noindent\emph{Few-shot ICD coding methods:}
AGMHT~\citep{song2021generalized}, RareCodes~\citep{chen2023rare}, GP~\citep{yang2023multi}, and KEPT~\citep{yang2022knowledge}.

\noindent\emph{Knowledge-injected models:}
KEMTL~\citep{li2023towards}, MRR~\citep{wang2024multi}, AKIL~\citep{wang2024auxiliary}, and GKI-ICD~\citep{zhang2025general}.

\phead{XMTC Baselines}: On the full \eurlex and \wikiten datasets (Table~\ref{tab:results_xmtc}) we compare \plant against PECOS \citep{yu2022pecos}, ICXML \citep{zhu2023icxml}, XRR \citep{xiong2023xrr}, X-Transformer with RDE \citep{shi2024residual}, MatchXML \citep{ye2024matchxml}, DE \citep{guptadual}, InceptionXML \citep{kharbanda2023inceptionxml} with GANDALF \citep{kharbanda2024labelcor}, CG \citep{chai2024compositional}, ELMO \citep{zhang2025elmo}, and ViXML \citep{ortego2026large}.

\phead{LLM Baselines}: For RQ5 and RQ6, the baselines are (i) the same LLM backbone fine-tuned with the same classification head but without \plant's Stage~1, and (ii) a training-free extract-then-link pipeline that uses the LLM off the shelf (App.~\ref{app:w1_w2_details}).

\section{Evaluation Metrics}
\label{sec:appendix_metrics}
\newcommand{\pspk}{\text{PSP@}k}

We focus on \microf, \macrof, \microp, \macrop, \micror, \macror, \microauc, \macroauc, \patk, \ratk, and \pspk~to compare with prior ICD studies. Micro-averaging treats each (text, code) pair individually, aggregating true positives, false positives, and false negatives across all instances. Macro-averaging computes metrics per label, giving more weight to infrequent labels. \microp~is the ratio of aggregated true positives to the sum of true positives and false positives, while \macrop~averages precision across all labels. \micror~is the ratio of aggregated true positives to the sum of true positives and false negatives, while \macror~averages recall across all labels. \microauc~computes the area under the ROC curve for all instances aggregated together, while \macroauc~averages the AUC scores across all labels. \patk~and \ratk~measure the proportion of the top $k$ predicted labels that match the ground truth, focusing on precision and recall, respectively.

\begin{align*}
\microp &= \frac{\sum_{i} \textsf{TP}_i}{\sum_{i} (\textsf{TP}_i + \textsf{FP}_i)} \\
\micror &= \frac{\sum_{i} \textsf{TP}_i}{\sum_{i} (\textsf{TP}_i + \textsf{FN}_i)} \\
\microf &= \frac{2 \cdot \sum_{i} \textsf{TP}_i}{\sum_{i} (\textsf{TP}_i + \textsf{FP}_i) + \sum_{i} (\textsf{TP}_i + \textsf{FN}_i)} \\
\microauc &= \int_{0}^{1} \textsf{TPR}_{\text{micro}}(\textsf{FPR}_{\text{micro}}) \, d\textsf{FPR}_{\text{micro}}
\end{align*}

\begin{align*}
\macrop &= \frac{1}{L} \sum_{i=1}^L \frac{\textsf{TP}_i}{\textsf{TP}_i + \textsf{FP}_i} \\
\macror &= \frac{1}{L} \sum_{i=1}^L \frac{\textsf{TP}_i}{\textsf{TP}_i + \textsf{FN}_i} \\
\macrof &= \frac{1}{L} \sum_{i=1}^L \frac{2 \cdot \textsf{TP}_i}{\textsf{TP}_i + \textsf{FP}_i + \textsf{TP}_i + \textsf{FN}_i} \\
\macroauc &= \frac{1}{L} \sum_{i=1}^L \int_{0}^{1} \textsf{TPR}_i(\textsf{FPR}_i) \, d\textsf{FPR}_i
\end{align*}

\begin{align*}
\patk &= \frac{1}{k} \sum_{i=1}^k \mathbb{I}\left[ \text{pred}_i \in Y \right] \\
\ratk &= \frac{1}{\min(k, |Y|)} \sum_{i=1}^k \mathbb{I}\left[ \text{pred}_i \in Y \right] \\
\pspk &= \frac{\sum_{l \in \text{rank}_k(\hat{y})} y_l / p_l}{\max_{\text{rank}_k} \sum_{l \in \text{rank}_k} y_l / p_l}
\end{align*}

where $\textsf{TP}_i$, $\textsf{FP}_i$, and $\textsf{FN}_i$ are the true positives, false positives, and false negatives for label $i$, respectively, $L$ is the total number of labels, $\textsf{TPR}_{\text{micro}}$ and $\textsf{FPR}_{\text{micro}}$ are the true positive rate and false positive rate for the aggregated micro-averaged data, $\textsf{TPR}_i$ and $\textsf{FPR}_i$ are the true positive rate and false positive rate for label $i$, $Y$ is the ground truth label set for an instance, and $\text{pred}_i$ is the $i$-th top predicted label, $\hat{y}$ denotes the ranked predicted labels, $y_l \in \{0,1\}$ indicates whether label $l$ is relevant, and $p_l$ is the propensity of label $l$ under the model of \citet{jain2016extreme} with $A=0.55$ and $B=1.5$ ($1/p_l$ is larger for rare labels). The denominator normalizes by the best score attainable from the instance's own relevant labels, so $\pspk \in [0,1]$.

\section{Statistical Significance}
\label{sec:appendix_stats_significance}

\paragraph{Statistical Significance via Wilcoxon Signed-Rank Test.}
We assess statistical significance using the non-parametric Wilcoxon Signed-Rank Test \citep{demvsar2006statistical} for comparing paired model outputs. 
For metrics computed at the instance level (e.g., \textsf{P@15}), we apply the test directly to the paired per-instance scores between the base model and its \plant-enhanced counterpart. 
For aggregate metrics such as \textsf{F1} and \textsf{AUC}, which are reported as single values over the full test set, we first collect $N$ paired scores—either from repeated evaluations (e.g., $N=10$ in $10$-fold cross-validation) or from $N$ bootstrap resamples. Let $\{a_1, a_2, \dots, a_N\}$ and $\{b_1, b_2, \dots, b_N\}$ denote the scores of the base model and the \plant-enhanced model, respectively. We compute the difference $d_i = b_i - a_i$ for each pair and rank the absolute values $|d_i|$ (excluding zeros), averaging ranks in the case of ties. Each rank is assigned the sign of $d_i$, and we compute the rank sums $W^+$ and $W^-$ over positive and negative differences. The test statistic is $W = \min(W^+, W^-)$.

For small $N$, statistical significance is determined using exact Wilcoxon critical values; for larger $N$, we apply the normal approximation with
\begin{align*}
\mu &= \frac{N(N+1)}{4}, \quad
\sigma = \sqrt{\frac{N(N+1)(2N+1)}{24}}, \\ 
z &= \frac{W - \mu}{\sigma}.
\end{align*}
We reject the null hypothesis of no difference if the resulting $p$-value is less than a threshold $\alpha$ (typically $0.05$). In our tables, statistically significant improvements are marked using \textsuperscript{\blueup}. This test is readily implemented in standard libraries such as \texttt{scipy.stats.wilcoxon} in Python or \texttt{wilcox.test(paired=TRUE)} in R. 

\paragraph{Reporting Gains with Confidence Intervals.}
We also report absolute gains along with $95\%$ confidence intervals (CI) using paired bootstrap resampling. For each evaluation metric, we draw $B = 1000$ bootstrap samples from the test set and compute the difference $\Delta_b = \mathsf{Metric}_b^{\text{\plant}} - \mathsf{Metric}_b^{\text{Base}}$ for each sample $b$. The reported gain is the mean $\hat{\mu}$ of $\{\Delta_b\}$, and the CI is computed using the percentile bootstrap method by taking the 2.5th and 97.5th percentiles of the empirical distribution of $\{\Delta_b\}$. 

We mark results as statistically significant only if the Wilcoxon signed-rank test ($\alpha{=}0.05$) is passed \emph{and} the 95\% CI excludes 0. In such cases, we annotate the score with a colored arrow: \textsuperscript{\blueup} for statistically significant gains and \textsuperscript{\reddown} for significant drops. If the CI includes 0, no arrow is shown.
For example, 
$14.7$\textsuperscript{\blueup}{ \textcolor{blue}{(+1.2},\ \textcolor{CIPlum}{[0.6,\ 1.8])}} 
indicates a statistically significant gain over the base model, while 
$70.1$\textsuperscript{\reddown}{\textcolor{red}{(-1.4},\ \textcolor{CIPlum}{[-2.1,\ -0.7])}} 
denotes a significant drop. In contrast, 
$73.8${ \textcolor{blue}{(+0.3},\ \textcolor{CIPlum}{[0.0,\ 0.6])}} 
is not statistically significant and is shown without an arrow.

\paragraph{Note on Bootstrap Resampling.}
When $N$ bootstrap resamples are used to obtain paired scores for the Wilcoxon signed-rank test, we note that the resamples are not strictly independent due to sampling with replacement. In such cases, we place primary emphasis on the bootstrap percentile confidence intervals for assessing the significance and magnitude of the difference; the Wilcoxon test is applied mainly for datasets where independent replicates (e.g., cross-validation folds) are available. Results are marked as statistically significant only when \emph{both} the 95\% CI excludes zero and the Wilcoxon test (where applicable) yields $p<0.05$.

\section{Additional Results}
\label{sec:appendix_results_detailed}

\begin{table*}[ht]
    \centering
    \small
    \resizebox{\textwidth}{!}{%
    \begin{tabular}{l cc cc c cc cc c}
        \toprule
        & \multicolumn{5}{c}{\textbf{\mimicfull}} 
        & \multicolumn{5}{c}{\textbf{\mimicfour}} \\
        \cmidrule(lr){2-6} \cmidrule(lr){7-11}
        \multirow{2}{*}{Model}
        & \multicolumn{2}{c}{\textsf{AUC}} 
        & \multicolumn{2}{c}{\textsf{F1}} 
        & \multirow{2}{*}{\textsf{P@15}}
        & \multicolumn{2}{c}{\textsf{AUC}} 
        & \multicolumn{2}{c}{\textsf{F1}} 
        & \multirow{2}{*}{\textsf{P@15}} \\
        \cmidrule(lr){2-3} \cmidrule(lr){4-5} 
        \cmidrule(lr){7-8} \cmidrule(lr){9-10}
        & Macro & Micro & Macro & Micro & 
        & Macro & Micro & Macro & Micro & \\
        \midrule
        \mistral                 & 90.8 & 98.9 & 13.5 & 62.0 & 63.5 & 90.2 & 98.7 & 20.0 & 57.0 & 53.8 \\
        \rowcolor{green!20} \mistral + \plant       & 
        \gain{98.1}{7.3} & \gain{\textbf{99.9}}{1.0} & 
        \gain{14.7}{1.2} & \gain{64.1}{2.1} & 
        \gain{65.8}{2.3} & 
        \gain{97.4}{7.2} & \gain{99.5}{0.8} & 
        \gain{23.0}{3.0} & \gain{59.2}{2.2} & 
        \gain{56.9}{3.1} \\

        \llama                  & 91.0 & 99.0 & 13.8 & 62.5 & 64.0 & 90.5 & 98.8 & 20.5 & 57.5 & 54.0 \\
        \rowcolor{green!20} \llama + \plant         & 
        \gain{\textbf{98.3}}{7.3} & \gain{99.8}{0.8} & 
        \gain{\textbf{15.0}}{1.2} & \gain{\textbf{64.5}}{2.0} & 
        \gain{\textbf{66.2}}{2.2} & 
        \gain{\textbf{97.6}}{7.1} & \gain{\textbf{99.6}}{0.8} & 
        \gain{\textbf{23.5}}{3.0} & \gain{\textbf{59.5}}{2.0} & 
        \gain{\textbf{57.0}}{3.0} \\

        \deepseek               & 90.6 & 98.8 & 13.2 & 61.8 & 63.2 & 90.0 & 98.6 & 19.8 & 56.8 & 53.5 \\
        \rowcolor{green!20} \deepseek + \plant     & 
        \gain{97.9}{7.3} & \gain{99.7}{0.9} & 
        \gain{14.5}{1.3} & \gain{64.0}{2.2} & 
        \gain{65.5}{2.3} & 
        \gain{97.2}{7.2} & \gain{99.4}{0.8} & 
        \gain{22.8}{3.0} & \gain{59.0}{2.2} & 
        \gain{56.5}{3.0} \\

        \phiThree               & 90.4 & 98.7 & 13.0 & 61.5 & 63.0 & 89.8 & 98.5 & 19.5 & 56.5 & 53.2 \\
        \rowcolor{green!20} \phiThree + \plant      & 
        \gain{97.7}{7.3} & \gain{99.6}{0.9} & 
        \gain{14.3}{1.3} & \gain{63.8}{2.3} & 
        \gain{65.3}{2.3} & 
        \gain{97.0}{7.2} & \gain{99.3}{0.8} & 
        \gain{22.5}{3.0} & \gain{58.8}{2.3} & 
        \gain{56.3}{3.1} \\

        \midrule
        \rowcolor{green!20}
        \textbf{Avg.\ gain with \plant} &
        \avgain{7.3} & \avgain{0.9} &
        \avgain{1.3} & \avgain{2.2} & \avgain{2.3} &
        \avgain{7.2} & \avgain{0.8} &
        \avgain{3.0} & \avgain{2.2} & \avgain{3.1} \\
        \bottomrule
    \end{tabular}%
    }
    \caption{\label{tab:sota_llm_vs_plant_full}
        \textbf{Performance of LLMs with and without \plant.} 
        Each model is evaluated standalone and with \plant on \mimicfull and \mimicfour. 
        Green rows denote results after integrating \plant. 
        Bold values indicate the best score for each metric. 
        A compact version with only \mimicfour results is provided in 
        Table~\ref{tab:sota_llm_vs_plant_mimiciv} in the main paper.
    }
\end{table*}

\begin{table*}[htp]
    \centering
    \small
    \resizebox{\textwidth}{!}{%
    \begin{tabular}{l cc cc cc cc cc c}
        \toprule
        & \multicolumn{6}{c}{\textbf{\mimicfull}} 
        & \multicolumn{5}{c}{\textbf{\mimicfifty}} \\
        \cmidrule(lr){2-7} \cmidrule(lr){8-12}
        \multirow{2}{*}{Model}
        & \multicolumn{2}{c}{$\mathsf{AUC}$} & \multicolumn{2}{c}{$\mathsf{F1}$} & \multicolumn{2}{c}{$\mathsf{Precision}$}
        & \multicolumn{2}{c}{$\mathsf{AUC}$} & \multicolumn{2}{c}{$\mathsf{F1}$} & \multirow{2}{*}{\patfive} \\
        \cmidrule(lr){2-3} \cmidrule(lr){4-5} \cmidrule(lr){6-7}
        \cmidrule(lr){8-9} \cmidrule(lr){10-11}
        & Macro & Micro & Macro & Micro & \pateight & \patfifteen 
        & Macro & Micro & Macro & Micro & \\
        \midrule
        Effective-CAN~\cite{liu-etal-2021-effective}       & 92.1 & 98.9 & 10.6 & 58.9 & 75.8 & 60.6 & 92.0 & 94.5 & 66.8 & 71.7 & 66.4 \\
        MSMN~\cite{yuan-etal-2022-code}                    & 95.0 & 99.2 & 10.3 & 58.4 & 75.2 & 59.9 & 92.8 & 94.7 & 68.3 & 72.5 & 68.0 \\
        PLM-ICD~\citep{huang2022plm}                       & 92.6 & 98.9 & 10.4 & 59.8 & 77.1 & 61.3 & 91.0 & 93.4 & 66.3 & 71.9 & 66.0 \\
        Contrastive + JointLAAT~\citep{lu2023towards}      & 94.1 & 98.8 & 11.5 & 58.3 & 73.9 & 59.4 & 91.3 & 93.7 & 67.2 & 72.0 & 67.9 \\
        KEMTL~\citep{li2023towards}                        & 95.3 & 99.6 & 12.7 & 58.3 & 75.6 & 59.3 & 94.8 & 95.5 & 69.5 & 72.9 & 70.8 \\
        AHDD~\citep{zhang2024novel}                        & 95.2 & 99.3 & 10.9 & 58.9 & 75.3 & 60.1 & 92.8 & 94.7 & 68.5 & 72.8 & 67.8 \\
        CoRelation~\citep{luo2024corelation}               & 95.2 & 99.2 & 10.2 & 59.1 & 76.2 & 60.7 & 93.3 & 95.1 & 69.3 & 73.1 & 68.3 \\
        PLM-CA~\citep{edin2024unsupervised}                & 91.6 & 98.9 & 10.3 & 59.9 & 77.2 & 61.6 & 91.6 & 93.6 & 67.1 & 71.0 & 66.4 \\
        MRR~\citep{wang2024multi}                          & 94.9 & 99.5 & 11.4 & 60.3 & 77.5 & 62.3 & 92.7 & 94.7 & 68.7 & 73.2 & 68.5 \\
        AKIL~\citep{wang2024auxiliary}                     & 94.8 & 99.4 & 11.2 & 60.5 & 78.4 & 63.7 & 92.8 & 95.0 & 69.2 & 73.4 & 68.3 \\
        GKI-ICD~\citep{zhang2025general}                   & 96.2 & 99.3 & 12.3 & 61.2 & 77.7 & 62.4 & 93.3 & 95.2 & 69.2 & 73.5 & 68.1 \\
        \midrule
        \rowcolor{green!20}
        \plant (Ours) & 
        \textbf{98.1} & \textbf{99.9} & \textbf{14.7} & \textbf{64.1} & \textbf{80.3} & \textbf{65.8} 
        & \textbf{95.1} & \textbf{96.1} & \textbf{69.9} & \textbf{73.8} & \textbf{70.9} \\

        \rowcolor{green!10}
        & 
        \gainonly{+1.9}{1.15}{2.72} & \gainonly{+0.3}{0.02}{0.61}
        & 
        \gainonly{+2.0}{1.26}{2.58} & \gainonly{+2.9}{2.14}{3.41} & \gainonly{+1.9}{1.02}{2.83} & \gainonly{+2.1}{1.33}{2.74}
        & \gainonly{+0.3}{-0.01}{0.58} & \gainonly{+0.6}{0.17}{1.06} & \gainonly{+0.4}{0.05}{0.64} %
        & \gainonly{+0.3}{-0.05}{0.63} & \gainonly{+0.1}{-0.19}{0.39} \\

        \rowcolor{green!10}
        & 
        \cionly{1.15}{2.72} & \cionly{0.02}{0.61}
        & 
        \cionly{1.26}{2.58} & \cionly{2.14}{3.41} & \cionly{1.02}{2.83} & \cionly{1.33}{2.74}
        & \cionly{-0.01}{0.58} & \cionly{0.17}{1.06} & \cionly{0.05}{0.64}
        & \cionly{-0.05}{0.63} & \cionly{-0.19}{0.39} \\
        \bottomrule
    \end{tabular}%
    }
    \caption{\label{tab:results_full_and_top50} 
    \textbf{\plant vs. SOTA models on \mimicfull and \mimicfifty.} 
    On \mimicfull, \plant achieves aggregate gains of \textbf{+1--3} across \textsf{AUC} (Macro), \textsf{F1}, and \textsf{Precision}, including a \textbf{+2} (95\% CI: \textbf{1.3--2.6}) gain in \textsf{F1} (Macro). 
    For \mimicfifty (top 50 most frequent codes), gains are more modest, averaging around \textbf{+0.3} (e.g., \textbf{+0.4} in \textsf{F1} (Macro), 95\% CI: \textbf{0.1--0.6}). 
    }
\end{table*}

\begin{table}[t]
\centering
\small
\resizebox{\columnwidth}{!}{%
\begin{tabular}{l c c c c c c}
\toprule
\multirow{2}{*}{Model} 
 & \multicolumn{2}{c}{AUC} 
 & \multicolumn{2}{c}{F1} 
 & \multicolumn{2}{c}{P@15 / PSP@15} \\
\cmidrule(lr){2-3} \cmidrule(lr){4-5} \cmidrule(lr){6-7}
 & Macro & Micro & Macro & Micro & P & PSP \\
\midrule

\mistral 
 & 90.2 & 98.7 & 20.0 & 57.0 & 53.8 & 21.5 \\

\rowcolor{green!20}
\mistral + \plant 
 & \gain{97.4}{7.2} & \gain{99.5}{0.8} 
 & \gain{23.0}{3.0} & \gain{59.2}{2.2} 
 & \gain{56.9}{3.1} & \gain{24.8}{3.3} \\

\llama 
 & 90.5 & 98.8 & 20.5 & 57.5 & 54.0 & 21.6 \\

\rowcolor{green!20}
\llama + \plant
 & \gain{\textbf{97.6}}{7.1} & \gain{\textbf{99.6}}{0.8}
 & \gain{\textbf{23.5}}{3.0} & \gain{\textbf{59.5}}{2.0}
 & \gain{\textbf{57.0}}{3.0} & \gain{\textbf{24.9}}{3.3} \\

\deepseek & 90.0 & 98.6 & 19.8 & 56.8 & 53.5 & 21.4 \\

\rowcolor{green!20} 
\deepseek + \plant &
\gain{97.2}{7.2} & \gain{99.4}{0.8} &
\gain{22.8}{3.0} & \gain{59.0}{2.2} &
\gain{56.5}{3.0} & \gain{24.7}{3.3} \\

\phiThree & 89.8 & 98.5 & 19.5 & 56.5 & 53.2 & 21.3 \\

\rowcolor{green!20} 
\phiThree + \plant &
\gain{97.0}{7.2} & \gain{99.3}{0.8} &
\gain{22.5}{3.0} & \gain{58.8}{2.3} &
\gain{56.3}{3.1} & \gain{24.6}{3.3} \\

\midrule

\rowcolor{green!20}
\textbf{Avg.\ gain with \plant}
 & \avgain{7.2} & \avgain{0.8}
 & \avgain{3.0} & \avgain{2.2} 
 & \avgain{3.1} & \avgain{3.3} \\

\bottomrule
\end{tabular}%
}
\caption{PLANT boosts LLMs across metrics on \mimicfour, with \textbf{PSP@15 emphasizing tail gains.} A compact version without propensity scores is provided in the main paper as Table~\ref{tab:sota_llm_vs_plant_mimiciv}.}
\label{tab:sota_llm_vs_plant_mimiciv_psp}
\end{table}

\begin{table*}[htb]
\centering
\small
\resizebox{\textwidth}{!}{%
\begin{tabular}{l cc cc cc cc cc}
\toprule
& \multicolumn{6}{c}{\textbf{\mimicfew}} & \multicolumn{4}{c}{\textbf{\mimicrare}} \\
\cmidrule(lr){2-7} \cmidrule(lr){8-11}
\multirow{2}{*}{Model}
& \multicolumn{2}{c}{$\mathsf{F1}$}
& \multicolumn{2}{c}{$\mathsf{Precision}$}
& \multicolumn{2}{c}{$\mathsf{Recall}$}
& \multicolumn{2}{c}{$\mathsf{AUC}$}
& \multicolumn{2}{c}{$\mathsf{F1}$} \\
\cmidrule(lr){2-3} \cmidrule(lr){4-5} \cmidrule(lr){6-7}
\cmidrule(lr){8-9} \cmidrule(lr){10-11}
& Macro & Micro & Macro & Micro & Macro & Micro
& Macro & Micro & Macro & Micro \\
\midrule
AGMHT~\citep{song2021generalized}
& 18.7 & 29.2 & 17.6 & 49.4 & 19.9 & 20.7
& --   & --   & --   & --   \\
KEPTLongformer~\citep{yang2022knowledge}
& --   & --   & --   & --   & --   & --
& 82.7 & 83.3 & 30.4 & 32.6 \\
MSMN~\citep{yuan-etal-2022-code}
& 4.3  & 8.5  & 4.5  & \textbf{70.9} & 4.2  & 4.5
& --   & --   & --   & --   \\
GP~\citep{yang2023multi}
& 30.2 & 35.3 & 27.9 & 38.5 & 32.9 & 32.6
& --   & --   & --   & --   \\
\midrule

\rowcolor{green!20}
\plant (Ours)
& \textbf{66.3} & \textbf{74.6} & \textbf{65.1} & 68.6 & \textbf{81.0} & \textbf{81.7}
& \textbf{95.6} & \textbf{96.0} & \textbf{82.6} & \textbf{84.2} \\

\rowcolor{green!10}
&
\gainonly{+36.1}{30.5}{41.7} & \gainonly{+39.3}{33.4}{44.8} & \gainonly{+37.2}{31.0}{43.5} &
\diponly{-2.3}{-3.7}{-1.0} & \gainonly{+48.1}{42.6}{54.0} & \gainonly{+49.1}{43.3}{54.8} &
\gainonly{+12.9}{9.5}{15.7} & \gainonly{+12.7}{9.6}{15.4} &
\gainonly{+52.2}{44.8}{60.0} & \gainonly{+51.6}{43.8}{58.4} \\

\rowcolor{green!10}
&
\cionly{30.5}{41.7} & \cionly{33.4}{44.8} & \cionly{31.0}{43.5} &
\cionly{-3.7}{-1.0} & \cionly{42.6}{54.0} & \cionly{43.3}{54.8} &
\cionly{9.5}{15.7} & \cionly{9.6}{15.4} &
\cionly{44.8}{60.0} & \cionly{43.8}{58.4} \\

\bottomrule
\end{tabular}%
}
\caption{\label{tab:results_mimic3_few_and_rare}
\textbf{Performance on rare labels.}
Baseline results are as reported by \citet{yang2023multi} (\mimicfew) and \citet{yang2022knowledge} (\mimicrare); -- = not reported. Gains are relative to the best baseline in each column.
\plant achieves substantial improvements on most metrics, with several gains exceeding +35 and percentile bootstrap CIs well-separated from zero.
}
\end{table*}

\begin{table*}[ht]
\centering
\small
\resizebox{\textwidth}{!}{
\begin{tabular}{lccc}
\toprule
\textbf{Ablation Config} & \macroAUC & \macrof & \patfifteen \\
\midrule
\multicolumn{4}{c}{\textbf{Dataset:} \mimicfull, \textbf{LLM:} \mistral} \\
\stageOne Single-Stage BCE                     
    & 92.0\dipci{-6.1}{-6.9}{-5.1} 
    & 10.0\dipci{-4.7}{-5.6}{-3.3} 
    & 60.0\dipci{-5.8}{-6.6}{-4.7} \\
\stageOne Single-Stage Focal Loss             
    & 93.5\dipci{-4.6}{-5.5}{-3.7} 
    & 11.5\dipci{-3.2}{-4.1}{-2.1} 
    & 61.5\dipci{-4.3}{-5.2}{-3.1} \\
\stageOne \plant\ w/ Vanilla BCE
    & 95.5\dipci{-2.6}{-3.3}{-1.9} 
    & 12.7\dipci{-2.0}{-2.6}{-1.3} 
    & 63.0\dipci{-2.8}{-3.4}{-2.1} \\
\stageTwo \plant w/o Label Smoothing         
    & 97.8\dipci{-0.3}{-0.6}{-0.1} 
    & 13.8\dipci{-0.9}{-1.4}{-0.4} 
    & 64.5\dipci{-1.3}{-1.9}{-0.7} \\
\stageTwo \plant w/o Hard Neg Mining         
    & 97.5\dipci{-0.6}{-1.0}{-0.2} 
    & 13.0\dipci{-1.7}{-2.3}{-1.1} 
    & 62.5\dipci{-3.3}{-4.0}{-2.4} \\
\stageOne \plant w/ Term Frequency           
    & 97.0\dipci{-1.1}{-1.8}{-0.5} 
    & 12.5\dipci{-2.2}{-2.9}{-1.6} 
    & 62.0\dipci{-3.8}{-4.5}{-2.9} \\
\stageOne \plant w/ MSE      
    & 96.5\dipci{-1.6}{-2.4}{-0.9} 
    & 12.0\dipci{-2.7}{-3.4}{-1.9} 
    & 61.8\dipci{-4.0}{-4.7}{-3.2} \\
\rowcolor{green!20}
\plant (full setup) 
    & \textbf{98.1} & \textbf{14.7} & \textbf{65.8} \\

\midrule
\multicolumn{4}{c}{\textbf{Dataset:} \mimicfour, \textbf{LLM:} \mistral} \\
\stageOne Single-Stage BCE                     
    & 91.0\dipci{-6.4}{-7.2}{-5.4} 
    & 18.0\dipci{-5.0}{-5.9}{-3.8} 
    & 52.0\dipci{-4.9}{-5.7}{-3.9} \\
\stageOne Single-Stage Focal Loss            
    & 92.5\dipci{-4.9}{-5.8}{-4.0} 
    & 19.5\dipci{-3.5}{-4.2}{-2.7} 
    & 53.5\dipci{-3.4}{-4.0}{-2.7} \\
\stageOne \plant w/ Vanilla BCE
    & 95.0\dipci{-2.4}{-3.0}{-1.7} 
    & 21.0\dipci{-2.0}{-2.6}{-1.4} 
    & 54.8\dipci{-2.1}{-2.7}{-1.5} \\
\stageTwo \plant w/o Label Smoothing         
    & 97.0\dipci{-0.4}{-0.7}{-0.2} 
    & 21.8\dipci{-1.2}{-1.8}{-0.7} 
    & 55.8\dipci{-1.1}{-1.7}{-0.6} \\
\stageTwo \plant w/o Hard Neg Mining         
    & 96.8\dipci{-0.6}{-1.0}{-0.3} 
    & 21.0\dipci{-2.0}{-2.7}{-1.3} 
    & 54.5\dipci{-2.4}{-3.1}{-1.7} \\
\stageOne \plant w/ Term Frequency           
    & 96.5\dipci{-0.9}{-1.4}{-0.5} 
    & 20.5\dipci{-2.5}{-3.2}{-1.8} 
    & 54.0\dipci{-2.9}{-3.6}{-2.1} \\
\stageOne \plant w/ MSE      
    & 96.0\dipci{-1.4}{-2.1}{-0.8} 
    & 20.0\dipci{-3.0}{-3.9}{-2.1} 
    & 53.8\dipci{-3.1}{-3.9}{-2.3} \\
\rowcolor{green!20}
\plant (full setup)
    & \textbf{97.4} & \textbf{23.0} & \textbf{56.9} \\
\bottomrule
\end{tabular}
}
\caption{\label{tab:plant_2stage_training_ablation_full}
Ablation results on \mimicfull and \mimicfour with base \mistral.
The single-stage rows use a different configuration from the baselines in Tables~\ref{tab:sota_llm_vs_plant_mimiciv} and~\ref{tab:sota_llm_vs_plant_full}, so their scores differ.
\plant's largest gains come from \textcolor{PlantGreen}{Stage~1} attention initialization via MIG{+}ranking, while \textcolor{Chocolate}{Stage~2} refinements (label smoothing, HNM, focal loss) add complementary improvements.
A compact version with only \mimicfour results is provided in
Table~\ref{tab:plant_2stage_training_ablation_mimiciv} in the main paper.
}
\end{table*}

\begin{figure*}[htp]
    \centering
    \begin{subfigure}{0.5\textwidth}
        \centering
        \includegraphics[width=\linewidth, keepaspectratio]{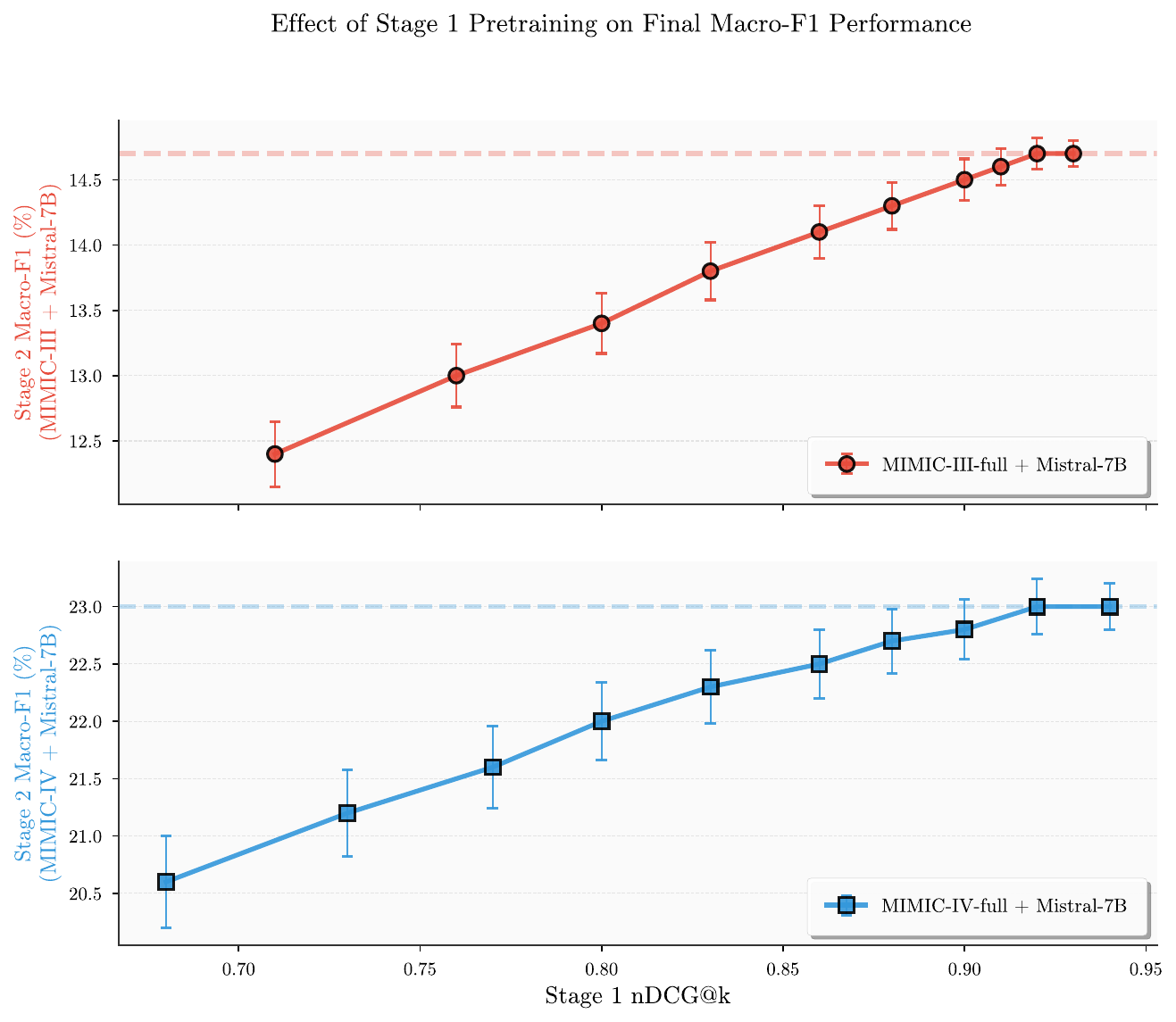}
    \end{subfigure}%
    \hfill%
    \begin{subfigure}{0.5\textwidth}
        \centering
        \includegraphics[width=\linewidth, keepaspectratio]{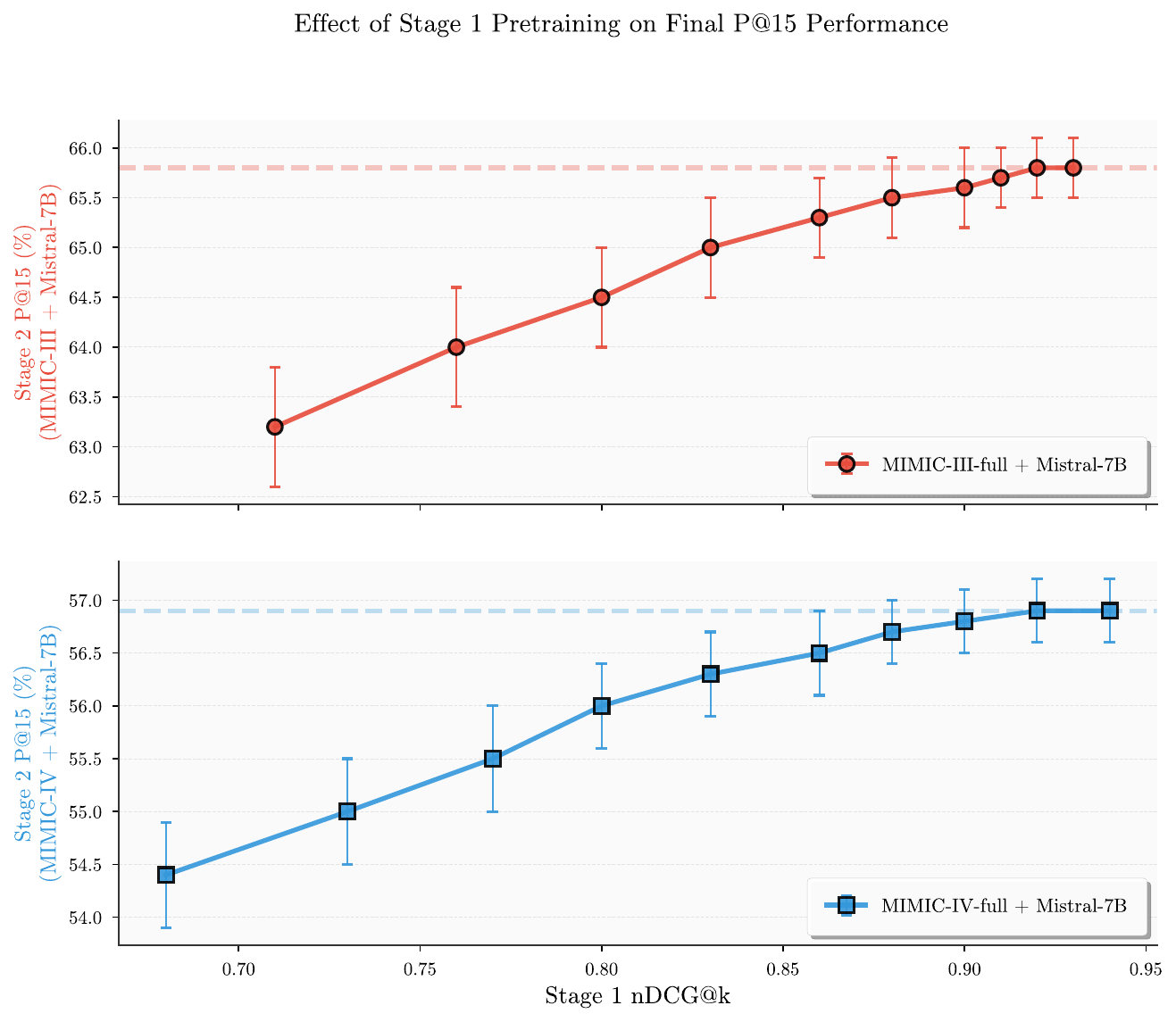}
    \end{subfigure}%
    \caption{%
     Effect of attention initialization quality on downstream performance on \mimicfull and \mimicfour with \mistral: as Stage~1 $\mathsf{nDCG}@k$ improves, final \macrof\ and \patfifteen\ after Stage~2 monotonically increase. The single-dataset view (\mimicfour, \macrof) is shown in the main paper as Figure~\ref{fig:effect_plant_initialization_plus_plant_vary_trn_size} (left).%
    }
    \label{fig:effect_plant_initialization}
\end{figure*}

\begin{figure*}[htp]
    \centering
    \begin{subfigure}{0.5\textwidth}
        \centering
        \includegraphics[width=\linewidth, height=4.28cm, keepaspectratio]{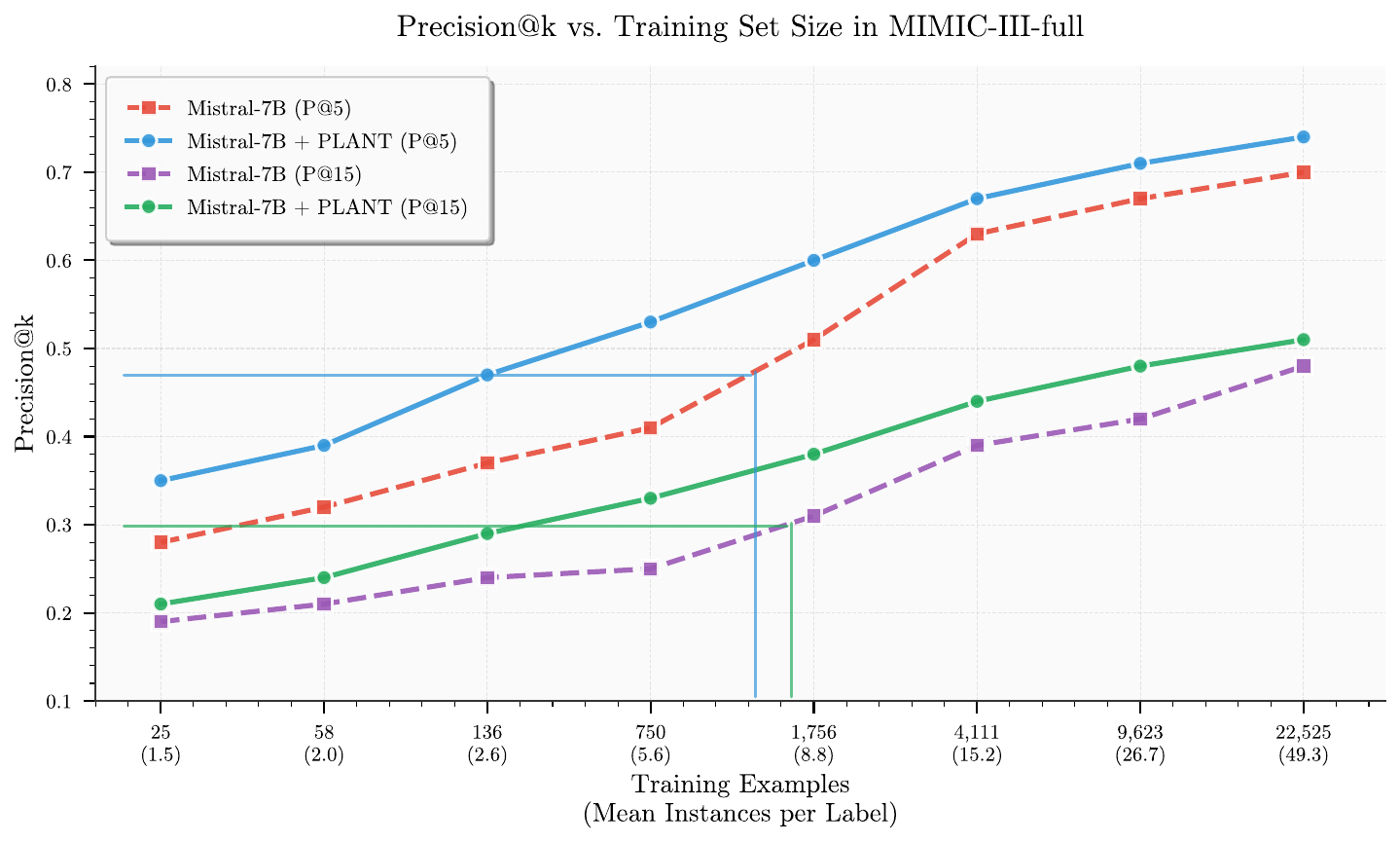}
    \end{subfigure}%
    \hfill%
    \begin{subfigure}{0.5\textwidth}
        \centering
        \includegraphics[width=\linewidth, height=4.28cm, keepaspectratio]{mimic4_icd10_Mistral7B_vs_Mistral7B_PLANT.pdf}
    \end{subfigure}%
    \caption{%
        \plant consistently boosts \mistral on \mimicfull\ (left) and \mimicfour\ (right) across training set sizes. 
        Solid lines (\mistral+\plant) outperform dashed lines (\mistral baseline) on both P@5 and P@15, with the largest gains appearing in low-data regimes. 
        Reference lines highlight that \plant reaches baseline performance using substantially fewer training examples.
        The single-dataset (\mimicfour only) view is shown in the main paper as Figure~\ref{fig:effect_plant_initialization_plus_plant_vary_trn_size} (right).%
    }
    \label{fig:plant_vary_trn_size}
\end{figure*}

\section{Controlled Comparisons with Fine-Tuned and Training-Free LLM Baselines}
\label{app:w1_w2_details}

This appendix gives the protocols and full results for RQ5 and RQ6 (Table~\ref{tab:w1_w2}).
Both studies use the \mistral and \llama checkpoints of the main experiments and \phiThreeSmall (7B) in place of \phiThree (App.~\ref{app:training_details}), on a single NVIDIA H100 80GB GPU, and report one run per configuration.
In both, the baseline and \plant share the same LoRA-adapted backbone $\mathcal{M}_{\mathsf{adapt}}$, data split, label space, classification head, and Stage~2 training configuration; the only difference is that \plant initializes the label-attention module with Stage~1 (MIG + L2R) before Stage~2, whereas the baseline trains it from random initialization. \plant therefore uses additional Stage~1 compute on top of a matched Stage~2 budget.
Unlike the main experiments (Section~\ref{subsection:architecture}), these runs initialize the label embeddings $\mathbf{E}$ randomly (truncated normal, standard deviation 0.01) rather than from label descriptions, for both the baseline and \plant.

\subsection{Fine-tuned LLMs outside the medical domain (RQ5)}
\label{app:w1_details}

\phead{Data}
Each subset keeps the 150 most frequent labels of the full dataset and samples 600 documents that carry them, with a fixed seed and 10\% held out for evaluation (540 train / 60 test).
After sampling, the median number of labels per document is 3 for \eurlex (6 in the full data) and 6 for \wikiten (19 in the full data).
The training labels remain skewed: 94 of 150 \eurlex labels (62.7\%) and 42 of 150 \wikiten labels (28.0\%) have fewer than 10 training examples.
Because the subsets keep only frequent labels and have small test sets, they complement but do not replace the full-label-space comparisons in Table~\ref{tab:results_xmtc}; we did not complete a matched full-dataset comparison.

\phead{Training budget and sensitivity}
On \eurlex, every stage was trained for 1 epoch. On \wikiten, the baseline and both \plant stages were trained for 3 epochs.
The \wikiten comparison was sensitive to this budget. With 1 epoch per stage, \plant trailed the \mistral baseline on every ranking metric (\textsf{P@5} 0.4674 vs.\ 0.4879; \textsf{P@25} 0.1607 vs.\ 0.1679; \textsf{R@25} 0.6630 vs.\ 0.6874; \textsf{PSP@25} 0.5248 vs.\ 0.5598). Training only \plant's Stage~1 for 3 epochs improved its rare-label F1 but not \textsf{P@5} (0.4698 vs.\ 0.5351 for a 1-epoch baseline). With 3 epochs for the baseline and both \plant stages, \plant outperformed the baseline on all ranking metrics (Table~\ref{tab:w1_full}).
Across these three runs, the \mistral baseline's \textsf{P@5} varied from 0.399 to 0.535, so single-run differences at this scale should be read as directional; \plant's 3-epoch \textsf{P@5} (0.599) exceeds all three baseline values.
\phiThreeSmall's \wikiten baseline is unusually weak (\textsf{P@5} 0.059), which inflates its relative gains; in absolute terms its \plant result is the lowest of the three backbones.

\begin{table*}[ht]
\centering
\small
\resizebox{\textwidth}{!}{%
\begin{tabular}{ll ccc ccc}
\toprule
Dataset & Metric & \multicolumn{2}{c}{\mistral} & \multicolumn{2}{c}{\llama} & \multicolumn{2}{c}{\phiThreeSmall} \\
\cmidrule(lr){3-4} \cmidrule(lr){5-6} \cmidrule(lr){7-8}
 & & FT & \plant & FT & \plant & FT & \plant \\
\midrule
\multirow{4}{*}{\eurlex (sample)}
 & \textsf{P@25}   & 0.0323 & 0.0465 & 0.0385 & 0.0540 & 0.0308 & 0.0428 \\
 & \textsf{R@25}   & 0.3164 & 0.4850 & 0.3805 & 0.5401 & 0.2906 & 0.3900 \\
 & \textsf{PSP@25} & 0.2617 & 0.4254 & 0.3189 & 0.4952 & 0.2463 & 0.3968 \\
 & \textsf{F1} (Micro) & 0.0352 & 0.0623 & 0.0272 & 0.0531 & 0.0348 & 0.1084 \\
\midrule
\multirow{6}{*}{\wikiten (sample)}
 & \textsf{P@5}    & 0.3988 & 0.5988 & 0.4065 & 0.5801 & 0.0591 & 0.4711 \\
 & \textsf{P@25}   & 0.1463 & 0.1908 & 0.1502 & 0.1938 & 0.0394 & 0.1545 \\
 & \textsf{R@25}   & 0.5753 & 0.7722 & 0.6252 & 0.7862 & 0.1272 & 0.6466 \\
 & \textsf{PSP@25} & 0.5228 & 0.6748 & 0.4651 & 0.6861 & 0.1674 & 0.4954 \\
 & \textsf{F1} (Micro) & 0.0927 & 0.5087 & 0.3502 & 0.4277 & 0.0824 & 0.1500 \\
 & \textsf{F1} (Macro) & 0.0763 & 0.2079 & 0.0196 & 0.2184 & 0.0720 & 0.0726 \\
\bottomrule
\end{tabular}%
}
\caption{\label{tab:w1_full}
Full results for RQ5: the same LLM fine-tuned without (FT) and with \plant on 600-document, 150-label subsets of \eurlex and \wikiten (60 test documents each; single runs). \eurlex: 1 epoch per stage; \wikiten: 3 epochs per stage.
}
\end{table*}

\subsection{Zero-shot and few-shot comparisons on MIMIC-IV-few (RQ6)}
\label{app:w2_details}

\phead{Dataset}
MIMIC-IV-few is built from \mimicfour (ICD-10). We selected the notes that carry at least one of 203 target codes occurring 3--7 times in the training split, kept each note's \emph{complete} ICD-10 annotation, and subsampled them, proportionally within each split and with a fixed seed, to 800 notes (516 train / 117 validation / 167 test). One of the 203 target codes loses all its notes in this subsample. The result has 3,343 labels and a median of 17 labels per note, so it remains an extreme multi-label task in which rare codes are over-represented relative to \mimicfour.

\phead{Frequency buckets and metrics}
Each label is assigned to a bucket using only its training-split frequency: zero-shot (0 occurrences; 358 labels), few-shot (1--7; 2,710 labels), and head ($\geq$8; 275 labels). We report \textsf{P@1}, \textsf{P@5}, and \textsf{P@8} within the zero-shot, few-shot, and head buckets, and \textsf{P@15} over all labels; each cutoff reflects the typical number of relevant labels per test note in that bucket. These threshold-free ranking metrics avoid threshold-calibration effects. A label's bucket (how often it was seen in training) is independent of a training-free model's prompting regime (how many in-context examples it receives).

\phead{Trained comparison}
The baseline and \plant were trained with the same data, label space, backbone, and budget (1 epoch for $\mathcal{M}_{\mathsf{adapt}}$; 6 epochs for the baseline and for each \plant stage; inputs truncated to 512 tokens). Both select checkpoints by validation few-shot-bucket recall@5.

\phead{Training-free extract-then-link pipeline}
For each test note, an off-the-shelf instruction-tuned LLM, with no MIMIC training and no list of candidate labels, extracts diagnoses and procedures as short phrases. In the zero-shot regime it receives no demonstrations; in the two-shot regime it receives two fixed, hand-authored generic examples, identical across models. (An earlier design that built the examples from training-note code descriptions led \llama to copy those descriptions into unrelated notes, so it was replaced.) SapBERT~\citep{liu2021self} links each extracted mention to the descriptions of the same 3,343 ICD-10 codes, and each code is scored by its best-matching mention, which yields a dense ranking evaluated with the same test notes, buckets, and precision@$k$ implementation as the trained models. We ran this pipeline for \mistral and \llama; we did not run it for \phiThreeSmall, and we did not evaluate \deepseek in either comparison because of its cost.

\phead{Findings}
\plant improves over the fine-tuned baseline in every bucket except \phiThreeSmall's few-shot bucket (\textsf{P@5} 0.0673 $\to$ 0.0581), where the baseline is the strongest fine-tuned few-shot baseline and \plant's few-shot validation recall stayed below that of the other two backbones. The training-free pipeline is competitive on rarely seen labels and exceeds \plant's few-shot-label \textsf{P@5} for both backbones on which it was run, while \plant is best on zero-shot-bucket, head, and overall precision (Table~\ref{tab:w2_full}).

\begin{table*}[ht]
\centering
\small
\resizebox{0.85\textwidth}{!}{%
\begin{tabular}{ll cccc}
\toprule
Backbone & Method & Zero-shot \textsf{P@1} & Few-shot \textsf{P@5} & Head \textsf{P@8} & Overall \textsf{P@15} \\
\midrule
\multirow{4}{*}{\mistral}
 & Fine-tuned baseline & 0.0055 & 0.0545 & 0.2766 & 0.2221 \\
 & Training-free, zero-shot prompt & 0.2754 & 0.1485 & 0.2964 & 0.2240 \\
 & Training-free, two-shot prompt & 0.2575 & 0.1545 & 0.2927 & 0.2271 \\
 & \plant & 0.3087 & 0.1160 & 0.3083 & 0.2460 \\
\midrule
\multirow{4}{*}{\llama}
 & Fine-tuned baseline & 0.0075 & 0.0087 & 0.1742 & 0.0985 \\
 & Training-free, zero-shot prompt & 0.2275 & 0.1389 & 0.2762 & 0.2104 \\
 & Training-free, two-shot prompt & 0.2335 & 0.1389 & 0.2837 & 0.2228 \\
 & \plant & 0.2839 & 0.1340 & 0.3423 & 0.2695 \\
\midrule
\multirow{2}{*}{\phiThreeSmall}
 & Fine-tuned baseline & 0.0112 & 0.0673 & 0.1716 & 0.1076 \\
 & \plant & 0.1118 & 0.0581 & 0.2901 & 0.2141 \\
\bottomrule
\end{tabular}%
}
\caption{\label{tab:w2_full}
Full results for RQ6 on MIMIC-IV-few (167 test notes; single runs). Buckets are defined by training-split label frequency.
}
\end{table*}

\section{Detailed Efficiency, Memory, and Inference Benchmarks}
\label{app:efficiency_benchmarks}

For completeness and reproducibility, this appendix provides expanded efficiency measurements, detailing wall-clock training time, GPU memory usage, and inference throughput across all experimental settings.

\plant adds no parameters beyond the label-attention head that vanilla fine-tuning also uses: the label embeddings $\mathbf{E} \in \mathbb{R}^{|\mathcal{L}| \times h}$ ($|\mathcal{L}|\,h$ parameters), the multi-head attention module $\mathsf{MultiHead}$ (standard multi-head attention, about $4h^2$ parameters), and the boosting matrices $\mathbf{B}_a, \mathbf{B}_m \in \mathbb{R}^{|\mathcal{L}| \times h}$ ($|\mathcal{L}|\,h$ parameters each). For a 7--8B backbone ($h = 4096$) on \mimicfour ($|\mathcal{L}| = 7{,}942$), $\mathbf{E}$ has about 32.5M parameters, $\mathsf{MultiHead}$ about 67M, and $\mathbf{B}_a, \mathbf{B}_m$ about 65M together: roughly 165M, or 2.4\% of a 7B backbone; the same head is used by the single-stage baselines. These parameters are optimized in Stage~1 and refined in Stage~2.

The incremental cost of \plant comes from Stage~1: MIG pre-computation on CPU followed by L2R optimization of $\mathcal{M}_{\mathsf{adapt}}$, $\mathsf{MultiHead}$, and $\mathbf{E}$ (at most 10 epochs). Relative to single-stage fine-tuning, this adds 12--28\% to total wall-clock training time on \mimicfour (20.8\% on average; Table~\ref{tab:efficiency-comparison-mimiciv}). Stage~2 employs end-to-end discriminative fine-tuning (up to 20 epochs with early stopping) on the full model. All timings and memory are empirically measured under the described regimen: 8$\times$ NVIDIA A100-80GB GPUs with DeepSpeed ZeRO-3 offloading, FP16 mixed-precision, global batch size 256 (per-device batch 8, 4$\times$ accumulation for effective per-device 32), sequence length 2048, and AdamW optimization. MIG pre-computation uses CPU (Intel Xeon, 64 cores) for efficiency. Early stopping (patience=2) typically halts Stage 1 at 7-8 epochs and Stage 2 at 12-15 epochs. Inference is a single forward pass per document (no generation) with batch size 1, on a single A100 GPU for the 7--8B backbones and on the same 8$\times$A100 node for \deepseek.

These costs are dominated by forward passes and full-model gradients in Stage 1 and 2, scaling with dataset size and model scale (e.g., DeepSeek-V3's 671B total/37B active MoE parameters incur $\sim$3$\times$ overhead vs. 7-8B dense models). With 4-bit quantization and LoRA, total training fits a standard 8-GPU node.

\begin{table}[t]
\centering
\small
\resizebox{\columnwidth}{!}{%
\begin{tabular}{l c c c c}
\toprule
\multirow{2}{*}{\textbf{Backbone}} 
  & \multirow{2}{*}{\textbf{Dataset}} 
  & \multicolumn{2}{c}{\textbf{Training Time (Wall-Clock Hours)}} 
  & \multirow{2}{*}{\textbf{Total}} \\
\cmidrule(lr){3-4}
& & \textbf{Stage 1 (MIG + L2R, $\leq$10 epochs)} 
  & \textbf{Stage 2 (End-to-End, up to 20 epochs)} 
  & \\ 
\midrule
\multirow{4}{*}{\mistral}
& MIMIC-III & 2.1 & 8.4 & 10.5 \\
& MIMIC-IV  & 4.2 & 18.7 & 22.9 \\
& EUR-LEX   & 1.4 & 3.2  & 4.6  \\
& WikiTen   & 1.3 & 2.9  & 4.2  \\
\midrule
\multirow{4}{*}{\llama}
& MIMIC-III & 2.2 & 9.1  & 11.3 \\
& MIMIC-IV  & 4.3 & 20.2 & 24.5 \\
& EUR-LEX   & 1.5 & 3.5  & 5.0  \\
& WikiTen   & 1.4 & 3.2  & 4.6  \\
\midrule
\multirow{4}{*}{\deepseek}
& MIMIC-III & 3.8 & 28.6 & 32.4 \\
& MIMIC-IV  & 7.5 & 63.4 & 70.9 \\
& EUR-LEX   & 2.6 & 11.8 & 14.4 \\
& WikiTen   & 2.4 & 10.7 & 13.1 \\
\midrule
\multirow{4}{*}{\phiThree}
& MIMIC-III & 1.6 & 5.2  & 6.8  \\
& MIMIC-IV  & 3.2 & 11.6 & 14.8 \\
& EUR-LEX   & 1.1 & 2.0  & 3.1  \\
& WikiTen   & 1.0 & 1.8  & 2.8  \\
\bottomrule
\end{tabular}%
}
\caption{Training Time (Wall-Clock Hours) by Backbone, Dataset, and Stage.}
\label{tab:training-time}
\end{table}

\noindent \textbf{Notes (Table~\ref{tab:training-time}):} Times scale with train-set size (MIMIC-III $\sim$47K docs; MIMIC-IV $\sim$89K; EUR-LEX/WikiTen $\sim$14--15K). MIG ($\sim$60\% of Stage~1) scales with document length (median 1{,}375--1{,}492 words for MIMIC). \phiThree (3.8B parameters) needs 33--40\% less training time than \mistral and \llama; DeepSeek-V3 is $\sim$3$\times$ slower due to scale. Multi-GPU scaling efficiency: 85--90\% (measured via strong scaling).

\begin{table}[t]
\centering
\small
\resizebox{\columnwidth}{!}{%
\begin{tabular}{l c c c}
\toprule
\textbf{Backbone} & \textbf{Dataset} & \textbf{Stage 1 (MIG + L2R)} & \textbf{Stage 2 (End-to-End)} \\
\midrule
\multirow{4}{*}{\mistral}
& MIMIC-III & 12.4 & 28.7 \\
& MIMIC-IV  & 12.8 & 29.4 \\
& EUR-LEX   & 11.9 & 27.2 \\
& WikiTen   & 11.7 & 26.9 \\
\midrule
\multirow{4}{*}{\llama}
& MIMIC-III & 12.4 & 30.2 \\
& MIMIC-IV  & 12.8 & 30.9 \\
& EUR-LEX   & 11.9 & 28.5 \\
& WikiTen   & 11.7 & 28.2 \\
\midrule
\multirow{4}{*}{\deepseek}
& MIMIC-III & 45.2 & 68.1 \\
& MIMIC-IV  & 46.3 & 69.5 \\
& EUR-LEX   & 43.8 & 65.4 \\
& WikiTen   & 43.4 & 64.9 \\
\midrule
\multirow{4}{*}{\phiThree}
& MIMIC-III & 8.7 & 18.5 \\
& MIMIC-IV  & 9.1 & 19.2 \\
& EUR-LEX   & 8.3 & 17.1 \\
& WikiTen   & 8.1 & 16.8 \\
\bottomrule
\end{tabular}%
}
\caption{Peak Memory Usage (GB VRAM per GPU) by Backbone, Dataset, and Stage.}
\label{tab:memory-usage}
\end{table}

\noindent \textbf{Notes (Table~\ref{tab:memory-usage}):} ZeRO-3 offloads optimizer states and activations to CPU/NVMe, enabling sub-70GB per-GPU peaks (up to $\sim$500--550GB across the node, for \deepseek). Stage~1 is lighter (33--59\% less), since it optimizes only the ranking loss with no classification head and no hard-negative mining. Peaks occur during backward passes in the later Stage~2 unfreezing phases and scale mildly with dataset length (longer MIMIC docs). In Stage~2, DeepSeek-V3 requires $\sim$2.3$\times$ more due to MoE (37B active params); all configurations fit 8$\times$A100 without spillover. Gradient checkpointing reduces memory by $\sim$20\%.

\begin{table}[t]
\centering
\small
\resizebox{\textwidth}{!}{%
\begin{tabular}{l c c c c}
\toprule
\textbf{Backbone} 
& \textbf{MIMIC-III (1,375 words)} 
& \textbf{MIMIC-IV (1,492 words)} 
& \textbf{EUR-LEX}
& \textbf{WikiTen} \\
\midrule
\mistral  & 1.8 & 1.9 & 0.7 & 1.0 \\
\llama    & 1.9 & 2.0 & 0.8 & 1.1 \\
\deepseek & 4.2 & 4.5 & 1.6 & 2.3 \\
\phiThree & 1.2 & 1.3 & 0.4 & 0.7 \\
\bottomrule
\end{tabular}%
}
\caption{Inference Time (Seconds per Document) by Backbone and Dataset.}
\label{tab:inference-time}
\end{table}

\noindent \textbf{Notes (Table~\ref{tab:inference-time}):} End-to-end (token selection + leveraged attention + classification); averaged over 1,000 test documents (FP16, batch=1) on a single A100 for the 7--8B backbones and on the 8$\times$A100 node for \deepseek. No stage distinction post-training. Inference scales approximately linearly with input length; DeepSeek-V3 is $\sim$2.3$\times$ slower due to MoE routing. Compared to vanilla LLM inference, PLANT adds about 8--12\% latency (Table~\ref{tab:efficiency-comparison-mimiciv}).

For direct comparability with the baselines in Table~\ref{tab:sota_llm_vs_plant_mimiciv}---which evaluate vanilla LLMs (\mistral, \llama, \deepseek, \phiThree) versus their \plant-augmented counterparts on \textsc{MIMIC-IV}---we provide below a focused breakdown of wall-clock training time (full pipeline for \plant vs.\ single-stage fine-tuning for vanilla) and per-document inference time (averaged over 1{,}000 test samples, FP16, batch size{=}1; a single A100 for the 7--8B backbones and the 8$\times$A100 node for \deepseek). All numbers are empirically measured under identical regimens (8$\times$A100--80GB with DeepSpeed ZeRO-3 for training; sequence length{=}2048), isolating the contribution of MIG and L2R pretraining.

Vanilla baselines incur single-stage end-to-end fine-tuning (up to 20 epochs, with early stopping typically at 12--15), mirroring \plant's Stage~2 but without attention pretraining; hence their training time approximates \plant's Stage~2 duration. \plant adds 8--12\% to vanilla inference time (Table~\ref{tab:efficiency-comparison-mimiciv}). DeepSeek-V3 remains an outlier due to MoE scaling, but \plant's relative gains hold consistently across all backbones.

\begin{table*}[t]
\centering
\small
\resizebox{\columnwidth}{!}{%
\begin{tabular}{lccc}
\toprule
\textbf{Backbone} & \textbf{Training Time (Hours): Vanilla} & \textbf{Training Time (Hours): \plant (Total)} & \textbf{Inference Time (s/doc): Vanilla vs.\ \plant} \\
\midrule
\mistral   & 18.7 & 22.9 (+22.5\%) & 1.7 vs.\ 1.9 (+11.8\%) \\
\llama     & 20.2 & 24.5 (+21.3\%) & 1.8 vs.\ 2.0 (+11.1\%) \\
\deepseek  & 63.4 & 70.9 (+11.8\%) & 4.1 vs.\ 4.5 (+9.8\%)  \\
\phiThree  & 11.6 & 14.8 (+27.6\%) & 1.2 vs.\ 1.3 (+8.3\%)  \\
\midrule
\textbf{Avg.\ Overhead} & --- & +20.8\% & +10.3\% \\
\bottomrule
\end{tabular}%
}
\caption{Training and inference efficiency for \plant vs.\ vanilla LLM baselines on \textsc{MIMIC-IV}. Relative overhead (\%) is shown in parentheses. Training times reflect multi-GPU wall-clock with early stopping. \plant's Stage~1 introduces modest training overhead yet yields substantial Macro-F1 improvements (average +3.0; see Table~\ref{tab:sota_llm_vs_plant_mimiciv}).}
\label{tab:efficiency-comparison-mimiciv}
\end{table*}

\noindent\textbf{Notes (Table~\ref{tab:efficiency-comparison-mimiciv})}: Relative training overhead decreases with model size, from +27.6\% for \phiThree (3.8B) to +11.8\% for \deepseek, because Stage~1's MIG computation is a larger share of the total when Stage~2 is short.
Per-document inference stays below 5\,s even for DeepSeek-V3.

\section{Extended Qualitative Analyses of PLANT’s Attention Over ICD–10 Codes in \mimicfour}
\label{sec:appendix_attention_analysis}

This appendix provides extended qualitative analyses of PLANT’s attention distributions across a diverse set of ICD–10 codes, illustrating  whether \plant's attention mechanism can reliably ‘find the needle in the haystack’—i.e., highlight the most clinically informative tokens despite scarce training signal.
Each case study lists tokens with their attention weights in parentheses, on a $\log_{10}$ scale and averaged over documents; values closer to zero indicate more attention.
\paragraph{Case Study: ICD--10--PCS \texttt{B211YZZ} (Coronary Angiography, Multiple Vessels).}
For the imaging procedure code \texttt{B211YZZ}, which corresponds to \emph{Fluoroscopy of Multiple Coronary Arteries using Other Contrast}—the PCS representation of multivessel coronary angiography—PLANT’s attention-ranked tokens cluster tightly around coronary anatomy and catheterization-report language. The highest-attention items, including \emph{domin} ($-1.75$) referencing coronary dominance, \emph{impress} ($-2.15$) echoing the “Impression” section of radiology/cardiology reports, and explicit vessel-branch markers such as \emph{diagonal} ($-2.45$), \emph{coron} ($-2.67$), and \emph{circum} ($-2.86$), directly mirror the nomenclature of LAD, diagonal, and LCx territories. Additional coronary-specific stems appear through \emph{desc} ($-2.89$) for the anterior descending artery, \emph{flex} ($-3.03$) and \emph{cx} ($-3.80$) for the circumflex, \emph{marginal} ($-3.03$) for obtuse marginal branches, and \emph{lad} ($-3.69$) itself. Catheterization workflow language surfaces via \emph{pci} ($-2.99$), \emph{block} ($-3.38$), \emph{cluded} ($-3.51$), \emph{attack} ($-3.54$), and \emph{tro} ($-3.65$), reflecting documentation of occlusions, myocardial infarction, and troponin status. Broader coronary-report vocabulary—\emph{vessel} ($-3.28$), \emph{vessels} ($-3.81$), \emph{segments} ($-3.43$), \emph{regional} ($-3.68$), and \emph{chamber} ($-3.74$)—aligns with standard angiographic interpretation of multivessel disease and ventricular chamber findings. While a few ranked items (\emph{water} at $-3.80$, \emph{rog} at $-3.80$, \emph{ho} at $-3.82$) are unrelated, reflecting noise typical for long-tail procedural codes, the top-ranked tokens concentrate on coronary anatomy, perfusion territories, ischemic terminology, and procedural descriptors characteristic of multivessel coronary angiography.

\paragraph{Case Study: ICD--10--PCS \texttt{6A551Z3} (Extracorporeal Plasma Exchange).}
For the procedure code \texttt{6A551Z3}, corresponding to \emph{Pheresis of Plasma, Multiple}, PLANT’s attention-ranked tokens form a strikingly coherent clinical signature. The highest-weight terms---\emph{plasma} ($-4.35$), \emph{filtered} ($-3.87$), \emph{exchange} ($-5.11$), and \emph{sessions} ($-4.28$)---match the code's semantics: pheresis (character~3 = 5) of plasma (character~7 = 3) over multiple sessions (character~5 = 1). Immunologic and hematologic cues such as \emph{kap} ($-5.19$), \emph{lambda} ($-4.96$), \emph{chain} ($-5.36$), and \emph{binding} ($-5.51$) reflect canonical indications for plasma exchange including removal of autoantibodies, paraproteins, or light chains in disorders like TTP, MGUS, or myasthenic crisis. Additional contextually aligned tokens---\emph{MOG} ($-5.27$), associated with antibody-mediated demyelinating disease, and \emph{replace} ($-5.38$), referring to the replacement-fluid component of plasmapheresis---further reinforce the procedural context. Some ranked items (e.g., \emph{shore} at $-5.38$, \emph{changes} at $-5.46$) display the expected semantic drift characteristic of long-tail rare codes, but most top-ranked tokens are tightly aligned with the mechanics, indications, and workflow of plasma exchange.

\paragraph{Case Study: ICD--10--PCS \texttt{3E0G76Z} (Enteral Tube Feeding).}
For the procedure code \texttt{3E0G76Z}, which denotes \emph{Introduction of Nutritional Substance into Upper GI, Via Natural or Artificial Opening}, PLANT’s attention-ranked tokens once again align tightly with the procedural semantics. The highest-attention terms---\emph{feed} ($-1.74$), \emph{feeding} ($-2.41$), \emph{tube} ($-2.46$), and \emph{peg} ($-2.45$)---directly correspond to enteral access, including PEG, NG, OG, and G-tube nutrition administration. Tokens linked to clinical indications and workflow, such as \emph{nutrition} ($-3.12$), \emph{swallow} ($-3.48$), and \emph{asp} ($-3.48$), reflect the typical contexts of dysphagia, aspiration risk, and nutritional compromise that prompt tube placement. Additional procedure-adjacent items---\emph{placement} ($-4.07$), \emph{placed} ($-4.19$), \emph{flush} ($-4.12$), and \emph{enter} ($-4.14$)---capture routine elements of enteral tube management, from tube positioning to maintenance flushing and enteral delivery checks. Even shorthand tokens frequently used in EHRs, such as \emph{tf} ($-3.05$) for “tube feed,” further reinforce contextual correctness. Some residual terms (e.g., \emph{home} at $-4.00$, \emph{video} at $-3.85$) exhibit expected drift yet remain plausibly adjacent to common documentation environments in nutritional support and discharge planning.

\paragraph{Case Study: ICD--10--PCS \texttt{10D00Z1} (Low Cervical Cesarean Section).}
For the obstetric procedure code \texttt{10D00Z1}, defined as \emph{Extraction of Products of Conception, Open Approach (Low Cervical Cesarean Section)}, PLANT’s attention-ranked tokens align almost perfectly with the linguistic and clinical setting of C\textsection\ delivery. The highest-attention items---\emph{ces} ($-1.78$), \emph{labor} ($-2.50$), and \emph{fet} ($-2.61$)---directly invoke cesarean delivery, active labor, and fetal extraction, which map precisely onto the PCS characters for extraction (character~3 = D) and products of conception (character~4 = 0). Additional obstetric markers such as \emph{bree} ($-2.77$), referencing breech presentation, and \emph{gest} ($-2.87$) and \emph{grav} ($-2.99$), denoting gestational age and gravida status, further reinforce labor and delivery context. Tokens reflecting pregnancy-related physiology and documentation---\emph{pregnancy} ($-3.37$), \emph{born} ($-3.10$), \emph{infant} ($-3.92$), and \emph{delivered} ($-4.02$)---capture routine narrative elements of cesarean operative notes. Procedure-form descriptors such as \emph{section} ($-3.88$), \emph{plac} ($-3.41$) for placenta, and \emph{fund} ($-4.26$) for fundal height or fundal pressure mirror common surgical and peripartum terminology. Other ranked tokens (e.g., \emph{bp} at $-4.23$, \emph{term} at $-3.89$) are consistent with surrounding obstetric charting. Overall, the top-ranked tokens center on vocabulary characteristic of cesarean extraction, gestational assessment, and delivery documentation.

\paragraph{Case Study: ICD--10--CM \texttt{Z85.828} (Personal History of Skin Malignancy).}
For the diagnosis code \texttt{Z85.828}, which denotes \emph{Personal History of Other Malignant Neoplasm of Skin}, PLANT’s attention-ranked tokens form an extraordinarily coherent dermatologic cancer signature. The dominant cluster---\emph{amous} ($-3.24$), \emph{squ} ($-3.31$), \emph{cell} ($-4.15$), \emph{car} ($-4.84$), and \emph{oma} ($-5.06$)---precisely reconstructs the morphology of \emph{squamous cell carcinoma}, the most common underlying condition referenced by this history code. Additional cutaneous oncology cues such as \emph{ker} ($-5.38$) for keratinocyte origin, \emph{cin} ($-5.43$), and \emph{situ} ($-6.22$) for carcinoma in situ further reinforce the malignant skin context. Anatomical-site terms frequently noted in dermatology documentation---\emph{scal} ($-6.67$), \emph{cheek} ($-6.77$), \emph{forehead} ($-6.86$), and \emph{temple} ($-7.06$)---reflect common SCC/BCC presentation areas. Surveillance and procedural tokens such as \emph{exc} ($-6.40$), referencing excision, and \emph{state} ($-6.80$), used in healed-treatment-status descriptions, align with the longitudinal follow-up nature of Z85.xx encounters. Some ranked items (e.g., \emph{daily} at $-6.87$, \emph{withdrawal} at $-6.79$) are typical outpatient-note background language, but most top-ranked tokens are morphologic and anatomic features characteristic of prior cutaneous malignancy.

\paragraph{Case Study: ICD--10--CM \texttt{C83.18} (Mantle Cell Lymphoma, Multiple Sites).}
For the diagnosis code \texttt{C83.18}, corresponding to \emph{Mantle Cell Lymphoma involving multiple lymph node regions}, PLANT’s attention-ranked tokens map strikingly well onto the characteristic vocabulary of B-cell lymphomas and hematopathology reporting. The top-ranked term, \emph{mant} ($-1.81$), directly invokes the mantle zone origin that defines this lymphoma subtype. Several additional high-attention tokens correspond to hallmark diagnostic and therapeutic features: \emph{chrom} ($-5.08$), referencing chromosomal abnormalities such as the canonical \emph{t(11;14)} translocation; \emph{hyper} ($-4.19$), capturing phrases like “hypercellular marrow”; \emph{rit} ($-4.43$), aligning with \emph{rituximab}—a standard anti-CD20 therapy; \emph{bend} ($-3.56$), suggestive of \emph{bendamustine}, a common MCL chemotherapeutic; and \emph{ki} ($-4.26$), which closely matches \emph{Ki-67}, the proliferation index routinely reported in mantle-cell pathology. Terms such as \emph{subset} ($-4.55$), \emph{characteristic} ($-4.69$), \emph{expression} ($-5.34$), and \emph{aggreg} ($-5.08$) reflect flow-cytometry and histopathology language describing immunophenotypic subsets, characteristic patterns, gene or protein expression, and atypical lymphoid aggregates. Additional pathology-adjacent items—\emph{oli} ($-5.14$) echoing monoclonality, \emph{phase} ($-4.98$) found in marrow-phase descriptors, and \emph{killer} ($-5.35$) associated with cytotoxic effector terminology—further reinforce the hematologic context. Some ranked terms (e.g., \emph{publicly} at $-5.19$, \emph{crowds} at $-5.12$) behave as expected sparse-class noise, while the top-ranked tokens concentrate on the morphologic, genetic, and therapeutic markers typical of mantle cell lymphoma.

\paragraph{Case Study: ICD--10--CM \texttt{H54.8} (Legal Blindness, U.S. Definition).}
For the diagnosis code \texttt{H54.8}, representing \emph{Legal Blindness as Defined in the U.S.A.}, PLANT’s attention-ranked tokens capture an ophthalmology-centric signal with striking precision. The most prominent items---\emph{legally} ($-1.34$), \emph{blind} ($-1.93$), and \emph{legal} ($-4.55$)---directly encode the definitional language of this code, which requires severe visual acuity or field loss in the better-seeing eye. Core ocular terminology appears immediately in tokens such as \emph{eye} ($-6.21$), \emph{ret} ($-6.36$) referencing the retina, \emph{mac} ($-5.83$) evoking macular disease, and \emph{degener} ($-6.92$), all of which reflect the major etiologies of profound vision loss, including macular degeneration and advanced retinal disorders. Additional high-salience terms---\emph{diab} ($-5.66$), consistent with diabetic retinopathy; \emph{drop} ($-6.21$) and \emph{drops} ($-6.24$), common in ophthalmic therapy documentation; and \emph{achment} ($-6.02$), suggestive of retinal detachment---further reinforce a pathology-driven visual impairment context. Symptom descriptors typical of low-vision notes, including \emph{shapes} ($-6.51$), \emph{shadows} ($-6.82$), and \emph{perception} ($-6.64$), likewise map to patient-reported experiences in severe visual loss. Other ranked terms (e.g., \emph{commission} at $-6.62$, \emph{indices} at $-6.42$) reflect administrative or evaluative language often co-documented in disability or certification settings. Overall, the top-ranked tokens center on the anatomical, etiologic, and functional descriptors characteristic of legal blindness assessments.

\paragraph{Case Study: ICD--10--CM \texttt{Z56.0} (Unemployment).}
For the socioeconomic code \texttt{Z56.0}, denoting \emph{Unemployment, Unspecified}, PLANT’s attention-ranked tokens yield a highly coherent social-determinants signature centered on joblessness, financial strain, and housing instability. The top-ranked items---\emph{unem} ($-3.62$), \emph{ployed} ($-3.63$), and \emph{unemployment} ($-3.69$)---explicitly encode the concept of lacking employment, which is the precise meaning of the code. Surrounding terms capture downstream consequences commonly documented in SDOH narratives: \emph{income} ($-5.02$) and \emph{money} ($-5.02$) reflecting financial insecurity; \emph{homeless} ($-4.67$), \emph{housing} ($-4.89$), and \emph{shelter} ($-4.97$) capturing housing precarity; and \emph{streets} ($-4.17$) evoking street exposure or unstable living conditions. Additional socio-environmental correlates such as \emph{illegal} ($-4.69$), \emph{criminal} ($-4.93$), and \emph{unsafe} ($-4.93$) mirror the high-risk social contexts frequently co-coded with Z56.x encounters. Psychosocial terms, including \emph{struggle} ($-4.74$), \emph{harm} ($-4.94$), and \emph{thoughts} ($-4.98$), align with mental-health stressors often accompanying unemployment. Workforce-barrier vocabulary, such as \emph{educ} ($-4.91$), \emph{skills} ($-4.90$), \emph{personal} ($-4.57$), and \emph{associations} ($-4.62$), reflects typical documentation in social-work assessments or care-coordination notes. While some ranked tokens are semantically diffuse, the top-ranked tokens concentrate on employment status, financial distress, and unstable housing—precisely the contextual cluster expected for \texttt{Z56.0}.

\section{Extended Discussion of Limitations}
\label{app:limitations}

This section expands on the limitations summarized in Section~\ref{sec:limitations}.

\phead{Dependence on labeled data}
Stage~1 relevance scores are mutual information computed from token--label co-occurrences in the labeled training split. Without label annotations these scores cannot be estimated, so \plant cannot pretrain attention in a fully unsupervised setting as currently formulated. Very sparse or noisy annotations make MIG estimates less reliable, which can reduce the benefit of the planted initialization; unsupervised relevance signals are a direction for future work. Within the evaluated settings, replacing MIG with normalized token frequency degrades rare-label and top-$k$ accuracy (Section~\ref{sec:ana}), so the choice of relevance signal matters.

\phead{Scale and domains}
We did not evaluate larger conventional XMTC benchmarks such as Wiki-500K or Amazon-3M, so scalability to label spaces of that size is untested. Most of our rare-label analyses use ICD coding. Outside the medical domain, the full-dataset comparisons with XMTC methods use a DistilBERT backbone (Table~\ref{tab:results_xmtc}), and the controlled comparisons with fine-tuned LLMs use 600-document subsets restricted to the 150 most frequent labels, evaluated on 60 documents per dataset with single runs (App.~\ref{app:w1_details}). That comparison was also sensitive to the training budget on \wikiten.

\phead{Rare and unseen labels}
On MIMIC-IV-few, a training-free extract-then-link pipeline exceeds \plant on few-shot-label precision, and \plant does not improve \phiThreeSmall's few-shot bucket (App.~\ref{app:w2_details}). These comparisons cover \mistral and \llama (training-free) and three backbones (fine-tuned), not \deepseek, and are single runs. \plant still depends on contextual evidence: labels with little semantic overlap with the input, or with insufficient evidence in the text, remain hard.

\phead{Negation and context}
Token-level MIG gives generic negation cues little label-specific relevance. Our account of why \plant still handles negated mentions, namely that attention operates over contextualized hidden states and Stage~2 trains end-to-end, is an explanation rather than a validated result: we have not run a controlled analysis with minimally different affirmed and negated inputs.

\phead{Transfer}
MIG and the attention initialization are computed on each target dataset's training split. The pretrained LLM backbone transfers across datasets, but planted attention is dataset-specific; transferring it between datasets that share labels or terminology may be possible but has not been evaluated.

\phead{Cost and modality}
Stage~1 adds 12--28\% to training time and about 10\% to inference latency on \mimicfour (App.~\ref{app:efficiency_benchmarks}), and the cost of MIG pre-computation grows with the label space. Our experiments use text classification with decoder-style LLM backbones (and DistilBERT for full-dataset \eurlex/\wikiten); other modalities are untested.

\section{Broader Impacts}
\label{app:broader_impacts}

This work aims to improve large-scale multi-label prediction systems by enabling more effective attention initialization for LLM-based classifiers. 
Potential positive applications include improved clinical coding, legal document organization, scientific literature tagging, and large-scale recommendation systems, particularly in settings with rare labels and limited supervision.

However, the same technology could also amplify biases present in pretrained models or training datasets. 
In sensitive domains such as healthcare, incorrect predictions or systematic bias against underrepresented classes could negatively affect downstream decision-making if deployed without appropriate human oversight. 
Additionally, improvements in scalable labeling systems may be misused for automated surveillance, profiling, or large-scale content categorization without adequate transparency. 
We therefore recommend that deployments involving high-stakes decisions include human review, careful auditing, and domain-specific validation.

%% file: sections/neurips_checklist.tex
\section*{NeurIPS Paper Checklist}

\begin{enumerate}

\item {\bf Claims}
    \item[] Question: Do the main claims made in the abstract and introduction accurately reflect the paper's contributions and scope?
    \item[] Answer: \answerYes{}
    \item[] Justification: The abstract and introduction state \plant's contribution and scope, including where it does not help (training-free baselines on few-shot labels; \phiThreeSmall's few-shot bucket) and that the controlled non-medical LLM comparisons are sample-scale (Sections~\ref{sec:exp}--\ref{sec:ana}).

\item {\bf Limitations}
    \item[] Question: Does the paper discuss the limitations of the work performed by the authors?
    \item[] Answer: \answerYes{}
    \item[] Justification: Section~\ref{sec:limitations} summarizes the limitations in the main paper, and Appendix~\ref{app:limitations} discusses them further: dependence on labeled data, untested larger benchmarks, sample-scale LLM comparisons, training-free baselines, negation, transfer, and cost.

\item {\bf Theory assumptions and proofs}
    \item[] Question: For each theoretical result, does the paper provide the full set of assumptions and a complete (and correct) proof?
    \item[] Answer: \answerNA{}
    \item[] Justification: The paper does not present theoretical results; its contribution is methodological and empirical.

    \item {\bf Experimental result reproducibility}
    \item[] Question: Does the paper fully disclose all the information needed to reproduce the main experimental results of the paper to the extent that it affects the main claims and/or conclusions of the paper (regardless of whether the code and data are provided or not)?
    \item[] Answer: \answerYes{}
    \item[] Justification: Sections~\ref{sec:approach}--\ref{sec:exp} and Appendices~\ref{app:training_details}, \ref{sec:appendix_implementation}, and \ref{app:w1_w2_details} describe datasets, architectures, optimization, evaluation, and the protocols of the controlled comparisons; code is released.

\item {\bf Open access to data and code}
    \item[] Question: Does the paper provide open access to the data and code, with sufficient instructions to faithfully reproduce the main experimental results, as described in supplemental material?
    \item[] Answer: \answerYes{}
    \item[] Justification: Code, preprocessing scripts, experiment configurations, and reproduction instructions are available at \url{https://github.com/deb-official/plant}. MIMIC data require credentialed PhysioNet access and are not redistributed; we provide scripts that rebuild the subsets used, including MIMIC-IV-few.

\item {\bf Experimental setting/details}
    \item[] Question: Does the paper specify all the training and test details (e.g., data splits, hyperparameters, how they were chosen, type of optimizer) necessary to understand the results?
    \item[] Answer: \answerYes{}
    \item[] Justification: Sections~\ref{sec:approach}--\ref{sec:exp} and Appendices~\ref{app:training_details}, \ref{sec:appendix_implementation}, and \ref{app:w1_w2_details} give splits, hyperparameters, optimizers, early stopping, and training budgets.

\item {\bf Experiment statistical significance}
    \item[] Question: Does the paper report error bars suitably and correctly defined or other appropriate information about the statistical significance of the experiments?
    \item[] Answer: \answerYes{}
    \item[] Justification: We report 95\% paired-bootstrap confidence intervals for gains (Appendix~\ref{sec:appendix_stats_significance}); Wilcoxon signed-rank tests are applied where independent replicates exist, and the causal analysis reports means $\pm$ s.d.\ over 5 folds. The controlled comparisons in RQ5--RQ6 are single runs without significance tests, as stated in the paper.

\item {\bf Experiments compute resources}
    \item[] Question: For each experiment, does the paper provide sufficient information on the computer resources (type of compute workers, memory, time of execution) needed to reproduce the experiments?
    \item[] Answer: \answerYes{}
    \item[] Justification: Main experiments use 8$\times$A100-80GB GPUs with DeepSpeed ZeRO-3 and FP16 (Appendix~\ref{app:training_details}); the RQ5--RQ6 comparisons use a single H100 80GB GPU (Appendix~\ref{app:w1_w2_details}). Appendix~\ref{app:efficiency_benchmarks} reports training time, memory, and inference time.
    
\item {\bf Code of ethics}
    \item[] Question: Does the research conducted in the paper conform, in every respect, with the NeurIPS Code of Ethics \url{https://neurips.cc/public/EthicsGuidelines}?
    \item[] Answer: \answerYes{}
    \item[] Justification: The research conforms to the NeurIPS Code of Ethics. Datasets and models are used according to their licenses; MIMIC data were accessed under PhysioNet's credentialed data use agreement.

\item {\bf Broader impacts}
    \item[] Question: Does the paper discuss both potential positive societal impacts and negative societal impacts of the work performed?
    \item[] Answer: \answerYes{}
    \item[] Justification: Appendix~\ref{app:broader_impacts} discusses beneficial applications (e.g., clinical coding and large-scale labeling) and risks, including bias, errors in high-stakes settings, and misuse.
    
\item {\bf Safeguards}
    \item[] Question: Does the paper describe safeguards that have been put in place for responsible release of data or models that have a high risk for misuse (e.g., pre-trained language models, image generators, or scraped datasets)?
    \item[] Answer: \answerNA{}
    \item[] Justification: The paper does not release models or datasets with a high risk of misuse beyond standard open LLM usage.

\item {\bf Licenses for existing assets}
    \item[] Question: Are the creators or original owners of assets (e.g., code, data, models), used in the paper, properly credited and are the license and terms of use explicitly mentioned and properly respected?
    \item[] Answer: \answerYes{}
    \item[] Justification: All datasets and pretrained models are cited (Section~\ref{sec:exp}, Appendices~\ref{app:training_details} and \ref{sec:appendix_implementation}). MIMIC-III/IV are used under the PhysioNet credentialed license, \eurlex and \wikiten come from the Extreme Classification Repository, and the LLM backbones are used under their Hugging Face licenses.

\item {\bf New assets}
    \item[] Question: Are new assets introduced in the paper well documented and is the documentation provided alongside the assets?
    \item[] Answer: \answerYes{}
    \item[] Justification: The released code and configurations are documented in the public repository. MIMIC-IV-few is not redistributed; we release the script that rebuilds it from MIMIC-IV.

\item {\bf Crowdsourcing and research with human subjects}
    \item[] Question: For crowdsourcing experiments and research with human subjects, does the paper include the full text of instructions given to participants and screenshots, if applicable, as well as details about compensation (if any)? 
    \item[] Answer: \answerNA{}
    \item[] Justification: The paper does not involve crowdsourcing or research with human subjects.

\item {\bf Institutional review board (IRB) approvals or equivalent for research with human subjects}
    \item[] Question: Does the paper describe potential risks incurred by study participants, whether such risks were disclosed to the subjects, and whether Institutional Review Board (IRB) approvals (or an equivalent approval/review based on the requirements of your country or institution) were obtained?
    \item[] Answer: \answerNA{}
    \item[] Justification: The work uses existing de-identified, credentialed-access clinical datasets (MIMIC-III/IV) and collects no new human-subject data.

\item {\bf Declaration of LLM usage}
    \item[] Question: Does the paper describe the usage of LLMs if it is an important, original, or non-standard component of the core methods in this research? Note that if the LLM is used only for writing, editing, or formatting purposes and does \emph{not} impact the core methodology, scientific rigor, or originality of the research, declaration is not required.
    \item[] Answer: \answerYes{}
    \item[] Justification: LLM backbones (\mistral, \llama, \deepseek, \phiThree) are core components of the \plant framework and are described and evaluated in the paper.

\end{enumerate}

%% file: main_neurips.bbl
\begin{thebibliography}{103}
\providecommand{\natexlab}[1]{#1}
\providecommand{\url}[1]{\texttt{#1}}
\expandafter\ifx\csname urlstyle\endcsname\relax
  \providecommand{\doi}[1]{doi: #1}\else
  \providecommand{\doi}{doi: \begingroup \urlstyle{rm}\Url}\fi

\bibitem[Abdin et~al.(2024)Abdin, Aneja, Awadalla, Awadallah, Awan, Bach,
  Bahree, Bakhtiari, Bao, Behl, et~al.]{abdin2024phi}
Marah Abdin, Jyoti Aneja, Hany Awadalla, Ahmed Awadallah, Ammar~Ahmad Awan,
  Nguyen Bach, Amit Bahree, Arash Bakhtiari, Jianmin Bao, Harkirat Behl, et~al.
\newblock Phi-3 technical report: A highly capable language model locally on
  your phone.
\newblock \emph{arXiv preprint arXiv:2404.14219}, 2024.

\bibitem[Aidouni(2024)]{aidouni2024masteringQLoRA}
Manal~El Aidouni.
\newblock Mastering {QLoRa}: A deep dive into 4‑bit quantization and {LoRa}
  parameter efficient fine‑tuning.
\newblock
  \url{https://manalelaidouni.github.io/4Bit-Quantization-Models-QLoRa.html},
  June 8 2024.

\bibitem[Asensio~Blasco et~al.(2025)Asensio~Blasco, Borrat~Frigola,
  Pastor~Duran, Conesa~Gonz{\'a}lez, Maci{\`a}, S{\'a}nchez~Barcenilla,
  Garrido~Bejar, and Frid]{asensio2025admission}
Elisa Asensio~Blasco, Xavier Borrat~Frigola, Xavier Pastor~Duran, Artur
  Conesa~Gonz{\'a}lez, Narc{\'\i}s Maci{\`a}, David S{\'a}nchez~Barcenilla,
  Ricardo Garrido~Bejar, and Santiago Frid.
\newblock From admission to discharge: Leveraging {NLP} for upstream primary
  coding with {SNOMED} {CT}.
\newblock \emph{Journal of Medical Systems}, 49\penalty0 (1):\penalty0 1--7,
  2025.

\bibitem[Bahdanau et~al.(2014)Bahdanau, Cho, and Bengio]{bahdanau2014neural}
Dzmitry Bahdanau, Kyunghyun Cho, and Yoshua Bengio.
\newblock Neural machine translation by jointly learning to align and
  translate.
\newblock \emph{arXiv preprint arXiv:1409.0473}, 2014.

\bibitem[Barreiros et~al.(2025)Barreiros, Coutinho, Correia, and
  Martins]{barreiros2025explainable}
Leonor Barreiros, Isabel Coutinho, Gon{\c{c}}alo~M Correia, and Bruno Martins.
\newblock Explainable {ICD} coding via entity linking.
\newblock \emph{arXiv preprint arXiv:2503.20508}, 2025.

\bibitem[Ben-Baruch et~al.(2020)Ben-Baruch, Ridnik, Zamir, Noy, Friedman,
  Protter, and Zelnik-Manor]{ben2020asymmetric}
Emanuel Ben-Baruch, Tal Ridnik, Nadav Zamir, Asaf Noy, Itamar Friedman, Matan
  Protter, and Lihi Zelnik-Manor.
\newblock Asymmetric loss for multi-label classification.
\newblock \emph{arXiv preprint arXiv:2009.14119}, 2020.

\bibitem[Bhatia et~al.(2016)Bhatia, Dahiya, Jain, Kar, Mittal, Prabhu, and
  Varma]{bhatia2016extreme}
Kush Bhatia, Kunal Dahiya, Himanshu Jain, Purushottam Kar, Anshul Mittal,
  Yashoteja Prabhu, and Manik Varma.
\newblock The extreme classification repository: Multi-label datasets and code.
\newblock \emph{URL http://manikvarma. org/downloads/XC/XMLRepository. html},
  2016.

\bibitem[Boukhers et~al.(2024)Boukhers, Khan, Ramadan, and
  Yang]{boukhers2024large}
Zeyd Boukhers, AmeerAli Khan, Qusai Ramadan, and Cong Yang.
\newblock Large language model in medical informatics: Direct classification
  and enhanced text representations for automatic {ICD} coding.
\newblock In \emph{2024 IEEE International Conference on Bioinformatics and
  Biomedicine (BIBM)}, pages 3066--3069. IEEE, 2024.

\bibitem[Boyle et~al.(2023)Boyle, Kascenas, Lok, Liakata, and
  O'Neil]{boyle2023automated}
Joseph~S Boyle, Antanas Kascenas, Pat Lok, Maria Liakata, and Alison~Q O'Neil.
\newblock Automated clinical coding using off-the-shelf large language models.
\newblock \emph{arXiv preprint arXiv:2310.06552}, 2023.

\bibitem[Cai et~al.(2024)Cai, Jiang, Wang, Tang, Kim, and Huang]{cai2024survey}
Weilin Cai, Juyong Jiang, Fan Wang, Jing Tang, Sunghun Kim, and Jiayi Huang.
\newblock A survey on mixture of experts.
\newblock \emph{arXiv preprint arXiv:2407.06204}, 2024.

\bibitem[Cao et~al.(2020)Cao, Chen, Liu, Zhao, Liu, and
  Chong]{cao-etal-2020-hypercore}
Pengfei Cao, Yubo Chen, Kang Liu, Jun Zhao, Shengping Liu, and Weifeng Chong.
\newblock {H}yper{C}ore: Hyperbolic and co-graph representation for automatic
  {ICD} coding.
\newblock In \emph{Proceedings of the 58th Annual Meeting of the Association
  for Computational Linguistics}, pages 3105--3114, Online, July 2020.
  Association for Computational Linguistics.
\newblock \doi{10.18653/v1/2020.acl-main.282}.
\newblock URL \url{https://aclanthology.org/2020.acl-main.282}.

\bibitem[{Centers for Medicare \& Medicaid Services}(2025)]{cms2026icd10}
{Centers for Medicare \& Medicaid Services}.
\newblock {ICD-10-CM} and {ICD-10-PCS} code files, {FY} 2026.
\newblock U.S. Department of Health and Human Services, 2025.
\newblock URL \url{https://www.cms.gov/medicare/coding-billing/icd-10-codes}.
\newblock 74,719 ICD-10-CM codes
  (\url{https://www.cms.gov/files/zip/2026-code-descriptions-tabular-order.zip})
  and 79,115 ICD-10-PCS codes
  (\url{https://www.cms.gov/files/zip/2026-icd-10-pcs-order-file-long-and-abbreviated-titles.zip}).
  Accessed: 2026-11-06.

\bibitem[Chai et~al.(2024)Chai, Li, Liu, Chen, Li, Ji, and
  Teng]{chai2024compositional}
Yuyang Chai, Zhuang Li, Jiahui Liu, Lei Chen, Fei Li, Donghong Ji, and Chong
  Teng.
\newblock Compositional generalization for multi-label text classification: A
  data-augmentation approach.
\newblock In \emph{Proceedings of the AAAI Conference on Artificial
  Intelligence}, volume~38, pages 17727--17735, 2024.

\bibitem[Chen et~al.(2023{\natexlab{a}})Chen, Li, Xi, Yu, and
  Xiong]{chen2023rare}
Jiamin Chen, Xuhong Li, Junting Xi, Lei Yu, and Haoyi Xiong.
\newblock Rare codes count: Mining inter-code relations for long-tail clinical
  text classification.
\newblock In \emph{Proceedings of the 5th Clinical Natural Language Processing
  Workshop}, pages 403--413, 2023{\natexlab{a}}.

\bibitem[Chen et~al.(2023{\natexlab{b}})Chen, Shen, Ding, Chen, Zhao,
  Learned-Miller, and Gan]{chen2023mod}
Zitian Chen, Yikang Shen, Mingyu Ding, Zhenfang Chen, Hengshuang Zhao, Erik~G
  Learned-Miller, and Chuang Gan.
\newblock Mod-squad: Designing mixtures of experts as modular multi-task
  learners.
\newblock In \emph{Proceedings of the IEEE/CVF Conference on Computer Vision
  and Pattern Recognition}, pages 11828--11837, 2023{\natexlab{b}}.

\bibitem[Dai et~al.(2022)Dai, Chalkidis, Darkner, and
  Elliott]{dai2022revisiting}
Xiang Dai, Ilias Chalkidis, Sune Darkner, and Desmond Elliott.
\newblock Revisiting transformer-based models for long document classification.
\newblock \emph{arXiv preprint arXiv:2204.06683}, 2022.

\bibitem[Dem{\v{s}}ar(2006)]{demvsar2006statistical}
Janez Dem{\v{s}}ar.
\newblock Statistical comparisons of classifiers over multiple data sets.
\newblock \emph{Journal of Machine learning research}, 7\penalty0
  (Jan):\penalty0 1--30, 2006.

\bibitem[Dettmers et~al.(2023)Dettmers, Pagnoni, Holtzman, and
  Zettlemoyer]{dettmers2023qlora}
Tim Dettmers, Artidoro Pagnoni, Ari Holtzman, and Luke Zettlemoyer.
\newblock {QLoRA}: Efficient finetuning of quantized {LLMs}.
\newblock \emph{Advances in neural information processing systems},
  36:\penalty0 10088--10115, 2023.

\bibitem[Dong et~al.(2021)Dong, Su{\'a}rez-Paniagua, Whiteley, and
  Wu]{dong2021explainable}
Hang Dong, V{\'\i}ctor Su{\'a}rez-Paniagua, William Whiteley, and Honghan Wu.
\newblock Explainable automated coding of clinical notes using hierarchical
  label-wise attention networks and label embedding initialisation.
\newblock \emph{Journal of biomedical informatics}, 116:\penalty0 103728, 2021.

\bibitem[Edin et~al.(2023)Edin, Junge, Havtorn, Borgholt, Maistro, Ruotsalo,
  and Maal{\o}e]{edin2023automated}
Joakim Edin, Alexander Junge, Jakob~D Havtorn, Lasse Borgholt, Maria Maistro,
  Tuukka Ruotsalo, and Lars Maal{\o}e.
\newblock Automated medical coding on {MIMIC-III} and {MIMIC-IV}: A critical
  review and replicability study.
\newblock \emph{arXiv preprint arXiv:2304.10909}, 2023.

\bibitem[Edin et~al.(2024)Edin, Maistro, Maal{\o}e, Borgholt, Havtorn, and
  Ruotsalo]{edin2024unsupervised}
Joakim Edin, Maria Maistro, Lars Maal{\o}e, Lasse Borgholt, Jakob~Drachmann
  Havtorn, and Tuukka Ruotsalo.
\newblock An unsupervised approach to achieve supervised-level explainability
  in healthcare records.
\newblock In Yaser Al-Onaizan, Mohit Bansal, and Yun-Nung Chen, editors,
  \emph{Proceedings of the 2024 Conference on Empirical Methods in Natural
  Language Processing}, pages 4869--4890, Miami, Florida, USA, November 2024.
  Association for Computational Linguistics.
\newblock \doi{10.18653/v1/2024.emnlp-main.280}.
\newblock URL \url{https://aclanthology.org/2024.emnlp-main.280/}.

\bibitem[Falis et~al.(2024)Falis, Gema, Dong, Daines, Basetti, Holder, Penfold,
  Birch, and Alex]{falis2024can}
Mat{\'u}{\v{s}} Falis, Aryo~Pradipta Gema, Hang Dong, Luke Daines, Siddharth
  Basetti, Michael Holder, Rose~S Penfold, Alexandra Birch, and Beatrice Alex.
\newblock Can {GPT-3.5} generate and code discharge summaries?
\newblock \emph{Journal of the American Medical Informatics Association},
  31\penalty0 (10):\penalty0 2284--2293, 2024.

\bibitem[Frantar et~al.(2022)Frantar, Ashkboos, Hoefler, and
  Alistarh]{frantar2022gptq}
Elias Frantar, Saleh Ashkboos, Torsten Hoefler, and Dan Alistarh.
\newblock Gptq: Accurate post-training quantization for generative pre-trained
  transformers.
\newblock \emph{arXiv preprint arXiv:2210.17323}, 2022.

\bibitem[Grattafiori et~al.(2024)Grattafiori, Dubey, Jauhri, Pandey, Kadian,
  Al-Dahle, Letman, Mathur, Schelten, Vaughan, et~al.]{grattafiori2024llama}
Aaron Grattafiori, Abhimanyu Dubey, Abhinav Jauhri, Abhinav Pandey, Abhishek
  Kadian, Ahmad Al-Dahle, Aiesha Letman, Akhil Mathur, Alan Schelten, Alex
  Vaughan, et~al.
\newblock The llama 3 herd of models.
\newblock \emph{arXiv preprint arXiv:2407.21783}, 2024.

\bibitem[Gupta et~al.(2024)Gupta, Devvrit, Rawat, Bhojanapalli, Jain, and
  Dhillon]{guptadual}
Nilesh Gupta, Fnu Devvrit, Ankit~Singh Rawat, Srinadh Bhojanapalli, Prateek
  Jain, and Inderjit~S Dhillon.
\newblock Dual-encoders for extreme multi-label classification.
\newblock In \emph{The Twelfth International Conference on Learning
  Representations}, 2024.

\bibitem[He et~al.(2025)He, Mao, Lin, Ruan, Lan, Feng, and
  Cambria]{he2025survey}
Kai He, Rui Mao, Qika Lin, Yucheng Ruan, Xiang Lan, Mengling Feng, and Erik
  Cambria.
\newblock A survey of large language models for healthcare: from data,
  technology, and applications to accountability and ethics.
\newblock \emph{Information Fusion}, page 102963, 2025.

\bibitem[Hou et~al.(2018)Hou, Pan, Loy, Wang, and Lin]{hou2018lifelong}
Saihui Hou, Xinyu Pan, Chen~Change Loy, Zilei Wang, and Dahua Lin.
\newblock Lifelong learning via progressive distillation and retrospection.
\newblock In \emph{Proceedings of the European Conference on Computer Vision
  (ECCV)}, pages 437--452, 2018.

\bibitem[Hsieh et~al.(2024)Hsieh, Sun, Kriman, Acharya, Rekesh, Jia, Zhang, and
  Ginsburg]{hsieh2024ruler}
Cheng-Ping Hsieh, Simeng Sun, Samuel Kriman, Shantanu Acharya, Dima Rekesh, Fei
  Jia, Yang Zhang, and Boris Ginsburg.
\newblock {RULER}: What's the real context size of your long-context language
  models?
\newblock \emph{arXiv preprint arXiv:2404.06654}, 2024.

\bibitem[Hu et~al.(2022)Hu, Shen, Wallis, Allen-Zhu, Li, Wang, Wang, Chen,
  et~al.]{hu2022lora}
Edward~J Hu, Yelong Shen, Phillip Wallis, Zeyuan Allen-Zhu, Yuanzhi Li, Shean
  Wang, Lu~Wang, Weizhu Chen, et~al.
\newblock Lora: Low-rank adaptation of large language models.
\newblock \emph{ICLR}, 1\penalty0 (2):\penalty0 3, 2022.

\bibitem[Huang et~al.(2022)Huang, Tsai, and Chen]{huang2022plm}
Chao-Wei Huang, Shang-Chi Tsai, and Yun-Nung Chen.
\newblock {PLM-ICD}: automatic {ICD} coding with pretrained language models.
\newblock \emph{arXiv preprint arXiv:2207.05289}, 2022.

\bibitem[Huang et~al.(2025)Huang, Yu, Yu, Qin, and Lin]{huang2025contrastive}
Hui Huang, Mingfeng Yu, Shuai Yu, Yongbin Qin, and Chuan Lin.
\newblock Contrastive learning-enhanced dual attention network for multi-label
  text classification.
\newblock \emph{Journal of King Saud University Computer and Information
  Sciences}, 37\penalty0 (6):\penalty0 136, 2025.

\bibitem[Jain et~al.(2016)Jain, Prabhu, and Varma]{jain2016extreme}
Himanshu Jain, Yashoteja Prabhu, and Manik Varma.
\newblock Extreme multi-label loss functions for recommendation, tagging,
  ranking \& other missing label applications.
\newblock In \emph{Proceedings of the 22nd ACM SIGKDD International Conference
  on Knowledge Discovery and Data Mining}, pages 935--944, 2016.

\bibitem[Jiang et~al.(2024)Jiang, Sablayrolles, Roux, Mensch, Savary, Bamford,
  Chaplot, Casas, Hanna, Bressand, et~al.]{jiang2024mixtral}
Albert~Q Jiang, Alexandre Sablayrolles, Antoine Roux, Arthur Mensch, Blanche
  Savary, Chris Bamford, Devendra~Singh Chaplot, Diego de~las Casas, Emma~Bou
  Hanna, Florian Bressand, et~al.
\newblock Mixtral of experts.
\newblock \emph{arXiv preprint arXiv:2401.04088}, 2024.

\bibitem[Johnson et~al.(2016)Johnson, Pollard, Shen, Lehman, Feng, Ghassemi,
  Moody, Szolovits, Anthony~Celi, and Mark]{johnson2016mimic}
Alistair~EW Johnson, Tom~J Pollard, Lu~Shen, Li-wei~H Lehman, Mengling Feng,
  Mohammad Ghassemi, Benjamin Moody, Peter Szolovits, Leo Anthony~Celi, and
  Roger~G Mark.
\newblock {MIMIC-III}, a freely accessible critical care database.
\newblock \emph{Scientific data}, 3\penalty0 (1):\penalty0 1--9, 2016.

\bibitem[Johnson et~al.(2023)Johnson, Bulgarelli, Shen, Gayles, Shammout,
  Horng, Pollard, Hao, Moody, Gow, et~al.]{johnson2023mimic}
Alistair~EW Johnson, Lucas Bulgarelli, Lu~Shen, Alvin Gayles, Ayad Shammout,
  Steven Horng, Tom~J Pollard, Sicheng Hao, Benjamin Moody, Brian Gow, et~al.
\newblock {MIMIC-IV}, a freely accessible electronic health record dataset.
\newblock \emph{Scientific data}, 10\penalty0 (1):\penalty0 1, 2023.

\bibitem[Kamradt(2023)]{kamradt2023needle}
Greg Kamradt.
\newblock Needle in a haystack - pressure testing {LLMs}.
\newblock \url{https://github.com/gkamradt/LLMTest_NeedleInAHaystack}, 2023.
\newblock Accessed: 2025-07-22.

\bibitem[Kang et~al.(2023)Kang, Wang, Xiong, Zhang, Zhou, Zhu, Zhang, and
  Tang]{kang2023automatic}
Beichen Kang, Xiaosu Wang, Yun Xiong, Yao Zhang, Chaofan Zhou, Yangyong Zhu,
  Jiawei Zhang, and Chunlei Tang.
\newblock Automatic {ICD} coding based on segmented {ClinicalBERT} with
  hierarchical tree structure learning.
\newblock In \emph{International Conference on Database Systems for Advanced
  Applications}, pages 250--265. Springer, 2023.

\bibitem[Kargupta et~al.(2023)Kargupta, Komarlu, Yoon, Wang, and
  Han]{kargupta2023megclass}
Priyanka Kargupta, Tanay Komarlu, Susik Yoon, Xuan Wang, and Jiawei Han.
\newblock Megclass: extremely weakly supervised text classification via
  mutually-enhancing text granularities.
\newblock \emph{arXiv preprint arXiv:2304.01969}, 2023.

\bibitem[Kharbanda et~al.(2022)Kharbanda, Banerjee, Schultheis, and
  Babbar]{kharbanda2022cascadexml}
Siddhant Kharbanda, Atmadeep Banerjee, Erik Schultheis, and Rohit Babbar.
\newblock Cascadexml: Rethinking transformers for end-to-end multi-resolution
  training in extreme multi-label classification.
\newblock \emph{Advances in neural information processing systems},
  35:\penalty0 2074--2087, 2022.

\bibitem[Kharbanda et~al.(2023)Kharbanda, Banerjee, Gupta, Palrecha, and
  Babbar]{kharbanda2023inceptionxml}
Siddhant Kharbanda, Atmadeep Banerjee, Devaansh Gupta, Akash Palrecha, and
  Rohit Babbar.
\newblock Inceptionxml: A lightweight framework with synchronized negative
  sampling for short text extreme classification.
\newblock In \emph{Proceedings of the 46th International ACM SIGIR Conference
  on Research and Development in Information Retrieval}, pages 760--769, 2023.

\bibitem[Kharbanda et~al.(2024)Kharbanda, Gupta, Schultheis, Banerjee, Hsieh,
  and Babbar]{kharbanda2024labelcor}
Siddhant Kharbanda, Devaansh Gupta, Erik Schultheis, Atmadeep Banerjee, Cho-Jui
  Hsieh, and Rohit Babbar.
\newblock Gandalf: Learning label-label correlations in extreme multi-label
  classification via label features.
\newblock In \emph{Proceedings of the 30th ACM SIGKDD Conference on Knowledge
  Discovery and Data Mining}, KDD '24, page 1360–1371, New York, NY, USA,
  2024. Association for Computing Machinery.
\newblock ISBN 9798400704901.
\newblock \doi{10.1145/3637528.3672063}.
\newblock URL \url{https://doi.org/10.1145/3637528.3672063}.

\bibitem[Li and Yu(2020)]{li2020icd}
Fei Li and Hong Yu.
\newblock {ICD} coding from clinical text using multi-filter residual
  convolutional neural network.
\newblock In \emph{proceedings of the AAAI conference on artificial
  intelligence}, volume~34, pages 8180--8187, 2020.

\bibitem[Li et~al.(2023)Li, Zhao, Zhang, and Xing]{li2023towards}
Xinhang Li, Xiangyu Zhao, Yong Zhang, and Chunxiao Xing.
\newblock Towards automatic {ICD} coding via knowledge enhanced multi-task
  learning.
\newblock In \emph{Proceedings of the 32nd ACM International Conference on
  Information and Knowledge Management}, pages 1238--1248, 2023.

\bibitem[Liu et~al.(2021)Liu, Shareghi, Meng, Basaldella, and
  Collier]{liu2021self}
Fangyu Liu, Ehsan Shareghi, Zaiqiao Meng, Marco Basaldella, and Nigel Collier.
\newblock Self-alignment pretraining for biomedical entity representations.
\newblock In \emph{Proceedings of the 2021 Conference of the North American
  Chapter of the Association for Computational Linguistics: Human Language
  Technologies}, pages 4228--4238, 2021.

\bibitem[Liu et~al.(2025{\natexlab{a}})Liu, Zhou, Gu, Zou, Huang, Wu, Li, Chen,
  Hua, Zhou, et~al.]{liu2025application}
Fenglin Liu, Hongjian Zhou, Boyang Gu, Xinyu Zou, Jinfa Huang, Jinge Wu, Yiru
  Li, Sam~S Chen, Yining Hua, Peilin Zhou, et~al.
\newblock Application of large language models in medicine.
\newblock \emph{Nature Reviews Bioengineering}, pages 1--20,
  2025{\natexlab{a}}.

\bibitem[Liu et~al.(2024{\natexlab{a}})Liu, Lin, Hewitt, Paranjape, Bevilacqua,
  Petroni, and Liang]{liu2024lost}
Nelson~F. Liu, Kevin Lin, John Hewitt, Ashwin Paranjape, Michele Bevilacqua,
  Fabio Petroni, and Percy Liang.
\newblock Lost in the middle: How language models use long contexts.
\newblock \emph{Transactions of the Association for Computational Linguistics},
  12:\penalty0 157--173, 2024{\natexlab{a}}.
\newblock \doi{10.1162/tacl_a_00638}.
\newblock URL \url{https://aclanthology.org/2024.tacl-1.9/}.

\bibitem[Liu et~al.(2024{\natexlab{b}})Liu, Wang, Yin, Molchanov, Wang, Cheng,
  and Chen]{liu2024dora}
Shih-Yang Liu, Chien-Yi Wang, Hongxu Yin, Pavlo Molchanov, Yu-Chiang~Frank
  Wang, Kwang-Ting Cheng, and Min-Hung Chen.
\newblock Dora: Weight-decomposed low-rank adaptation.
\newblock In \emph{Forty-first International Conference on Machine Learning},
  2024{\natexlab{b}}.

\bibitem[liu et~al.(2021)liu, cheng, klopfer, gormley, and
  schaaf]{liu-etal-2021-effective}
yang liu, hua cheng, russell klopfer, matthew~r. gormley, and thomas schaaf.
\newblock effective convolutional attention network for multi-label clinical
  document classification.
\newblock In \emph{proceedings of the 2021 conference on empirical methods in
  natural language processing}, pages 5941--5953, online and punta cana,
  dominican republic, November 2021. association for computational linguistics.
\newblock \doi{10.18653/v1/2021.emnlp-main.481}.
\newblock URL \url{https://aclanthology.org/2021.emnlp-main.481}.

\bibitem[Liu et~al.(2025{\natexlab{b}})Liu, Huang, Xia, and Zhang]{liu2025all}
Zhi Liu, Yunjie Huang, Xincheng Xia, and Yihao Zhang.
\newblock All is attention for multi-label text classification.
\newblock \emph{Knowledge and Information Systems}, 67\penalty0 (2):\penalty0
  1249--1270, 2025{\natexlab{b}}.

\bibitem[Loshchilov and Hutter(2017)]{loshchilov2017decoupled}
Ilya Loshchilov and Frank Hutter.
\newblock Decoupled weight decay regularization.
\newblock \emph{arXiv preprint arXiv:1711.05101}, 2017.

\bibitem[Lu et~al.(2023)Lu, Reddy, Wang, and Ning]{lu2023towards}
Chang Lu, Chandan Reddy, Ping Wang, and Yue Ning.
\newblock Towards semi-structured automatic {ICD} coding via tree-based
  contrastive learning.
\newblock \emph{Advances in Neural Information Processing Systems},
  36:\penalty0 68300--68315, 2023.

\bibitem[Luo et~al.(2024)Luo, Wang, Wang, Chang, Wang, and
  Ma]{luo2024corelation}
Junyu Luo, Xiaochen Wang, Jiaqi Wang, Aofei Chang, Yaqing Wang, and Fenglong
  Ma.
\newblock {CoRelation}: Boosting automatic {ICD} coding through contextualized
  code relation learning.
\newblock \emph{arXiv preprint arXiv:2402.15700}, 2024.

\bibitem[Ma et~al.(2025)Ma, Chochlakis, Pandiyan, Thomason, and
  Narayanan]{ma2025large}
Marcus Ma, Georgios Chochlakis, Niyantha~Maruthu Pandiyan, Jesse Thomason, and
  Shrikanth Narayanan.
\newblock Large language models do multi-label classification differently.
\newblock In \emph{Proceedings of the 2025 Conference on Empirical Methods in
  Natural Language Processing}, pages 2472--2495, 2025.

\bibitem[Madan et~al.(2024)Madan, Lentzen, Brandt, Rueckert, Hofmann-Apitius,
  and Fr{\"o}hlich]{madan2024transformer}
Sumit Madan, Manuel Lentzen, Johannes Brandt, Daniel Rueckert, Martin
  Hofmann-Apitius, and Holger Fr{\"o}hlich.
\newblock Transformer models in biomedicine.
\newblock \emph{BMC Medical Informatics and Decision Making}, 24\penalty0
  (1):\penalty0 214, 2024.

\bibitem[Michalopoulos et~al.(2022)Michalopoulos, Malyska, Sahar, Wong, and
  Chen]{michalopoulos2022icdbigbird}
George Michalopoulos, Michal Malyska, Nicola Sahar, Alexander Wong, and Helen
  Chen.
\newblock {ICDBigBird}: a contextual embedding model for {ICD} code
  classification.
\newblock \emph{arXiv preprint arXiv:2204.10408}, 2022.

\bibitem[Moons et~al.(2020)Moons, Khanna, Akkasi, and
  Moens]{moons2020comparison}
Elias Moons, Aditya Khanna, Abbas Akkasi, and Marie-Francine Moens.
\newblock A comparison of deep learning methods for {ICD} coding of clinical
  records.
\newblock \emph{Applied Sciences}, 10\penalty0 (15):\penalty0 5262, 2020.

\bibitem[Mullenbach et~al.(2018)Mullenbach, Wiegreffe, Duke, Sun, and
  Eisenstein]{mullenbach-etal-2018-explainable}
James Mullenbach, Sarah Wiegreffe, Jon Duke, Jimeng Sun, and Jacob Eisenstein.
\newblock Explainable prediction of medical codes from clinical text.
\newblock In \emph{Proceedings of the 2018 Conference of the North {A}merican
  Chapter of the Association for Computational Linguistics: Human Language
  Technologies, Volume 1 (Long Papers)}, pages 1101--1111, New Orleans,
  Louisiana, June 2018. Association for Computational Linguistics.
\newblock \doi{10.18653/v1/N18-1100}.
\newblock URL \url{https://aclanthology.org/N18-1100}.

\bibitem[Mullenbach et~al.(2021)Mullenbach, Pruksachatkun, Adler, Seale,
  Swartz, McKelvey, Dai, Yang, and Sontag]{mullenbach2021clip}
James Mullenbach, Yada Pruksachatkun, Sean Adler, Jennifer Seale, Jordan
  Swartz, T~Greg McKelvey, Hui Dai, Yi~Yang, and David Sontag.
\newblock {CLIP}: A dataset for extracting action items for physicians from
  hospital discharge notes.
\newblock \emph{arXiv preprint arXiv:2106.02524}, 2021.

\bibitem[Nerella et~al.(2024)Nerella, Bandyopadhyay, Zhang, Contreras, Siegel,
  Bumin, Silva, Sena, Shickel, Bihorac, et~al.]{nerella2024transformers}
Subhash Nerella, Sabyasachi Bandyopadhyay, Jiaqing Zhang, Miguel Contreras,
  Scott Siegel, Aysegul Bumin, Brandon Silva, Jessica Sena, Benjamin Shickel,
  Azra Bihorac, et~al.
\newblock Transformers and large language models in healthcare: A review.
\newblock \emph{Artificial intelligence in medicine}, page 102900, 2024.

\bibitem[Ng et~al.(2023)Ng, Santos, and Rei]{ng2023modelling}
Clarence Boon~Liang Ng, Diogo Santos, and Marek Rei.
\newblock Modelling temporal document sequences for clinical {ICD} coding.
\newblock \emph{arXiv preprint arXiv:2302.12666}, 2023.

\bibitem[Nguyen et~al.(2023)Nguyen, Schlegel, Kashyap, Winkler, Huang, Liu, and
  Lin]{nguyen2023mimic}
Thanh-Tung Nguyen, Viktor Schlegel, Abhinav Kashyap, Stefan Winkler, Shao-Syuan
  Huang, Jie-Jyun Liu, and Chih-Jen Lin.
\newblock {MIMIC-IV-ICD}: A new benchmark for extreme multilabel
  classification.
\newblock \emph{arXiv preprint arXiv:2304.13998}, 2023.

\bibitem[Ortego et~al.(2026)Ortego, Rodr{\'\i}guez, Almagro, Dahiya,
  Jim{\'e}nez, and SanMiguel]{ortego2026large}
Diego Ortego, Marlon Rodr{\'\i}guez, Mario Almagro, Kunal Dahiya, David
  Jim{\'e}nez, and Juan~C SanMiguel.
\newblock Large language models meet extreme multi-label classification:
  Scaling and multi-modal framework.
\newblock In \emph{Proceedings of the AAAI Conference on Artificial
  Intelligence}, volume~40, pages 24630--24638, 2026.

\bibitem[Peysakhovich and Lerer(2023)]{peysakhovich2023attention}
Alexander Peysakhovich and Adam Lerer.
\newblock Attention sorting combats recency bias in long context language
  models.
\newblock \emph{arXiv preprint arXiv:2310.01427}, 2023.

\bibitem[Reimers and Gurevych(2019)]{reimers2019sentence}
Nils Reimers and Iryna Gurevych.
\newblock Sentence-{BERT}: Sentence embeddings using {S}iamese {BERT}-networks.
\newblock In \emph{Proceedings of the 2019 Conference on Empirical Methods in
  Natural Language Processing and the 9th International Joint Conference on
  Natural Language Processing (EMNLP-IJCNLP)}, pages 3982--3992, 2019.

\bibitem[Sakai and Lam(2025)]{sakai2025large}
Hajar Sakai and Sarah~S Lam.
\newblock Large language models for healthcare text classification: A
  systematic review.
\newblock \emph{arXiv preprint arXiv:2503.01159}, 2025.

\bibitem[Shi et~al.(2024)Shi, Wei, and Li]{shi2024residual}
Jiang-Xin Shi, Tong Wei, and Yu-Feng Li.
\newblock Residual diverse ensemble for long-tailed multi-label text
  classification.
\newblock \emph{Science CHINA Information Science}, 2024.

\bibitem[Song et~al.(2021)Song, Zhang, Sadoughi, Xie, and
  Xing]{song2021generalized}
Congzheng Song, Shanghang Zhang, Najmeh Sadoughi, Pengtao Xie, and Eric Xing.
\newblock Generalized zero-shot text classification for {ICD} coding.
\newblock In \emph{Proceedings of the Twenty-Ninth International Conference on
  International Joint Conferences on Artificial Intelligence}, pages
  4018--4024, 2021.

\bibitem[Tabatabaei et~al.(2025)Tabatabaei, Fancher, Parsons, and
  Askari]{tabatabaei2025can}
Seyed~Amin Tabatabaei, Sarah Fancher, Michael Parsons, and Arian Askari.
\newblock Can large language models serve as effective classifiers for
  hierarchical multi-label classification of scientific documents at industrial
  scale?
\newblock In \emph{Proceedings of the 31st International Conference on
  Computational Linguistics: Industry Track}, pages 163--174, 2025.

\bibitem[Team et~al.(2023)Team, Anil, Borgeaud, Alayrac, Yu, Soricut,
  Schalkwyk, Dai, Hauth, Millican, et~al.]{team2023gemini}
Gemini Team, Rohan Anil, Sebastian Borgeaud, Jean-Baptiste Alayrac, Jiahui Yu,
  Radu Soricut, Johan Schalkwyk, Andrew~M Dai, Anja Hauth, Katie Millican,
  et~al.
\newblock Gemini: a family of highly capable multimodal models.
\newblock \emph{arXiv preprint arXiv:2312.11805}, 2023.

\bibitem[Teng et~al.(2024)Teng, Zhang, Zhou, Hu, and Li]{teng2024few}
Fei Teng, Quanmei Zhang, Xiaomin Zhou, Jie Hu, and Tianrui Li.
\newblock Few-shot {ICD} coding with knowledge transfer and evidence
  representation.
\newblock \emph{Expert Systems with Applications}, 238:\penalty0 121861, 2024.

\bibitem[Vandemoortele et~al.(2025)Vandemoortele, Steenwinckel, Ongenae, and
  Van~Hoecke]{vandemoortele2025haystack}
Nathan Vandemoortele, Bram Steenwinckel, Femke Ongenae, and Sofie Van~Hoecke.
\newblock From haystack to needle: Label space reduction for zero-shot
  classification.
\newblock \emph{arXiv preprint arXiv:2502.08436}, 2025.

\bibitem[Vaswani et~al.(2017)Vaswani, Shazeer, Parmar, Uszkoreit, Jones, Gomez,
  Kaiser, and Polosukhin]{vaswani2017attention}
Ashish Vaswani, Noam Shazeer, Niki Parmar, Jakob Uszkoreit, Llion Jones,
  Aidan~N Gomez, {\L}ukasz Kaiser, and Illia Polosukhin.
\newblock Attention is all you need.
\newblock \emph{Advances in neural information processing systems}, 30, 2017.

\bibitem[Vu et~al.(2021)Vu, Nguyen, and Nguyen]{vu-label-attn2021}
Thanh Vu, Dat~Quoc Nguyen, and Anthony Nguyen.
\newblock A label attention model for {ICD} coding from clinical text.
\newblock In \emph{Proceedings of the Twenty-Ninth International Joint
  Conference on Artificial Intelligence}, IJCAI'20, 2021.
\newblock ISBN 9780999241165.

\bibitem[Wang et~al.(2024{\natexlab{a}})Wang, Du, Jiang, Liu, Li, Chen, Gao,
  Xie, and Lee]{wang2024label}
Gang Wang, Yajun Du, Yurui Jiang, Jia Liu, Xianyong Li, Xiaoliang Chen, Hongmei
  Gao, Chunzhi Xie, and Yan-li Lee.
\newblock Label-text bi-attention capsule networks model for multi-label text
  classification.
\newblock \emph{Neurocomputing}, 588:\penalty0 127671, 2024{\natexlab{a}}.

\bibitem[Wang et~al.(2023{\natexlab{a}})Wang, Chen, Qin, He, and
  Lin]{wang2023multi}
Jiyao Wang, Zijie Chen, Yang Qin, Dengbo He, and Fangzhen Lin.
\newblock Multi-aspect co-attentional collaborative filtering for extreme
  multi-label text classification.
\newblock \emph{Knowledge-Based Systems}, 260:\penalty0 110110,
  2023{\natexlab{a}}.

\bibitem[Wang et~al.(2024{\natexlab{b}})Wang, Mercer, and
  Rudzicz]{wang2024multi}
Xindi Wang, Robert Mercer, and Frank Rudzicz.
\newblock Multi-stage retrieve and re-rank model for automatic medical coding
  recommendation.
\newblock In Kevin Duh, Helena Gomez, and Steven Bethard, editors,
  \emph{Proceedings of the 2024 Conference of the North American Chapter of the
  Association for Computational Linguistics: Human Language Technologies
  (Volume 1: Long Papers)}, pages 4881--4891, Mexico City, Mexico, June
  2024{\natexlab{b}}. Association for Computational Linguistics.
\newblock \doi{10.18653/v1/2024.naacl-long.273}.
\newblock URL \url{https://aclanthology.org/2024.naacl-long.273/}.

\bibitem[Wang et~al.(2024{\natexlab{c}})Wang, Mercer, and
  Rudzicz]{wang2024auxiliary}
Xindi Wang, Robert~E. Mercer, and Frank Rudzicz.
\newblock Auxiliary knowledge-induced learning for automatic multi-label
  medical document classification.
\newblock In Nicoletta Calzolari, Min-Yen Kan, Veronique Hoste, Alessandro
  Lenci, Sakriani Sakti, and Nianwen Xue, editors, \emph{Proceedings of the
  2024 Joint International Conference on Computational Linguistics, Language
  Resources and Evaluation (LREC-COLING 2024)}, pages 2006--2016, Torino,
  Italia, May 2024{\natexlab{c}}. ELRA and ICCL.
\newblock URL \url{https://aclanthology.org/2024.lrec-main.181/}.

\bibitem[Wang et~al.(2024{\natexlab{d}})Wang, Wang, Zhang, Wang, Qi, Chen,
  Sastry, Johnson, and De]{wang2024icdxml}
Zeqiang Wang, Yuqi Wang, Haiyang Zhang, Wei Wang, Jun Qi, Jianjun Chen,
  Nishanth Sastry, Jon Johnson, and Suparna De.
\newblock {ICDXML}: enhancing {ICD} coding with probabilistic label trees and
  dynamic semantic representations.
\newblock \emph{Scientific Reports}, 14\penalty0 (1):\penalty0 18319,
  2024{\natexlab{d}}.

\bibitem[Wang et~al.(2023{\natexlab{b}})Wang, Wang, Mekala, and
  Shang]{wang2023benchmark}
Zihan Wang, Tianle Wang, Dheeraj Mekala, and Jingbo Shang.
\newblock A benchmark on extremely weakly supervised text classification:
  Reconcile seed matching and prompting approaches.
\newblock \emph{arXiv preprint arXiv:2305.12749}, 2023{\natexlab{b}}.

\bibitem[WHO(2025)]{who2025icd}
WHO.
\newblock \emph{International Statistical Classification of Diseases and
  Related Health Problems}.
\newblock 11th edition, 2025.
\newblock URL
  \url{https://www.who.int/standards/classifications/classification-of-diseases}.

\bibitem[Wu et~al.(2024)Wu, Wu, and Sun]{wu2024dila}
John Wu, David Wu, and Jimeng Sun.
\newblock {DILA}: Dictionary label attention for mechanistic interpretability
  in high-dimensional multi-label medical coding prediction.
\newblock \emph{arXiv preprint arXiv:2409.10504}, 2024.

\bibitem[Xie et~al.(2019)Xie, Xiong, Yu, and Zhu]{xie-knowgraph19}
Xiancheng Xie, Yun Xiong, Philip~S. Yu, and Yangyong Zhu.
\newblock {EHR} coding with multi-scale feature attention and structured
  knowledge graph propagation.
\newblock In \emph{Proceedings of the 28th ACM International Conference on
  Information and Knowledge Management}, CIKM '19, page 649–658, New York,
  NY, USA, 2019. Association for Computing Machinery.
\newblock ISBN 9781450369763.
\newblock \doi{10.1145/3357384.3357897}.
\newblock URL \url{https://doi.org/10.1145/3357384.3357897}.

\bibitem[Xiong et~al.(2023)Xiong, Yu, Niu, and Leng]{xiong2023xrr}
Jie Xiong, Li~Yu, Xi~Niu, and Youfang Leng.
\newblock {XRR}: Extreme multi-label text classification with candidate
  retrieving and deep ranking.
\newblock \emph{Information Sciences}, 622:\penalty0 115--132, 2023.

\bibitem[Yan et~al.(2025)Yan, Liu, and Zhang]{yan2025labelcorank}
Yan Yan, Junyuan Liu, and Bo-Wen Zhang.
\newblock {LabelCoRank}: Revolutionizing long tail multi-label classification
  with co-occurrence reranking.
\newblock \emph{arXiv preprint arXiv:2503.07968}, 2025.

\bibitem[Yang et~al.(2022{\natexlab{a}})Yang, Chen, PourNejatian, Shin, Smith,
  Parisien, Compas, Martin, Costa, Flores, et~al.]{yang2022large}
Xi~Yang, Aokun Chen, Nima PourNejatian, Hoo~Chang Shin, Kaleb~E Smith,
  Christopher Parisien, Colin Compas, Cheryl Martin, Anthony~B Costa, Mona~G
  Flores, et~al.
\newblock A large language model for electronic health records.
\newblock \emph{NPJ digital medicine}, 5\penalty0 (1):\penalty0 194,
  2022{\natexlab{a}}.

\bibitem[Yang et~al.(2022{\natexlab{b}})Yang, Wang, Rawat, Mitra, and
  Yu]{yang2022knowledge}
Zhichao Yang, Shufan Wang, Bhanu Pratap~Singh Rawat, Avijit Mitra, and Hong Yu.
\newblock Knowledge injected prompt based fine-tuning for multi-label few-shot
  {ICD} coding.
\newblock In \emph{Proceedings of the Conference on Empirical Methods in
  Natural Language Processing. Conference on Empirical Methods in Natural
  Language Processing}, volume 2022, page 1767. NIH Public Access,
  2022{\natexlab{b}}.

\bibitem[Yang et~al.(2023{\natexlab{a}})Yang, Batra, Stremmel, and
  Halperin]{yang2023surpassing}
Zhichao Yang, Sanjit~Singh Batra, Joel Stremmel, and Eran Halperin.
\newblock Surpassing {GPT-4} medical coding with a two-stage approach.
\newblock \emph{arXiv preprint arXiv:2311.13735}, 2023{\natexlab{a}}.

\bibitem[Yang et~al.(2023{\natexlab{b}})Yang, Kwon, Yao, and Yu]{yang2023multi}
Zhichao Yang, Sunjae Kwon, Zonghai Yao, and Hong Yu.
\newblock Multi-label few-shot {ICD} coding as autoregressive generation with
  prompt.
\newblock In \emph{Proceedings of the AAAI Conference on Artificial
  Intelligence}, volume~37, pages 5366--5374, 2023{\natexlab{b}}.

\bibitem[Yang et~al.(2023{\natexlab{c}})Yang, Mitra, Liu, Berlowitz, and
  Yu]{yang2023transformehr}
Zhichao Yang, Avijit Mitra, Weisong Liu, Dan Berlowitz, and Hong Yu.
\newblock {TransformEHR}: transformer-based encoder-decoder generative model to
  enhance prediction of disease outcomes using electronic health records.
\newblock \emph{Nature communications}, 14\penalty0 (1):\penalty0 7857,
  2023{\natexlab{c}}.

\bibitem[Ye et~al.(2024)Ye, Sunderraman, and Ji]{ye2024matchxml}
Hui Ye, Rajshekhar Sunderraman, and Shihao Ji.
\newblock {MatchXML}: an efficient text-label matching framework for extreme
  multi-label text classification.
\newblock \emph{IEEE Transactions on Knowledge and Data Engineering},
  36\penalty0 (9):\penalty0 4781--4793, 2024.

\bibitem[You et~al.(2019)You, Zhang, Wang, Dai, Mamitsuka, and
  Zhu]{you2019attentionxml}
Ronghui You, Zihan Zhang, Ziye Wang, Suyang Dai, Hiroshi Mamitsuka, and
  Shanfeng Zhu.
\newblock Attentionxml: Label tree-based attention-aware deep model for
  high-performance extreme multi-label text classification.
\newblock \emph{Advances in neural information processing systems}, 32, 2019.

\bibitem[Yu et~al.(2022)Yu, Zhong, Zhang, Chang, and Dhillon]{yu2022pecos}
Hsiang-Fu Yu, Kai Zhong, Jiong Zhang, Wei-Cheng Chang, and Inderjit~S Dhillon.
\newblock Pecos: Prediction for enormous and correlated output spaces.
\newblock \emph{Journal of Machine Learning Research}, 23\penalty0
  (98):\penalty0 1--32, 2022.

\bibitem[Yu et~al.(2024)Yu, Zhuge, Zhang, Hu, Wang, Lu, and He]{yu2024boosting}
Jiazuo Yu, Yunzhi Zhuge, Lu~Zhang, Ping Hu, Dong Wang, Huchuan Lu, and You He.
\newblock Boosting continual learning of vision-language models via
  mixture-of-experts adapters.
\newblock In \emph{Proceedings of the IEEE/CVF Conference on Computer Vision
  and Pattern Recognition}, pages 23219--23230, 2024.

\bibitem[Yuan et~al.(2025)Yuan, Yoon, Gu, Munby, Walker, Zhu, and
  Eyre]{yuan2025transformers}
Kevin Yuan, Chang~Ho Yoon, Qingze Gu, Henry Munby, A~Sarah Walker, Tingting
  Zhu, and David~W Eyre.
\newblock Transformers and large language models are efficient feature
  extractors for electronic health record studies.
\newblock \emph{Communications Medicine}, 5\penalty0 (1):\penalty0 83, 2025.

\bibitem[Yuan et~al.(2024)Yuan, Xu, Sun, Yu, Wei, and Zhou]{yuan2024research}
Ling Yuan, Xinyi Xu, Ping Sun, Hai~ping Yu, Yin~Zhen Wei, and Jun~jie Zhou.
\newblock Research of multi-label text classification based on label attention
  and correlation networks.
\newblock \emph{PloS one}, 19\penalty0 (9):\penalty0 e0311305, 2024.

\bibitem[Yuan et~al.(2022)Yuan, Tan, and Huang]{yuan-etal-2022-code}
Zheng Yuan, Chuanqi Tan, and Songfang Huang.
\newblock Code synonyms do matter: Multiple synonyms matching network for
  automatic {ICD} coding.
\newblock In \emph{Proceedings of the 60th Annual Meeting of the Association
  for Computational Linguistics (Volume 2: Short Papers)}, pages 808--814,
  Dublin, Ireland, May 2022. Association for Computational Linguistics.
\newblock \doi{10.18653/v1/2022.acl-short.91}.
\newblock URL \url{https://aclanthology.org/2022.acl-short.91}.

\bibitem[Zhang and Wang(2024)]{zhang2024novel}
Bin Zhang and Junli Wang.
\newblock A novel {ICD} coding framework based on associated and hierarchical
  code description distillation.
\newblock \emph{arXiv preprint arXiv:2404.11132}, 2024.

\bibitem[Zhang et~al.(2025{\natexlab{a}})Zhang, Ullah, Schultheis, and
  Babbar]{zhang2025elmo}
Jinbin Zhang, Nasib Ullah, Erik Schultheis, and Rohit Babbar.
\newblock Elmo: Efficiency via low-precision and peak memory optimization in
  large output spaces.
\newblock \emph{arXiv preprint arXiv:2510.11168}, 2025{\natexlab{a}}.

\bibitem[Zhang et~al.(2021)Zhang, Chang, Yu, and Dhillon]{zhang2021fast}
Jiong Zhang, Wei-Cheng Chang, Hsiang-Fu Yu, and Inderjit Dhillon.
\newblock Fast multi-resolution transformer fine-tuning for extreme multi-label
  text classification.
\newblock \emph{Advances in Neural Information Processing Systems},
  34:\penalty0 7267--7280, 2021.

\bibitem[Zhang et~al.(2022)Zhang, Zhang, Zhang, Sang, and
  Yang]{zhang-etal-2022-automatic}
Shurui Zhang, Bozheng Zhang, Fuxin Zhang, Bo~Sang, and Wanchun Yang.
\newblock Automatic {ICD} coding exploiting discourse structure and reconciled
  code embeddings.
\newblock In \emph{Proceedings of the 29th International Conference on
  Computational Linguistics}, pages 2883--2891, Gyeongju, Republic of Korea,
  October 2022. International Committee on Computational Linguistics.
\newblock URL \url{https://aclanthology.org/2022.coling-1.254}.

\bibitem[Zhang et~al.(2025{\natexlab{b}})Zhang, Zhang, Ma, Wang, Wu, Li, and
  Zhou]{zhang2025general}
Xu~Zhang, Kun Zhang, Wenxin Ma, Rongsheng Wang, Chenxu Wu, Yingtai Li, and
  S~Kevin Zhou.
\newblock A general knowledge injection framework for {ICD} coding.
\newblock \emph{arXiv preprint arXiv:2505.18708}, 2025{\natexlab{b}}.

\bibitem[Zhou et~al.(2021)Zhou, Cao, Chen, Liu, Zhao, Niu, Chong, and
  Liu]{zhou-etal-2021-automatic}
Tong Zhou, Pengfei Cao, Yubo Chen, Kang Liu, Jun Zhao, Kun Niu, Weifeng Chong,
  and Shengping Liu.
\newblock Automatic {ICD} coding via interactive shared representation networks
  with self-distillation mechanism.
\newblock In \emph{Proceedings of the 59th Annual Meeting of the Association
  for Computational Linguistics and the 11th International Joint Conference on
  Natural Language Processing (Volume 1: Long Papers)}, pages 5948--5957,
  Online, August 2021. Association for Computational Linguistics.
\newblock \doi{10.18653/v1/2021.acl-long.463}.
\newblock URL \url{https://aclanthology.org/2021.acl-long.463}.

\bibitem[Zhu and Zamani(2023)]{zhu2023icxml}
Yaxin Zhu and Hamed Zamani.
\newblock {ICXML}: An in-context learning framework for zero-shot extreme
  multi-label classification.
\newblock \emph{arXiv preprint arXiv:2311.09649}, 2023.

\end{thebibliography}
